\documentclass[journal]{IEEEtran}
\usepackage{lscape}
\usepackage{times}
\usepackage{graphicx} 
\usepackage[margin=1in]{geometry}
\usepackage{subfigure} 
\usepackage{algpseudocode}
\usepackage{algorithm}
\usepackage{hyperref}

\newcommand{\rec}{\mathrm{rec}}
\newcommand{\nn}{\nonumber}
\usepackage{url}
\usepackage{graphics}
\usepackage{amsmath}
\usepackage{color}
\usepackage{fancyhdr}
\usepackage{datetime}
\usepackage{amssymb}
\usepackage{amsthm}
\graphicspath{{./Figures/}}
\begin{document}
\title{Deep Sentence Embedding Using Long Short-Term Memory Networks: Analysis and Application to Information Retrieval}


\author{Hamid~Palangi, Li~Deng, Yelong~Shen, Jianfeng~Gao, Xiaodong~He, Jianshu~Chen, Xinying~Song, Rabab~Ward
\thanks{H. Palangi and R. Ward  are with the Department
of Electrical and Computer Engineering, University of British Columbia, Vancouver,
BC, V6T 1Z4 Canada (e-mail: \{hamidp,rababw\}@ece.ubc.ca)}
\thanks{L. Deng, Y. Shen, J.Gao, X. He, J. Chen and X. Song are with Microsoft Research, Redmond, WA 98052 USA (e-mail: \{deng,jfgao,xiaohe,yeshen,jianshuc,xinson\}@microsoft.com)}}

\maketitle
\begin{abstract}
This paper develops a model that addresses sentence embedding, a hot topic in current natural language processing research, using recurrent neural networks (RNN) with Long Short-Term Memory (LSTM) cells. The proposed LSTM-RNN model sequentially takes each word in a sentence, extracts its information, and embeds it into a semantic vector. Due to its ability to capture long term memory, the LSTM-RNN accumulates increasingly richer information as it goes through the sentence, and when it reaches the last word, the hidden layer of the network provides a semantic representation of the whole sentence. In this paper, the LSTM-RNN is trained in a \emph{weakly supervised} manner on user click-through data logged by a commercial web search engine. Visualization and analysis are performed to understand how the embedding process works. The model is found to automatically attenuate the unimportant words and detects the salient keywords in the sentence. Furthermore, these detected keywords are found to automatically activate different cells of the LSTM-RNN, where words belonging to a similar topic activate the same cell. As a semantic representation of the sentence, the embedding vector can be used in many different applications. These automatic keyword detection and topic allocation abilities enabled by the LSTM-RNN allow the network to perform document retrieval, a difficult language processing task, where the similarity between the query and documents can be measured by the distance between their corresponding sentence embedding vectors computed by the LSTM-RNN. On a web search task, the LSTM-RNN embedding is shown to significantly outperform several existing state of the art methods. We emphasize that the proposed model generates sentence embedding vectors that are specially useful for web document retrieval tasks. A comparison with a well known general sentence embedding method, the Paragraph Vector, is performed. The results show that the proposed method in this paper significantly outperforms it for web document retrieval task.
\end{abstract}
\IEEEpeerreviewmaketitle

\section{Introduction}
\label{sec:intro}
\IEEEPARstart{L}{earning} a good representation (or features) of input data is an important task in machine learning. In text and language processing, one such problem is  learning of an embedding vector for a sentence; that is, to train a model that can automatically transform a sentence to a vector that encodes the semantic meaning of the sentence. 
While word embedding is learned using a loss function defined on word pairs, sentence embedding is learned using a loss function defined on sentence pairs. In the sentence embedding usually the relationship among words in the sentence, i.e., the context information, is taken into consideration. Therefore, sentence embedding is more suitable for tasks that require computing semantic similarities between text strings.
By mapping texts into a unified semantic representation, the embedding vector can be further used for different language processing applications, such as machine translation \cite{sutskever2014sequence}, sentiment analysis \cite{le2014distributed}, and information retrieval \cite{DSSM}. In machine translation, the recurrent neural networks (RNN) with Long Short-Term Memory (LSTM) cells, or the LSTM-RNN, is used to encode an English sentence into a vector, which contains the semantic meaning of the input sentence, and then another LSTM-RNN is used to generate a French (or another target language) sentence from the vector. The model is trained to best predict the output sentence. In \cite{le2014distributed}, a paragraph vector is learned in an unsupervised manner as a distributed representation of sentences and documents, which are then used for sentiment analysis. Sentence embedding can also be applied to information retrieval, where the contextual information are properly represented by the vectors in the same space for fuzzy text matching \cite{DSSM}.

In this paper, we propose to use an RNN to sequentially accept each word in a sentence and recurrently map it into a latent space together with the historical information. As the RNN reaches the last word in the sentence, the hidden activations form a natural embedding vector for the contextual information of the sentence. We further incorporate the LSTM cells into the RNN model (i.e. the LSTM-RNN) to address the difficulty of learning long term memory in RNN. The learning of such a model is performed in a \emph{weakly supervised} manner on the click-through data logged by a commercial web search engine. Although manually labelled data are insufficient in machine learning, logged data with limited feedback signals are massively available due to the widely used commercial web search engines. Limited feedback information such as click-through data provides a weak supervision signal that indicates the semantic similarity between the text on the query side and the clicked text on the document side. To exploit such a signal, the objective of our training is to maximize the similarity between the two vectors mapped by the LSTM-RNN from the query and the clicked document, respectively. Consequently, the learned embedding vectors of the query and clicked document are specifically useful for web document retrieval task.

An important contribution of this paper is to analyse the embedding process of the LSTM-RNN by visualizing the internal activation behaviours in response to different text inputs. We show that the embedding process of the learned LSTM-RNN effectively detects the keywords, while attenuating less important words, in the sentence automatically by switching on and off the gates within the LSTM-RNN cells.  We further show that different cells in the learned model indeed correspond to different topics, and the keywords associated with a similar topic activate the same cell unit in the model. As the LSTM-RNN reads to the end of the sentence, the topic activation accumulates and the hidden vector at the last word encodes the rich contextual information of the entire sentence. For this reason, a natural application of sentence embedding is web search ranking, in which the embedding vector from the query can be used to match the embedding vectors of the candidate documents according to the maximum cosine similarity rule. Evaluated on a real web document ranking task, our proposed method significantly outperforms many of the existing state of the art methods in NDCG scores. Please note that when we refer to document in the paper we mean the title (headline) of the document. 

\section{Related Work}
\label{sec:related work}

Inspired by the word embedding method \cite{mikolov2013distributed,mikolov2013efficient}, the authors in \cite{le2014distributed} proposed an unsupervised learning method to learn a paragraph vector as a distributed representation of sentences and documents, which are then used for sentiment analysis with superior performance. However, the model is not designed to capture the fine-grained sentence structure. In \cite{Skip-Thought}, an unsupervised sentence embedding method is proposed with great performance on large corpus of contiguous text corpus, e.g., the BookCorpus \cite{BookCorpus}. The main idea is to encode the sentence $s(t)$ and then decode previous and next sentences, i.e., $s(t-1)$ and $s(t+1)$, using two separate decoders. The encoder and decoders are RNNs with Gated Recurrent Unit (GRU) \cite{GRU}. However, this sentence embedding method is not designed for document retrieval task having a supervision among queries and clicked and unclicked documents. In \cite{RAE}, a Semi-Supervised Recursive Autoencoder (RAE) is proposed and used for sentiment prediction. Similar to our proposed method, it does not need any language specific sentiment parsers. A greedy approximation method is proposed to construct a tree structure for the input sentence. It assigns a vector per word. It can become practically problematic for large vocabularies. It also works both on unlabeled data and supervised sentiment data.

Similar to the recurrent models in this paper, The DSSM \cite{DSSM} and CLSM \cite{CDSSM} models, developed for information retrieval,  can also be interpreted as sentence embedding methods. However, DSSM treats the input sentence as a bag-of-words and does not model word dependencies explicitly. CLSM treats a sentence as a bag of n-grams, where n is defined by a window, and can capture \emph{local} word dependencies. Then a Max-pooling layer is used to form a global feature vector. Methods in \cite{collobert:2008} are also convolutional based networks for Natural Language Processing (NLP). These models, by design, cannot capture long distance dependencies, i.e., dependencies among words belonging to non-overlapping n-grams. In \cite{DCNN} a Dynamic Convolutional Neural Network (DCNN) is proposed for sentence embedding. Similar to CLSM, DCNN does not rely on a parse tree and is easily applicable to any language. However, different from CLSM where a regular max-pooling is used, in DCNN a dynamic $k$-max-pooling is used. This means that instead of just keeping the largest entries among word vectors in one vector, $k$ largest entries are kept in $k$ different vectors. DCNN has shown good performance in sentiment prediction and question type classification tasks. In \cite{CNN_Matching}, a convolutional neural network architecture is proposed for sentence matching. It has shown great performance in several matching tasks. In \cite{PhraseEmbedding}, a Bilingually-constrained Recursive Auto-encoders (BRAE) is proposed to create semantic vector representation for phrases. Through experiments it is shown that the proposed method has great performance in two end-to-end SMT tasks. 

Long short-term memory networks were developed in \cite{lstm} to address the difficulty of capturing long term memory in RNN. It has been successfully applied to speech recognition, which achieves state-of-art performance \cite{graves2013speech,sak2014long}. In text analysis, LSTM-RNN treats a sentence as a sequence of words with internal structures, i.e., word dependencies. It encodes a semantic vector of a sentence incrementally which differs from DSSM and CLSM. The encoding process is performed left-to-right, word-by-word. At each time step, a new word is encoded into the semantic vector, and the word dependencies embedded in the vector are ``updated''. When the process reaches the end of the sentence, the semantic vector has embedded all the words and their dependencies, hence, can be viewed as a feature vector representation of the whole sentence. In the machine translation work \cite{sutskever2014sequence}, an input English sentence is converted into a vector representation using LSTM-RNN, and then another LSTM-RNN is used to generate an output French sentence. The model is trained to maximize the probability of predicting the correct output sentence. In \cite{hermann2014multilingual}, there are two main composition models, “ADD” model that is bag of words and “BI” model that is a summation over bi-gram pairs plus a non-linearity. In our proposed model, instead of simple summation, we have used LSTM model with letter tri-grams which keeps valuable information over long intervals (for long sentences) and throws away useless information. In \cite{NMTbengio}, an encoder-decoder approach is proposed to jointly learn to align and translate sentences from English to French using RNNs. The concept of ``attention'' in the decoder, discussed in this paper, is closely related to how our proposed model extracts keywords in the document side. For further explanations please see section \ref{sec:exKeyWord}. In \cite{RNNvisualization} a set of visualizations are presented for RNNs with and without LSTM cells and GRUs. Different from our work where the target task is sentence embedding for document retrieval, the target tasks in \cite{RNNvisualization} were character level sequence modelling for text characters and source codes. Interesting observations about interpretability of some LSTM cells and statistics of gates activations are presented. In section \ref{sec:Analysis} we show that some of the results of our visualization are consistent with the observations reported in \cite{RNNvisualization}. We also present more detailed visualization specific to the document retrieval task using click-through data. We also present visualizations about how our proposed model can be used for keyword detection. 

Different from the aforementioned studies, the method developed in this paper trains the model so that sentences that are paraphrase of each other are close in their semantic embedding vectors --- see the description in Sec. \ref{sec:LearningMethod} further ahead. Another reason that LSTM-RNN is particularly effective for sentence embedding, is its robustness to noise. For example, in the web document ranking task, the noise comes from two sources: (i) Not every word in query / document is equally important, and we only want to ``remember'' salient words using the limited ``memory''. (ii) A word or phrase that is important to a document may not be relevant to a given query, and we only want to ``remember'' related words that are useful to compute the relevance of the document for a given query. We will illustrate robustness of LSTM-RNN in this paper. The structure of LSTM-RNN will also circumvent the serious limitation of using a fixed window size in CLSM. Our experiments show that this difference leads to significantly better results in web document retrieval task. Furthermore, it has other advantages. It allows us to capture keywords and key topics effectively. The models in this paper also do not need the extra max-pooling layer, as required by the CLSM, to capture global contextual information and they do so more effectively.


\section{Sentence Embedding Using RNNs with and without LSTM Cells}
\label{sec:senEmbed}

In this section, we introduce the model of recurrent neural networks and its long short-term memory version for learning the sentence embedding vectors. We start with the basic RNN and then proceed to LSTM-RNN.

\subsection{The basic version of RNN}
\label{sec:RDSSM}
\begin{figure}[t]
\center
\includegraphics[width=0.45\textwidth]{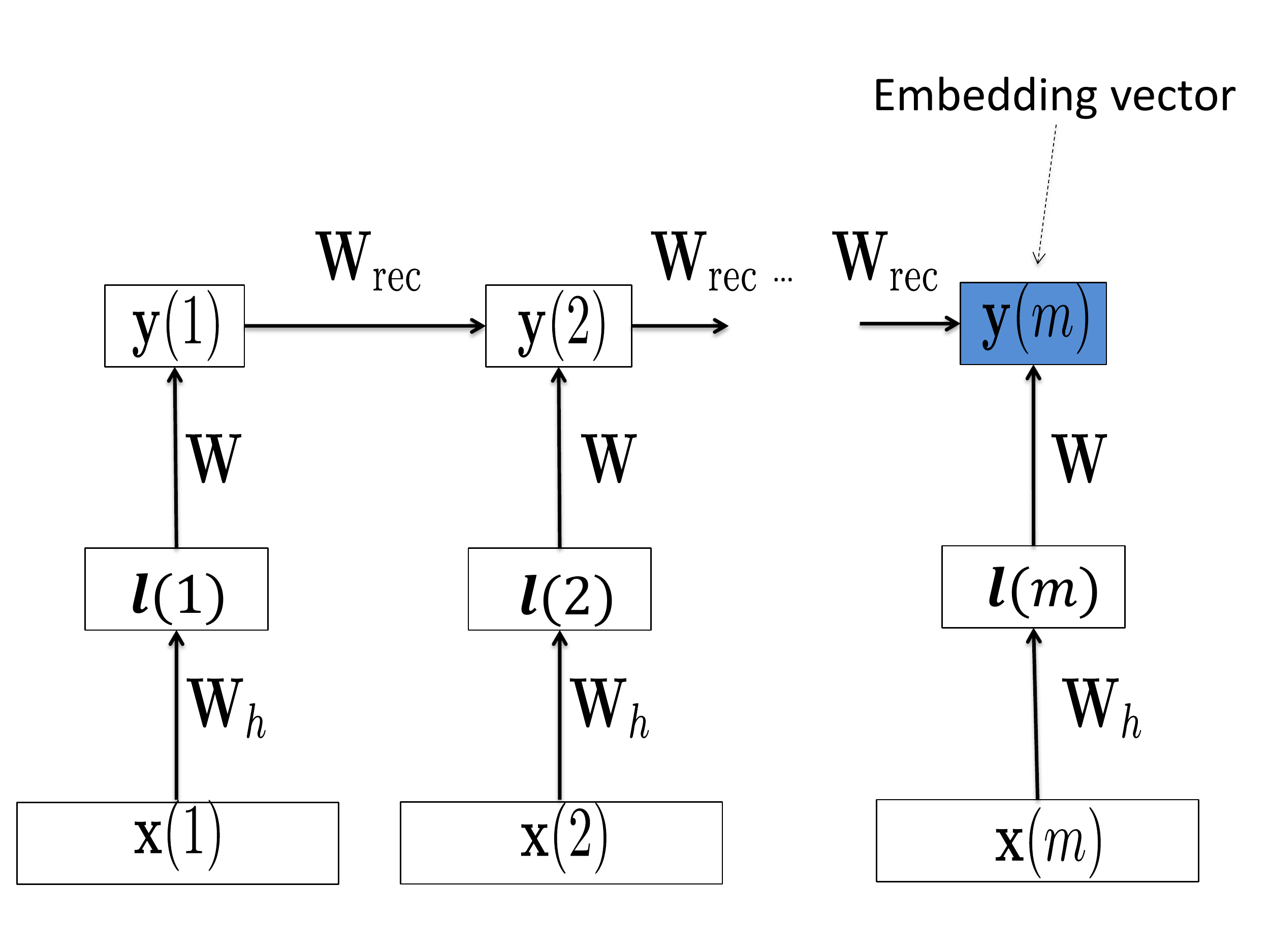}
\caption{The basic architecture of the RNN for sentence embedding, where temporal recurrence is used to model the contextual information across words in the text string. The hidden activation vector corresponding to the last word is the sentence embedding vector (blue).}
\label{Fig:Fig_RNNSentenceEmbed}
\end{figure}

The RNN is a type of deep neural networks that are ``deep'' in temporal dimension and it has been used extensively in time sequence modelling \cite{RNNref,robinson1994application,deng1994analysis,mikolov2010recurrent,graves2012sequence,bengio2013advances,PrimalDual,mesnilinvestigation,DengChen14}. The main idea of using RNN for sentence embedding is to find a dense and low dimensional semantic representation by sequentially and recurrently processing each word in a sentence and mapping it into a low dimensional vector. In this model, the global contextual features of the whole text will be in the semantic representation of the last word in the text sequence --- see Figure \ref{Fig:Fig_RNNSentenceEmbed}, where $\mathbf{x}(t)$ is the $t$-th word, coded as a 1-hot vector, $\mathbf{W}_h$ is a fixed hashing operator similar to the one used in \cite{DSSM} that converts the word vector to a letter tri-gram vector, $\mathbf{W}$ is the input weight matrix, $\mathbf{W}_{rec}$ is the recurrent weight matrix, $\mathbf{y}(t)$ is the hidden activation vector of the RNN, which can be used as a semantic representation of the $t$-th word, and $\mathbf{y}(t)$ associated to the last word $\mathbf{x}(m)$ is the semantic representation vector of the entire sentence. Note that this is very different from the approach in \cite{DSSM} where the bag-of-words representation is used for the whole text and no context information is used. This is also different from \cite{CDSSM} where the sliding window of a fixed size (akin to an FIR filter) is used to capture local features and a max-pooling layer on the top to capture global features. In the RNN there is neither a fixed-sized window nor a max-pooling layer; rather the recurrence is used to capture the context information in the sequence (akin to an IIR filter). 



The mathematical formulation of the above RNN model for sentence embedding can be expressed as
\begin{align}
\label{eq:model}
&\mathbf{l}(t) = \mathbf{W}_h\mathbf{x}(t)\nn\\
&\mathbf{y}(t) = \textit{f}( \mathbf{W}\mathbf{l}(t) + \mathbf{W}_{\rec}\mathbf{y}(t-1) + \mathbf{b} )
\end{align}
where $\mathbf{W}$ and $\mathbf{W}_{\rec}$ are the input and recurrent matrices to be learned, $\mathbf{W}_h$ is a fixed word hashing operator, $\mathbf{b}$ is the bias vector and $f(\cdot)$ is assumed to be $\tanh(\cdot)$. Note that the architecture proposed here for sentence embedding is slightly different from traditional RNN in that there is a word hashing layer that convert the high dimensional input into a relatively lower dimensional letter tri-gram representation. There is also no per word supervision during training, instead, the whole sentence has a label. This is explained in more detail in section \ref{sec:LearningMethod}.

\subsection{The RNN with LSTM cells}
\label{sec:LSTM-DSSM}
Although RNN performs the transformation from the sentence to a vector in a principled manner, it is generally difficult to learn the long term dependency within the sequence due to vanishing gradients problem. One of the effective solutions for this problem in RNNs is using memory cells instead of neurons originally proposed in \cite{lstm} as Long Short-Term Memory (LSTM) and completed in \cite{lstm_forget} and \cite{lstm_peephole} by adding forget gate and peephole connections to the architecture. 

\begin{figure}[t]
\center
\includegraphics[width=0.48\textwidth]{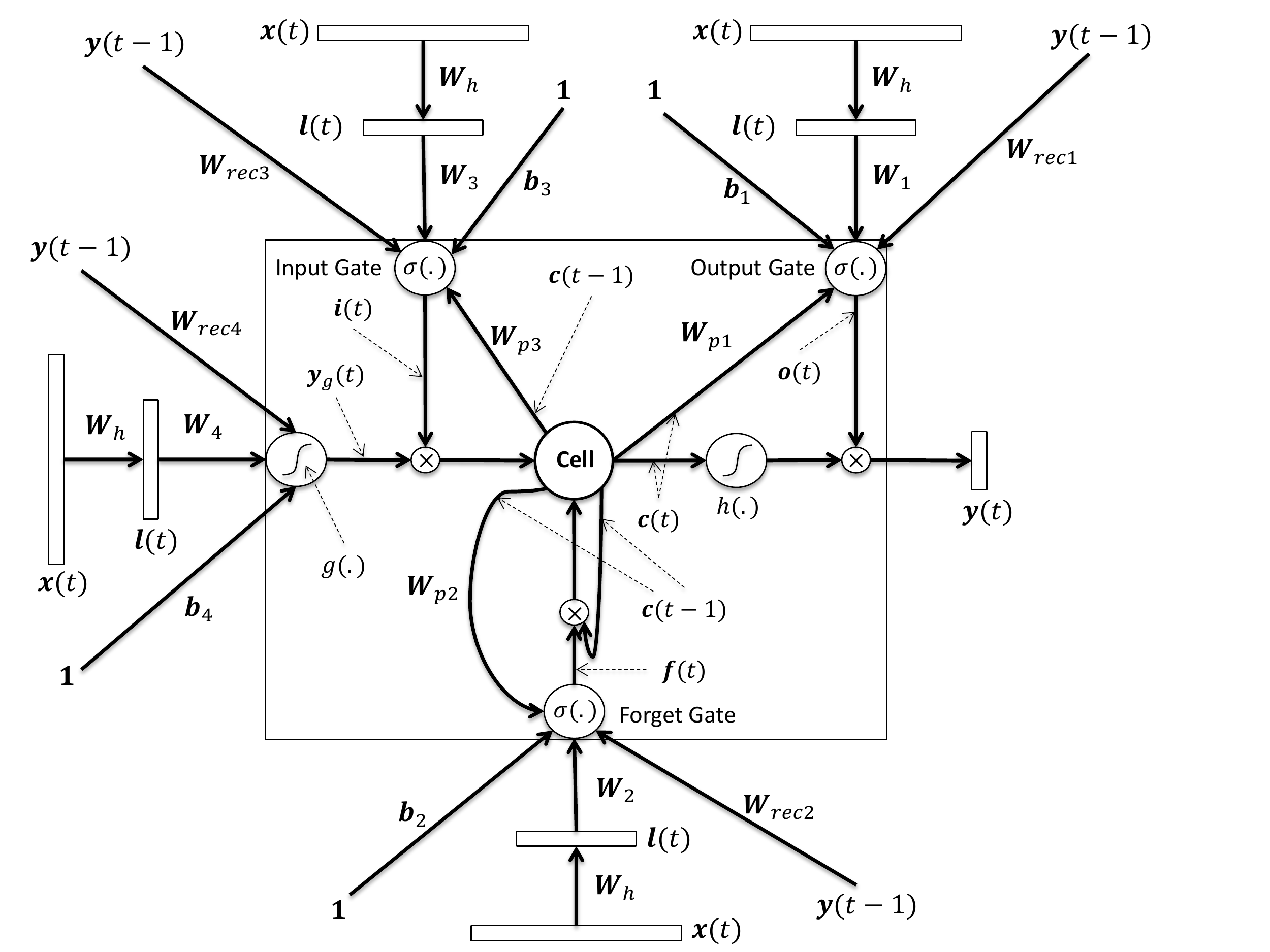}
\caption{The basic LSTM architecture used for sentence embedding}
\label{fig:LSTM Architecture}
\end{figure}

We use the architecture of LSTM illustrated in Fig. \ref{fig:LSTM Architecture} for the proposed sentence embedding method. In this figure, $\mathbf{i}(t),\;\mathbf{f}(t)\;,\mathbf{o}(t)\;, \mathbf{c}(t)$ are input gate, forget gate, output gate and cell state vector respectively, $\mathbf{W}_{p1},\;\mathbf{W}_{p2}$ and $\mathbf{W}_{p3}$ are peephole connections, $\mathbf{W}_i$, $\mathbf{W}_{reci}$ and  $\mathbf{b}_i$, $i=1,2,3,4$ are input connections, recurrent connections and bias values, respectively, $g(\cdot)$ and $h(\cdot)$ are $\tanh(\cdot)$ function and $\sigma(\cdot)$ is the sigmoid function. We use this architecture to find $\mathbf{y}$ for each word, then use the $\mathbf{y}(m)$ corresponding to the last word in the sentence as the semantic vector for the entire sentence. 

Considering Fig.~\ref{fig:LSTM Architecture}, the forward pass for LSTM-RNN model is as follows:
\begin{align}
\label{eq:lstm_forward}
&\mathbf{y}_g(t) = g(\mathbf{W}_4\mathbf{l}(t) + \mathbf{W}_{\rec 4}\mathbf{y}(t-1) + \mathbf{b}_4)\nn\\
&\mathbf{i}(t) = \sigma(\mathbf{W}_3\mathbf{l}(t) + \mathbf{W}_{\rec 3}\mathbf{y}(t-1) + \mathbf{W}_{p3}\mathbf{c}(t-1) + \mathbf{b}_3)\nn\\
&\mathbf{f}(t) = \sigma(\mathbf{W}_2\mathbf{l}(t) + \mathbf{W}_{\rec 2}\mathbf{y}(t-1) + \mathbf{W}_{p2}\mathbf{c}(t-1) + \mathbf{b}_2)\nn\\
&\mathbf{c}(t) = \mathbf{f}(t)\circ\mathbf{c}(t-1) + \mathbf{i}(t)\circ\mathbf{y}_g(t)\nn\\
&\mathbf{o}(t) = \sigma(\mathbf{W}_1\mathbf{l}(t) + \mathbf{W}_{\rec 1}\mathbf{y}(t-1) + \mathbf{W}_{p1}\mathbf{c}(t) + \mathbf{b}_1)\nn\\
&\mathbf{y}(t) = \mathbf{o}(t)\circ h(\mathbf{c}(t))
\end{align}
where $\circ$ denotes Hadamard (element-wise) product. A diagram of the proposed model with more details is presented in section VI of Supplementary Materials.

\section{Learning Method}
\label{sec:LearningMethod}

To learn a good semantic representation of the input sentence, our objective is \emph{to make the embedding vectors for sentences of similar meaning as close as possible, and meanwhile, to make sentences of different meanings as far apart as possible}. This is challenging in practice since it is hard to collect a large amount of manually labelled data that give the semantic similarity signal between different sentences. Nevertheless, the widely used commercial web search engine is able to log massive amount of data with some limited user feedback signals. For example, given a particular query, the click-through information about the user-clicked document among many candidates is usually recorded and can be used as a weak (binary) supervision signal to indicate the semantic similarity between two sentences (on the query side and the document side). In this section, we explain how to leverage such a weak supervision signal to learn a sentence embedding vector that achieves the aforementioned training objective. Please also note that above objective to make sentences with similar meaning as close as possible is similar to machine translation tasks where two sentences belong to two different languages with similar meanings and we want to make their semantic representation as close as possible.

\begin{figure}[t]
\centering
\includegraphics[width=0.45\textwidth]{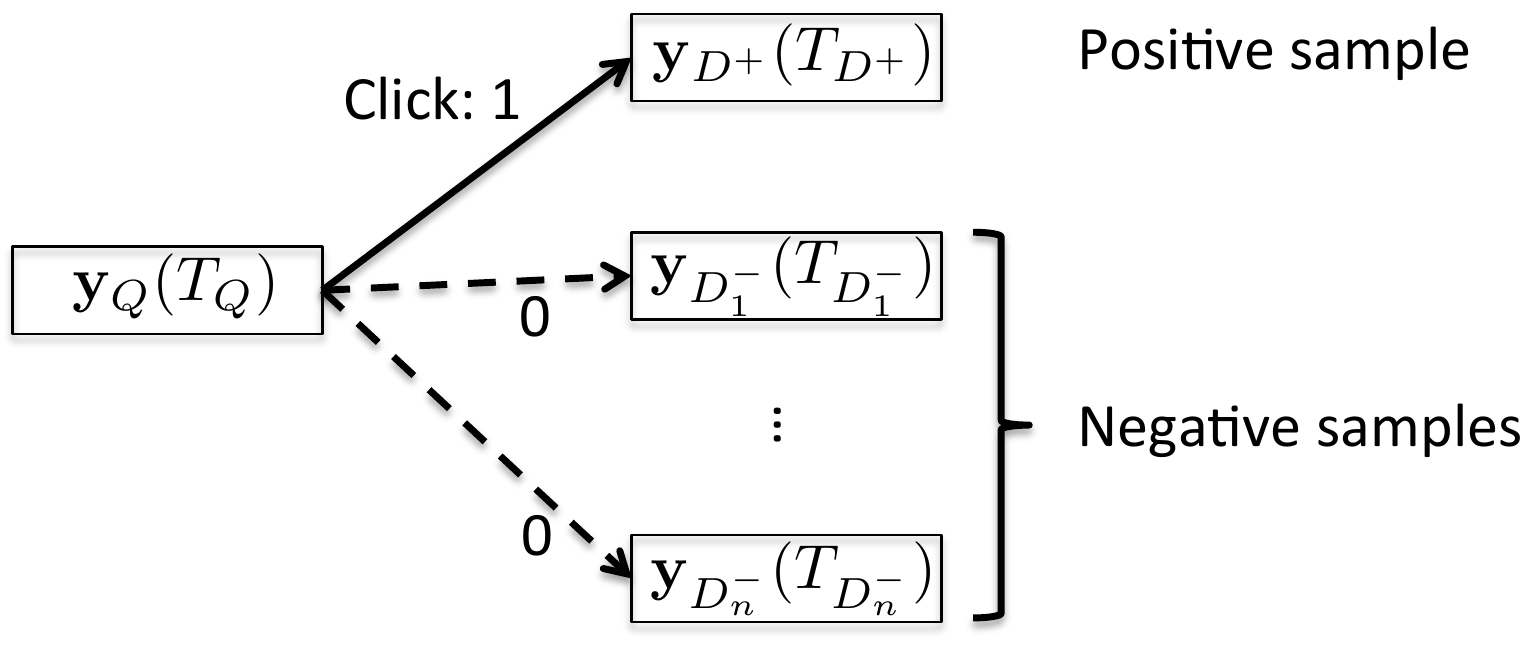}
\caption{The click-through signal can be used as a (binary) indication of the semantic similarity between the sentence on the query side and the sentence on the document side. The negative samples are randomly sampled from the training data.}
\label{Fig:Fig_ClickThroughMatching}
\end{figure}

We now describe how to train the model to achieve the above objective using the click-through data logged by a commercial search engine. For a complete description of the click-through data please refer to section 2 in \cite{ClickThroughRef}. To begin with, we adopt the cosine similarity between the semantic vectors of two sentences as a measure for their similarity:
\begin{equation}
\label{eq:similarity_func}
R(Q,D) = \frac{\mathbf{y}_Q(T_Q)^T\mathbf{y}_D(T_D)}{\Vert\mathbf{y}_Q(T_Q)\Vert \cdot \Vert\mathbf{y}_D(T_D)\Vert}
\end{equation}
where $T_Q$ and $T_D$ are the lengths of the sentence $Q$ and sentence $D$, respectively. In the context of training over click-through data, we will use $Q$ and $D$ to denote ``query'' and ``document'', respectively. In Figure \ref{Fig:Fig_ClickThroughMatching}, we show the sentence embedding vectors corresponding to the query, $\mathbf{y}_Q(T_Q)$, and all the documents, $\{\mathbf{y}_{D^{+}}(T_{D^{+}}), \mathbf{y}_{D_1^{-}}(T_{D_1^{-}}), \ldots, \mathbf{y}_{D_n^{-}}(T_{D_n^{-}})\}$, where the subscript $D^{+}$ denotes the (clicked) positive sample among the documents, and the subscript $D_j^{-}$ denotes the $j$-th (un-clicked) negative sample. All these embedding vectors are generated by feeding the sentences into the RNN or LSTM-RNN model described in Sec. \ref{sec:senEmbed} and take the $\mathbf{y}$ corresponding to the last word --- see the blue box in Figure \ref{Fig:Fig_RNNSentenceEmbed}.

We want to maximize the likelihood of the clicked document given query, which can be formulated as the following optimization problem:
\begin{equation}
\label{eq:objective_func}
L(\mathbf{\Lambda})=\underset{\mathbf{\Lambda}}{\operatorname{min}} 
\left\{ -\log \prod_{r=1}^{N}P(D_r^+\vert Q_r) \right\} = 
\underset{\mathbf{\Lambda}}{\operatorname{min}} \sum_{r=1}^{N}l_r(\mathbf{\Lambda})
\end{equation}
where $\mathbf{\Lambda}$ denotes the collection of the model parameters; in regular RNN case, it includes $ \mathbf{W}_{rec}$ and $\mathbf{W}$ in Figure \ref{Fig:Fig_RNNSentenceEmbed}, and in LSTM-RNN case, it includes $\mathbf{W}_1$, $\mathbf{W}_2$, $\mathbf{W}_3$, $\mathbf{W}_4$, $\mathbf{W}_{rec1}$, $\mathbf{W}_{rec2}$, $\mathbf{W}_{rec3}$, $\mathbf{W}_{rec4}$, $\mathbf{W}_{p1}$, $\mathbf{W}_{p2}$, $\mathbf{W}_{p3}$, $\mathbf{b}_1$, $\mathbf{b}_2$, $\mathbf{b}_3$ and $\mathbf{b}_4$ in Figure \ref{fig:LSTM Architecture}. $D_r^+$ is the clicked document for $r$-th query, $P(D_r^+\vert Q_r)$ is the probability of clicked document given the $r$-th query, $N$ is number of query / clicked-document pairs in the corpus and 
\begin{align}
\label{eq:L_r}
l_r(\mathbf{\Lambda}) 	
					&= 		-\log\left( 
								\frac{e^{\gamma R(Q_r,D_r^+)}}
								{
									e^{\gamma R(Q_r,D_r^+)}
									+
									\sum_{i=j}^n 
									e^{\gamma R(Q_r,D_{r,j}^{-})}
								} 
							\right) 
							\nn\\
					&= 		\log\left( 
								1 
								+ 
								\sum_{j=1}^n 
								e^{- \gamma \cdot \Delta_{r,j}}
							\right)
\end{align}
where $\Delta_{r,j} = R(Q_r,D_r^+) - R(Q_r,D_{r,j}^{-})$, $R(\cdot,\cdot)$ was defined earlier in \eqref{eq:similarity_func}, $D_{r,j}^{-}$ is the $j$-th negative candidate document for $r$-th query and $n$ denotes the number of negative samples used during training. 

The expression in \eqref{eq:L_r} is a logistic loss over $\Delta_{r,j}$. It upper-bounds the pairwise accuracy, i.e., the 0 - 1 loss. Since the similarity measure is the cosine function, $\Delta_{r,j} \in [-2,2]$. To have a larger range for $\Delta_{r,j}$, we use $\gamma$ for scaling. It helps to penalize the prediction error more. Its value is set empirically by experiments on a held out dataset. 

To train the RNN and LSTM-RNN, we use Back Propagation Through Time (BPTT). The update equations for parameter $\mathbf{\Lambda}$ at epoch $k$ are as follows:
\begin{align}
\label{eq:Nesterov}
&\triangle\mathbf{\Lambda} _k = \mathbf{\Lambda} _k - \mathbf{\Lambda} _{k-1}\nn\\
&\triangle\mathbf{\Lambda} _k = \mu_{k-1}\triangle\mathbf{\Lambda} _{k-1} - \epsilon_{k-1}\nabla L(\mathbf{\Lambda} _{k-1} + \mu_{k-1}\triangle\mathbf{\Lambda} _{k-1})
\end{align}
where $\nabla L(\cdot)$ is the gradient of the cost function in \eqref{eq:objective_func}, $\epsilon$ is the learning rate and $\mu_k$ is a momentum parameter determined by the scheduling scheme used for training. Above equations are equivalent to Nesterov
 method in \cite{Nestrov1983}. To see why, please refer to appendix A.1 of \cite{RNNhinton2013} where Nesterov method is derived as a momentum method. The gradient of the cost function, $\nabla L(\mathbf{\Lambda})$, is: 
\begin{equation}
\label{eq:rdssmTrain1}
\nabla L(\mathbf{\Lambda}) = \underbrace{-\sum_{r=1}^N\sum_{j=1}^n\sum_{\tau=0}^{T}\alpha_{r,j}\frac{\partial \Delta_{r,j,\tau}}{\partial\mathbf{\Lambda}}}_{\mathrm{one\; large\; update}}
\end{equation}
where $T$ is the number of time steps that we unfold the network over time and
\begin{equation}
\label{eq:rdssmTrain2}
\alpha_{r,j} = \frac{-\gamma e^{-\gamma\Delta_{r,j}}}{1+\sum_{j=1}^ne^{-\gamma \Delta_{r,j}}}.
\end{equation}

$\frac{\partial \Delta_{r,j,\tau}}{\partial\mathbf{\Lambda}}$ in \eqref{eq:rdssmTrain1} and error signals for different parameters of RNN and LSTM-RNN that are necessary for training are presented in Appendix \ref{sec:appGradientExpress}. Full derivation of gradients in both models is presented in section III of supplementary materials.

To accelerate training by parallelization, we use mini-batch training and one large update instead of incremental updates during back propagation through time. To resolve the gradient explosion problem we use gradient re-normalization method described in \cite{RazvanRNN,mikolov2010recurrent}. To accelerate the convergence, we use Nesterov method \cite{Nestrov1983} and found it effective in training both RNN and LSTM-RNN for sentence embedding. 

We have used a simple yet effective scheduling for $\mu_k$ for both RNN and LSTM-RNN models, in the first and last 2\% of all parameter updates $\mu_k = 0.9$ and for the other 96\% of all parameter updates $\mu_k = 0.995$. We have used a fixed step size for training RNN and a fixed step size for training LSTM-RNN.

A summary of training method for LSTM-RNN is presented in Algorithm \ref{alg1}.
\begin{algorithm}[t]
\caption{Training LSTM-RNN for Sentence Embedding}
\label{alg1}
\begin{algorithmic}
\scriptsize
\State \textbf{Inputs}: Fixed step size ``$\epsilon$'', Scheduling for ``$\mu$'', Gradient clip threshold ``$th_G$'', Maximum number of Epochs ``$nEpoch$'', Total number of query / clicked-document pairs ``$N$'', Total number of un-clicked (negative) documents for a given query ``$n$'', Maximum sequence length for truncated BPTT ``$T$''. 
\State \textbf{Outputs}: Two trained models, one in query side ``$\mathbf{\Lambda}_Q$'', one in document side ``$\mathbf{\Lambda}_D$''.
\State \textbf{Initialization}: Set all parameters in $\mathbf{\Lambda}_Q$ and $\mathbf{\Lambda}_D$ to small random numbers, $i=0$, $k=1$.
\Procedure {LSTM-RNN}{$\mathbf{\Lambda}_Q$,$\mathbf{\Lambda}_D$}
\While {$i \leq nEpoch$}
\For {``first minibatch'' $\rightarrow$ ``last minibatch''}
\State $r \leftarrow 1$
\While{$r \leq N$}
\For{$j=1 \rightarrow n$}
\State Compute $\alpha_{r,j}$\Comment{use \eqref{eq:rdssmTrain2}}
\State Compute $\sum_{\tau=0}^{T}\alpha_{r,j}\frac{\partial \Delta_{r,j,\tau}}{\partial\mathbf{\Lambda}_{k,Q}}$
\State \Comment{use \eqref{eq:lstmTrain1} to \eqref{eq:LSTM47} in appendix \ref{sec:appGradientExpress}}
\State Compute $\sum_{\tau=0}^{T}\alpha_{r,j}\frac{\partial \Delta_{r,j,\tau}}{\partial\mathbf{\Lambda}_{k,D}}$
\State \Comment{use \eqref{eq:lstmTrain1} to \eqref{eq:LSTM47} in appendix \ref{sec:appGradientExpress}}
\State sum above terms for $Q$ and $D$ over $j$
\EndFor
\State sum above terms for $Q$ and $D$ over $r$
\State $r \leftarrow r+1$
\EndWhile
\State Compute $\nabla L(\mathbf{\Lambda}_{k,Q})$\Comment{use \eqref{eq:rdssmTrain1}}
\State Compute $\nabla L(\mathbf{\Lambda}_{k,D})$\Comment{use \eqref{eq:rdssmTrain1}}
\If{$\| \nabla L(\mathbf{\Lambda}_{k,Q})\| > th_G$}
\State $\nabla L(\mathbf{\Lambda}_{k,Q}) \leftarrow th_G \cdot \frac{\nabla L(\mathbf{\Lambda}_{k,Q}) }{\|\nabla L(\mathbf{\Lambda}_{k,Q}) \|}$
\EndIf
\If{$ \| \nabla L(\mathbf{\Lambda}_{k,D})\| > th_G$}
\State $\nabla L(\mathbf{\Lambda}_{k,D}) \leftarrow th_G \cdot \frac{\nabla L(\mathbf{\Lambda}_{k,D})}{\|\nabla L(\mathbf{\Lambda}_{k,D})\|}$
\EndIf
\State Compute $\triangle\mathbf{\Lambda} _{k,Q}$\Comment {use \eqref{eq:Nesterov}}
\State Compute $\triangle\mathbf{\Lambda} _{k,D}$\Comment {use \eqref{eq:Nesterov}}
\State Update: $\mathbf{\Lambda} _{k,Q} \leftarrow \triangle\mathbf{\Lambda} _{k,Q} + \mathbf{\Lambda} _{k-1,Q}$
\State Update: $\mathbf{\Lambda} _{k,D} \leftarrow \triangle\mathbf{\Lambda} _{k,D} + \mathbf{\Lambda} _{k-1,D}$
\State $k \leftarrow k+1$
\EndFor
\State $i \leftarrow i+1$
\EndWhile
\EndProcedure
\end{algorithmic}
\end{algorithm}

\section{Analysis of the Sentence Embedding Process and Performance Evaluation}
\label{sec: Experiments}

To understand how the LSTM-RNN performs sentence embedding, we use visualization tools to analyze the semantic vectors generated by our model. We would like to answer the following questions: (i) How are word dependencies and context information captured? (ii) How does LSTM-RNN attenuate unimportant information and detect critical information from the input sentence? Or, how are the keywords embedded into the semantic vector? (iii) How are the global topics identified by LSTM-RNN?

To answer these questions, we train the RNN with and without LSTM cells on the click-through dataset which are logged by a commercial web search engine. The training method has been described in Sec. \ref{sec:LearningMethod}. Description of the corpus is as follows. The training set includes 200,000 positive query / document pairs where only the
clicked signal is used as a weak supervision for training LSTM. The relevance judgement
set (test set) is constructed as follows. First, the queries are sampled from a
year of search engine logs. Adult, spam, and bot queries are all removed. Queries are
“de-duped” so that only unique queries remain. To reflex a natural query distribution,
we do not try to control the quality of these queries. For example, in our query sets,
there are around 20\% misspelled queries, and around 20\% navigational queries and
10\% transactional queries, etc. Second, for each query, we collect Web documents to
be judged by issuing the query to several popular search engines (e.g., Google, Bing)
and fetching top-10 retrieval results from each. Finally, the query-document pairs are
judged by a group of well-trained assessors. In this study all the queries are preprocessed
as follows. The text is white-space tokenized and lower-cased, numbers are
retained, and no stemming/inflection treatment is performed. Unless stated otherwise, in the experiments we used 4 negative samples, i.e., $n=4$ in Fig. \ref{Fig:Fig_ClickThroughMatching}.

We now proceed to perform a  comprehensive analysis by visualizing the trained RNN and LSTM-RNN models. In particular, we will visualize the on-and-off behaviors of the input gates, output gates, cell states, and the semantic vectors in LSTM-RNN model, which reveals how the model extracts useful information from the input sentence and embeds it properly into the semantic vector according to the topic information.

Although giving the full learning formula for all the model parameters in the previous section, we will remove the peephole connections and the forget gate from the LSTM-RNN model in the current task. This is because the length of each sequence, i.e., the number of words in a query or a document, is known in advance, and we set the state of each cell to zero in the beginning of a new sequence. Therefore, forget gates are not a great help here. Also, as long as the order of words is kept, the precise timing in the sequence is not of great concern. Therefore, peephole connections are not that important as well. Removing peephole connections and forget gate will also reduce the amount of training time, since a smaller number of parameters need to be learned.

\subsection{Analysis}
\label{sec:Analysis}

In this section we would like to examine how the information in the input sentence is sequentially extracted and embedded into the semantic vector over time by the LSTM-RNN model. 

\subsubsection{Attenuating Unimportant Information}
\label{sec:lstm_semantic_vectors}

\begin{figure}[t]
\centerline{
\subfigure[$\mathbf{i}(t)$]{\includegraphics[width = 0.24\textwidth]{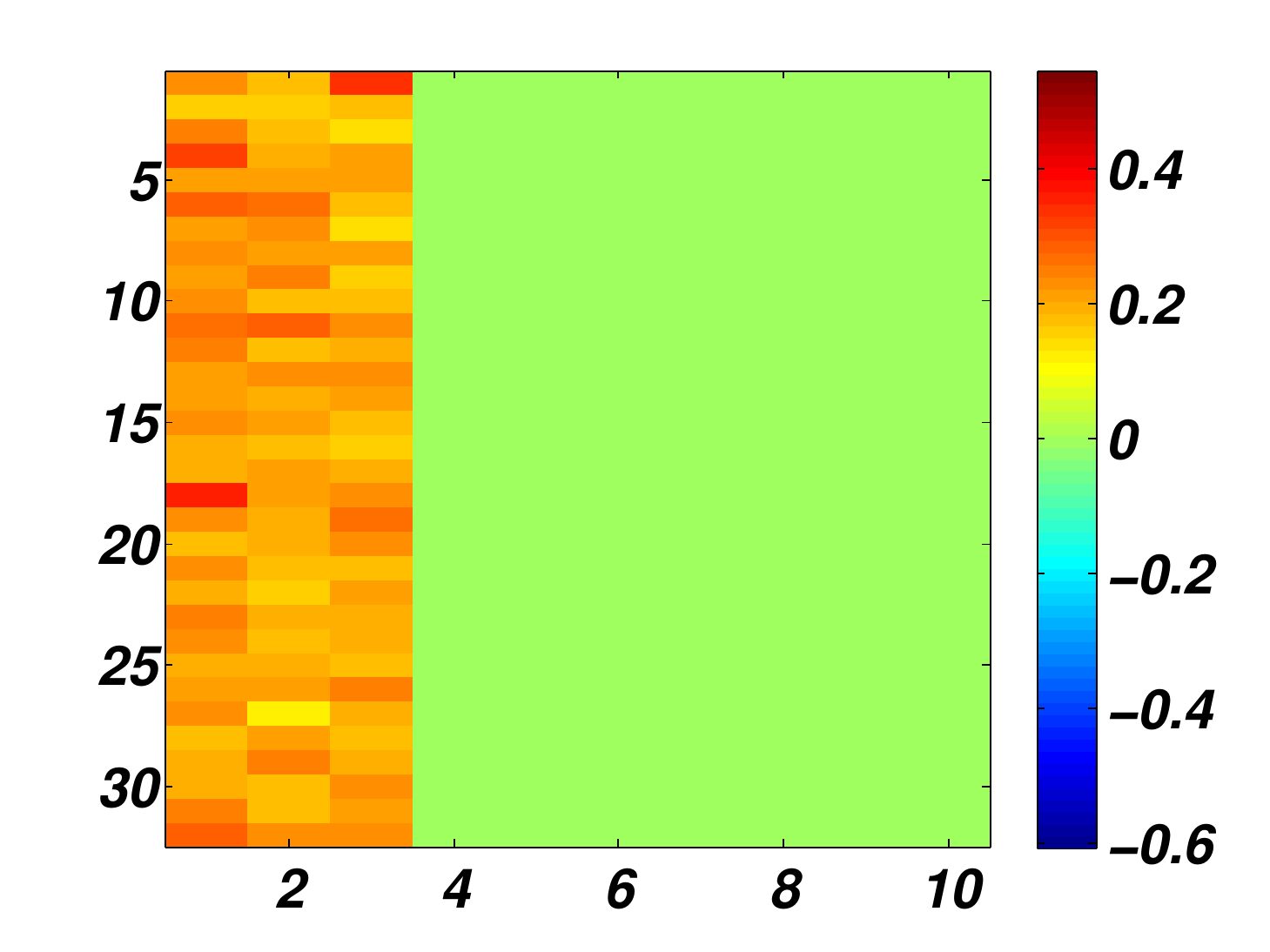}}
\subfigure[$\mathbf{c}(t)$]{\includegraphics[width = 0.24\textwidth]{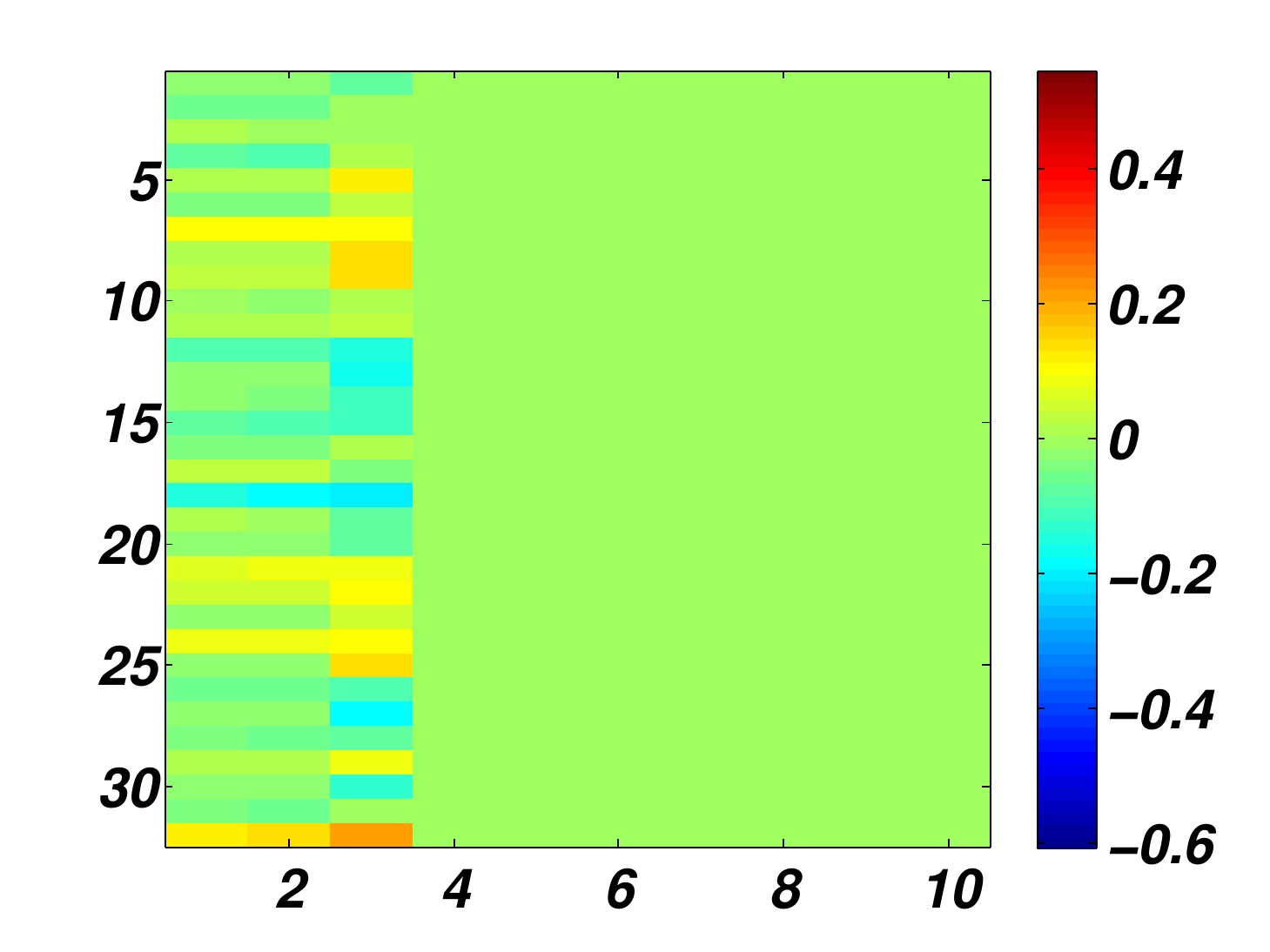}}
}
\centerline{
\subfigure[$\mathbf{o}(t)$]{\includegraphics[width = 0.24\textwidth]{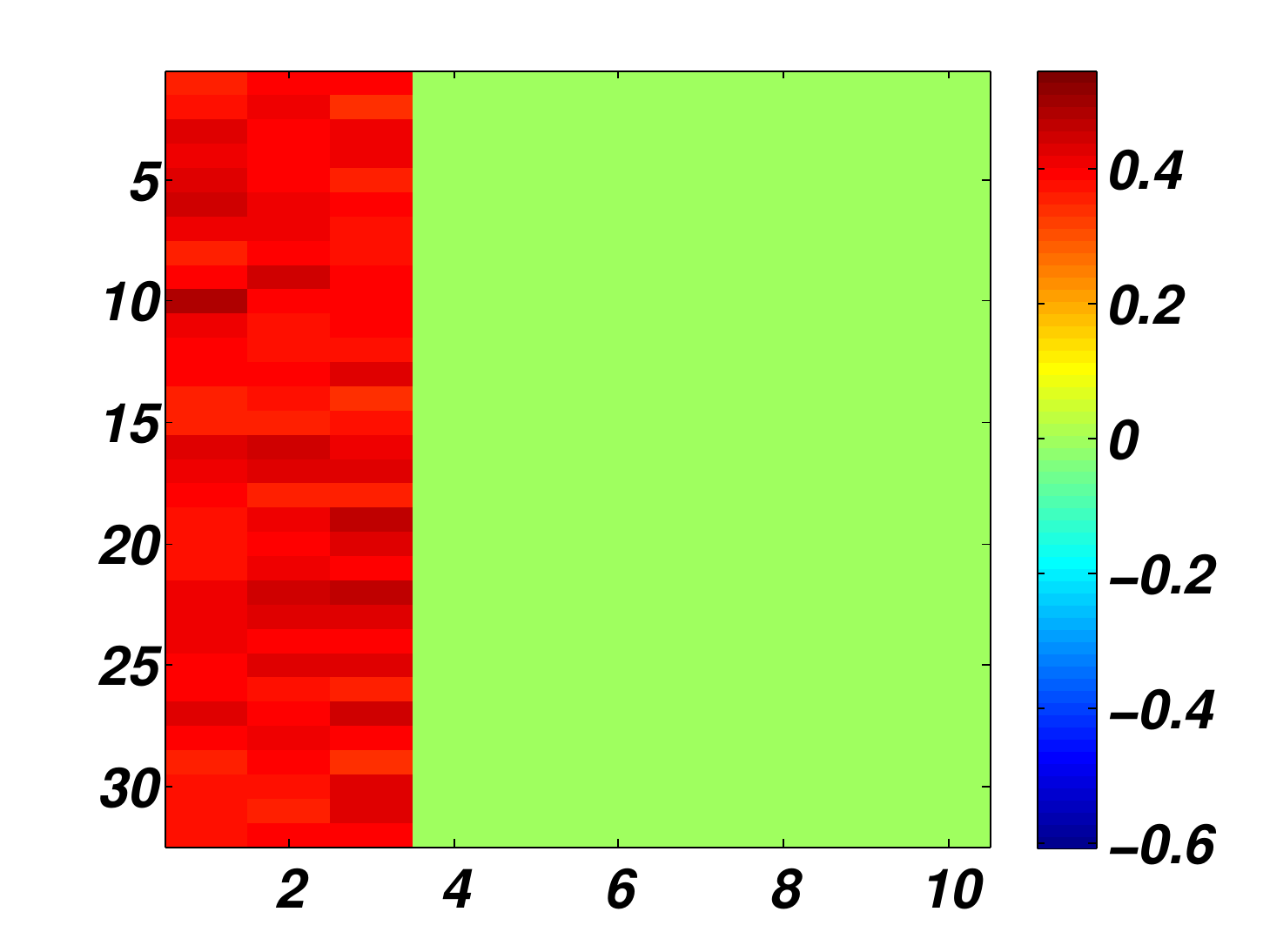}}
\subfigure[$\mathbf{y}(t)$]{\includegraphics[width = 0.24\textwidth]{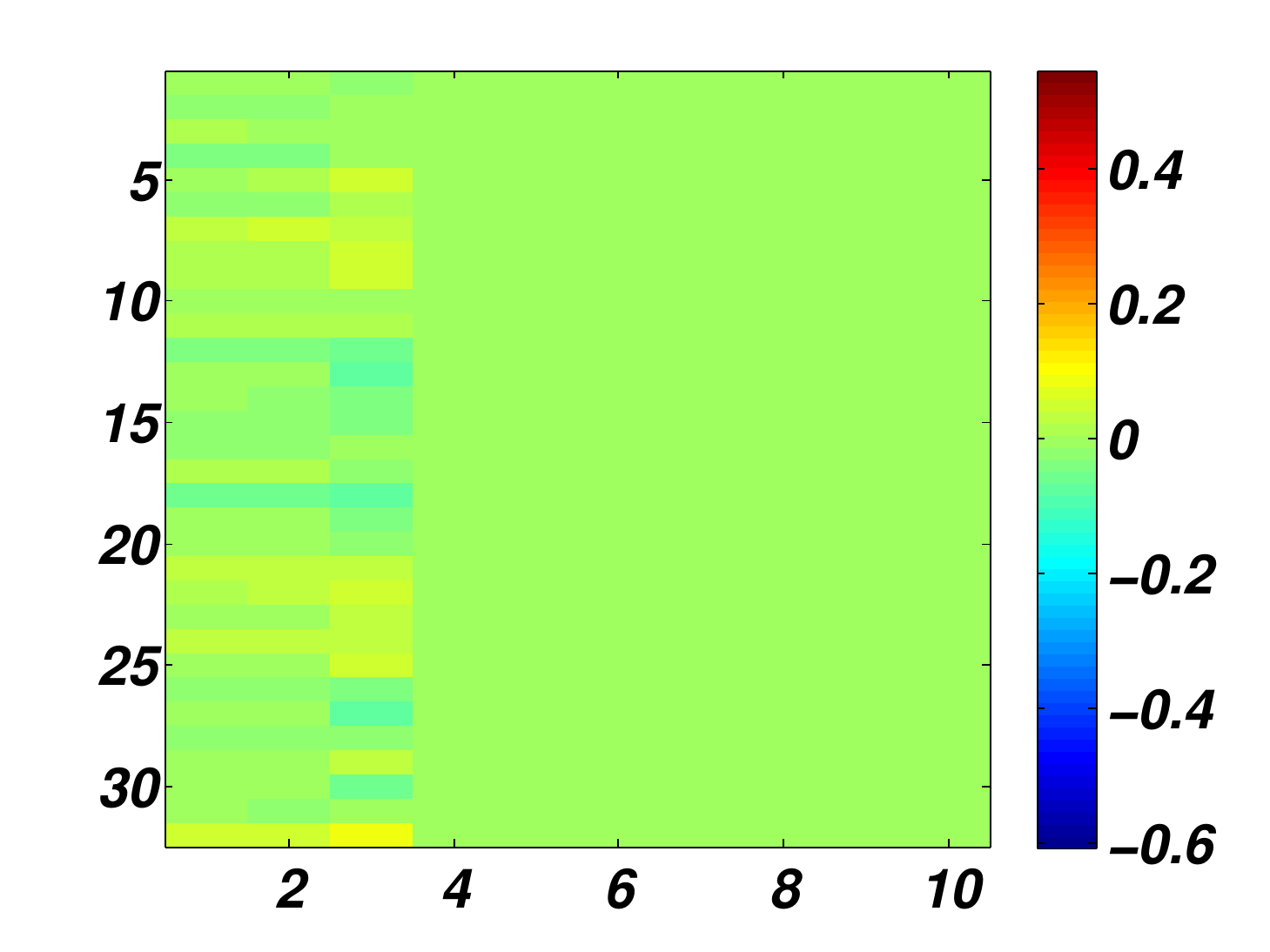}}
}
\caption{Query: ``$hotels\; in \; shanghai$''. Since the sentence ends at the third word, all the values to the right of it are zero (green color). }
\label{fig:ShanghaiQ}
\end{figure}

\begin{figure}[t]
\centerline{
\subfigure[$\mathbf{i}(t)$]{\includegraphics[width = 0.24\textwidth]{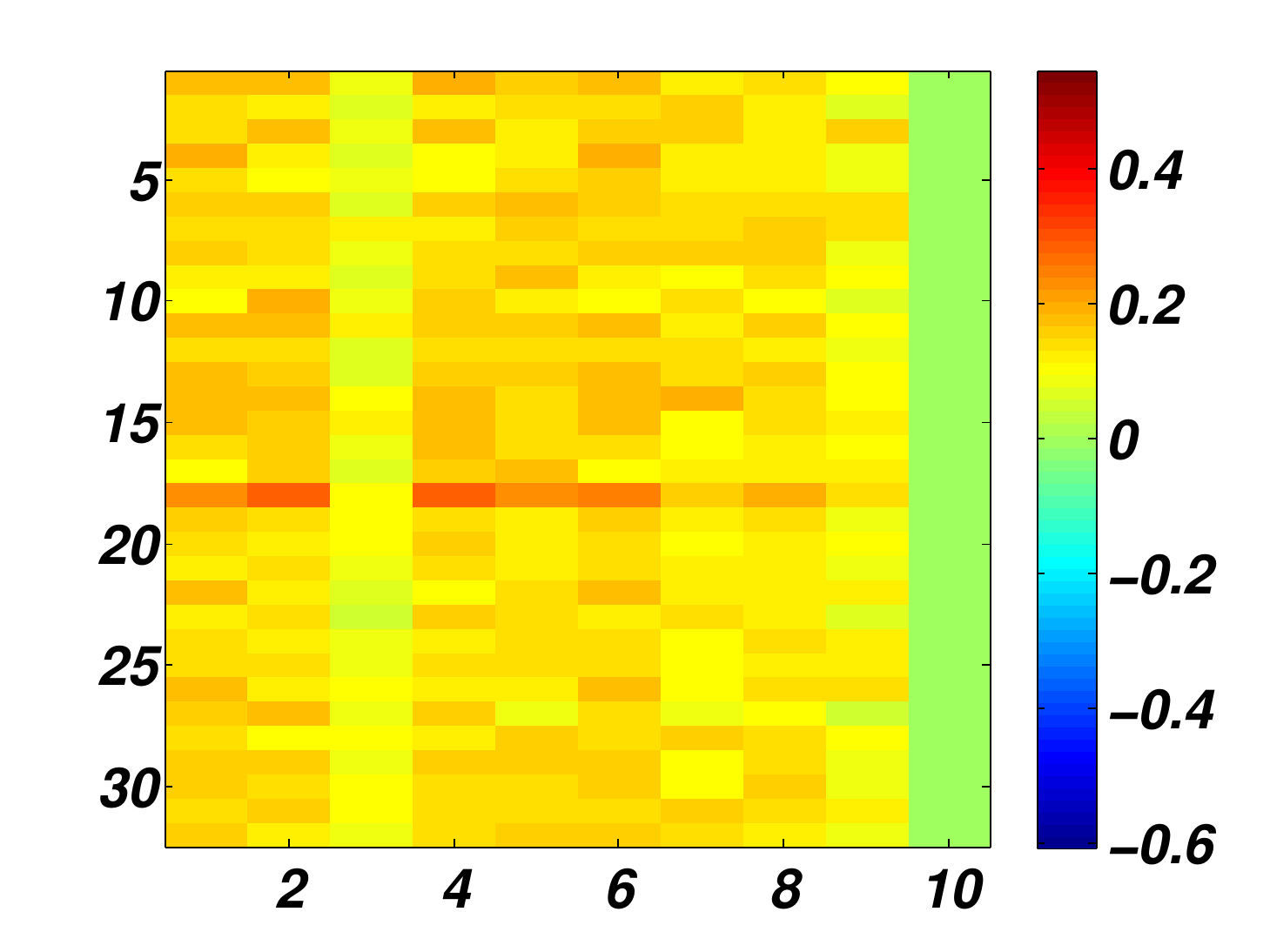}}
\subfigure[$\mathbf{c}(t)$]{\includegraphics[width = 0.24\textwidth]{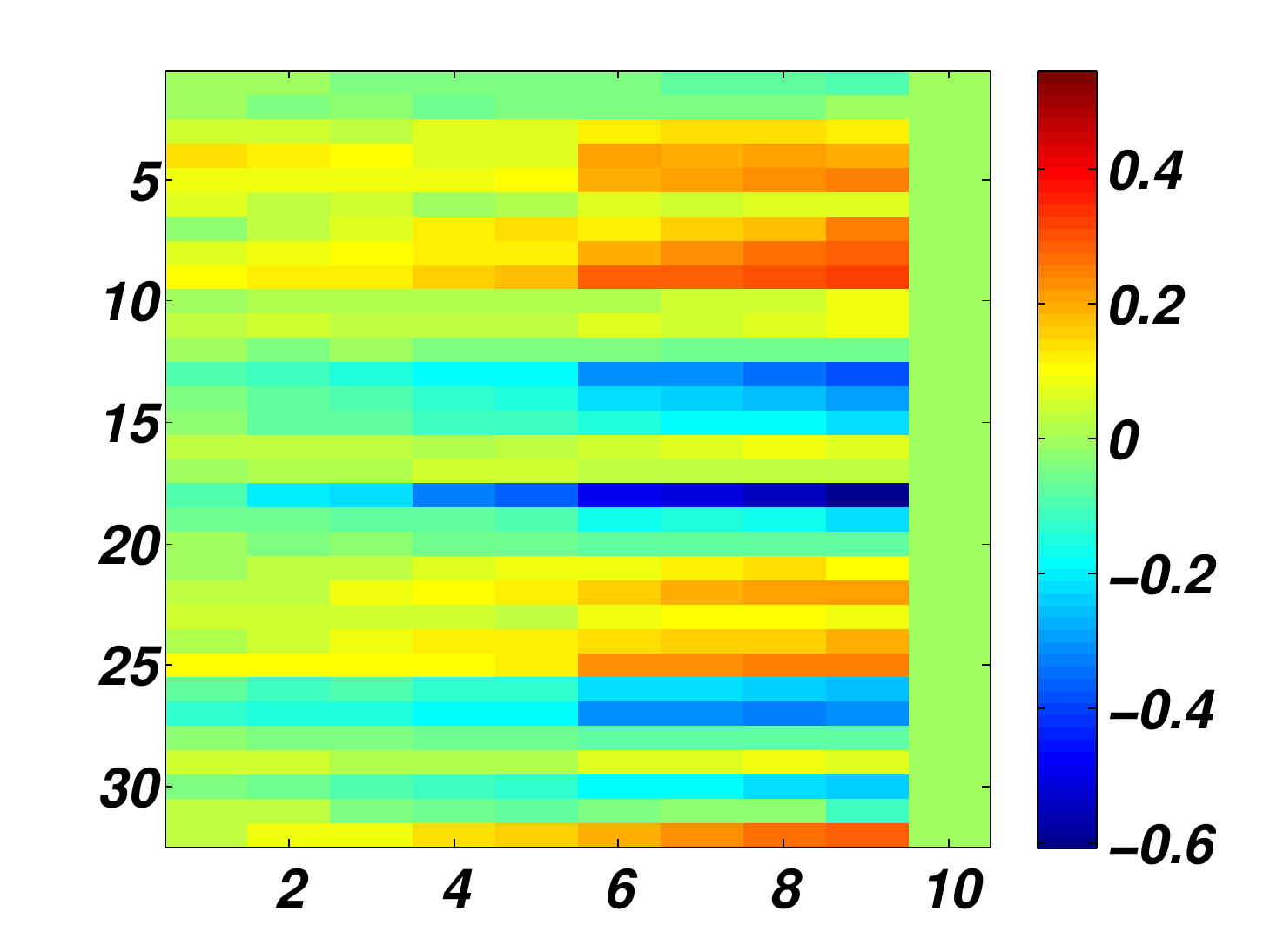}}
}
\centerline{
\subfigure[$\mathbf{o}(t)$]{\includegraphics[width = 0.24\textwidth]{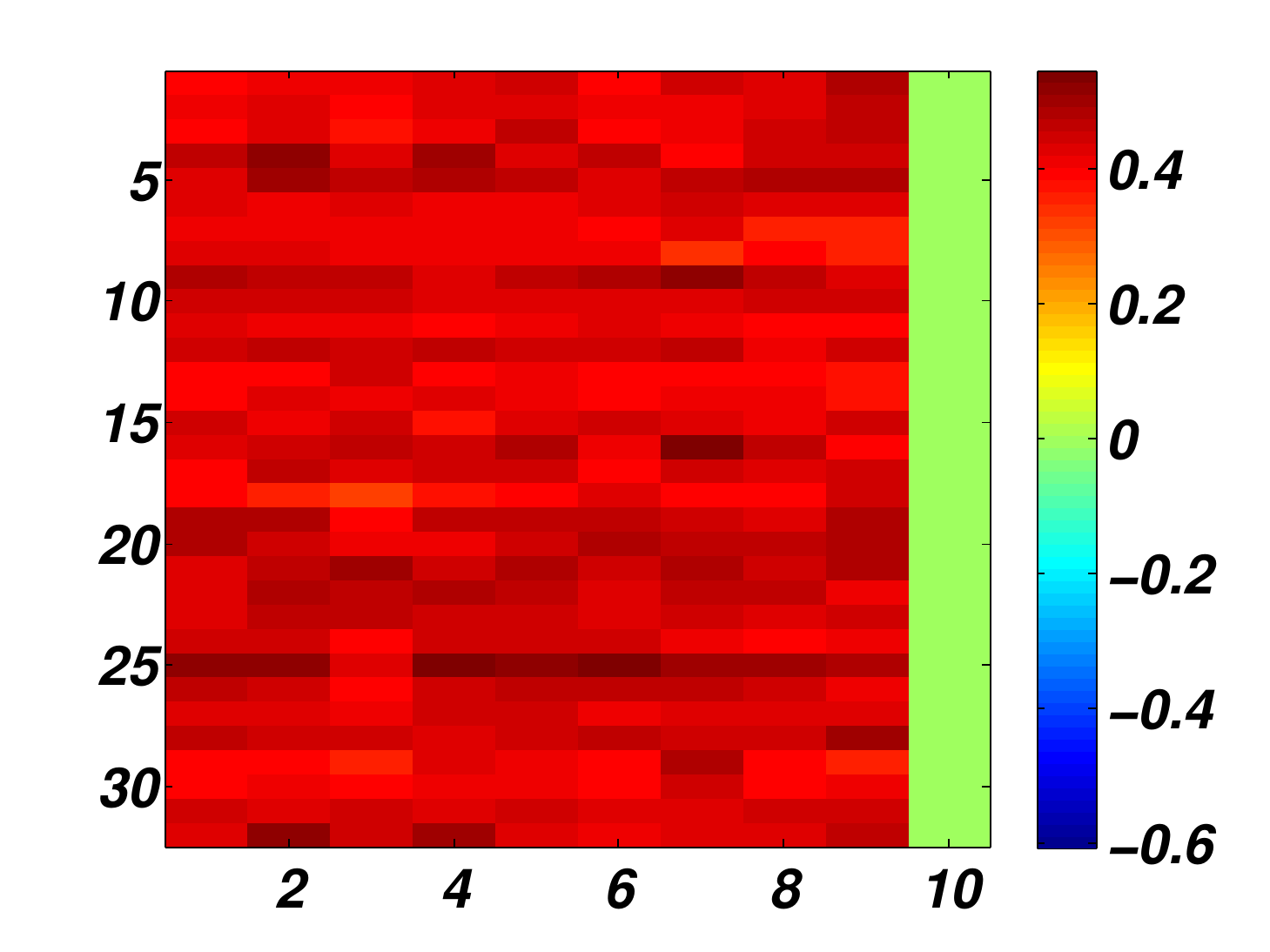}}
\subfigure[$\mathbf{y}(t)$]{\includegraphics[width = 0.24\textwidth]{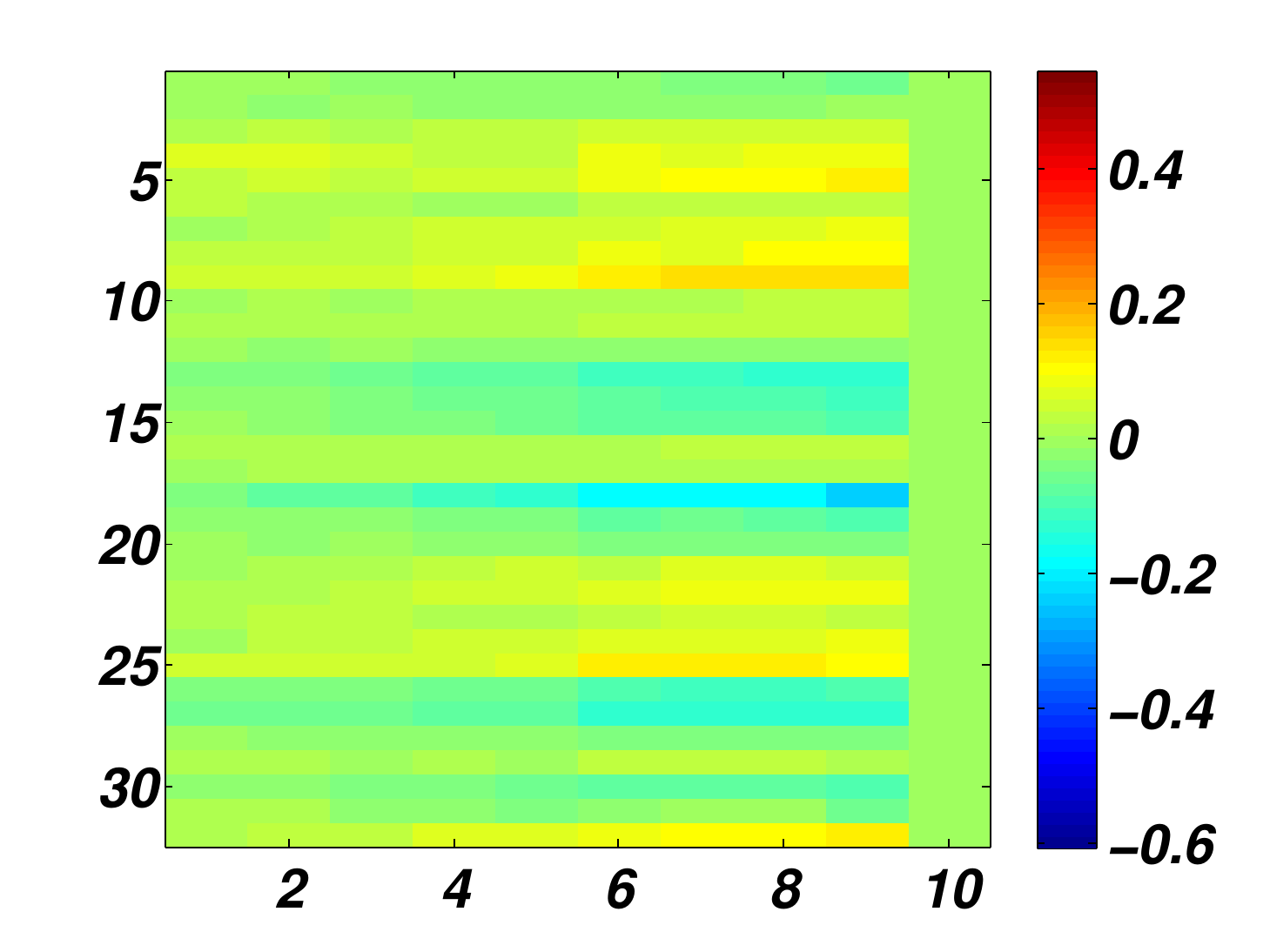}}
}
\caption{Document: ``$shanghai\; hotels\; accommodation\; hotel$ \newline $in\; shanghai\; discount\; and\; reservation$''. Since the sentence ends at the ninth word, all the values to the right of it are zero (green color).}
\label{fig:ShanghaiD}
\end{figure}

First, we examine the evolution of the semantic vector and how unimportant words are attenuated. Specifically, we feed the following input sentences from the test dataset into the trained LSTM-RNN model:
\begin{itemize}
\item Query: ``\emph{hotels  in   shanghai}''
\item Document: ``\emph{shanghai  hotels  accommodation  hotel in  shanghai  discount  and  reservation}''
\end{itemize}
Activations of input gate, output gate, cell state and the embedding vector for each cell for query and document are shown in Fig. \ref{fig:ShanghaiQ} and Fig. \ref{fig:ShanghaiD}, respectively. The vertical axis is the cell index from $1$ to $32$, and the horizontal axis is the word index from $1$ to $10$ numbered from left to right in a sequence of words and color codes show activation values.
From Figs.\ref{fig:ShanghaiQ}--\ref{fig:ShanghaiD}, we make the following observations:
\begin{itemize}

\item Semantic representation $\mathbf{y}(t)$ and cell states $\mathbf{c}(t)$ are evolving over time. Valuable context information is gradually absorbed into $\mathbf{c}(t)$ and $\mathbf{y}(t)$, so that the information in these two vectors becomes richer over time, and  the semantic information of the entire input sentence is embedded into vector $\mathbf{y}(t)$, which is obtained by applying output gates to the cell states $\mathbf{c}(t)$. 

\item The input gates evolve in such a way that it attenuates the unimportant information and detects the important information from the input sentence. For example, in Fig. \ref{fig:ShanghaiD}(a), most of the input gate values corresponding to word $3$, word $7$ and word $9$ have very small values (light green-yellow color)\footnote{If this is not clearly visible, please refer to Fig. 1 in section I of supplementary materials. We have adjusted color bar for all figures to have the same range, for this reason the structure might not be clearly visible. More visualization examples could also be found in section IV of Supplementary Materials}, which corresponds to the words ``$accommodation$'', ``$discount$'' and ``$reservation$'', respectively, in the document sentence. Interestingly, input gates reduce the effect of these three words in the final semantic representation, $\mathbf{y}(t)$, such that the semantic similarity between sentences from query and document sides are not affected by these words.

\end{itemize}

\subsubsection{Keywords Extraction}
\label{sec:exKeyWord}
\begin{figure}[t]
\centerline{
\subfigure[$\mathbf{y}(t)$ top 10 for query]{\includegraphics[width = 0.24\textwidth]{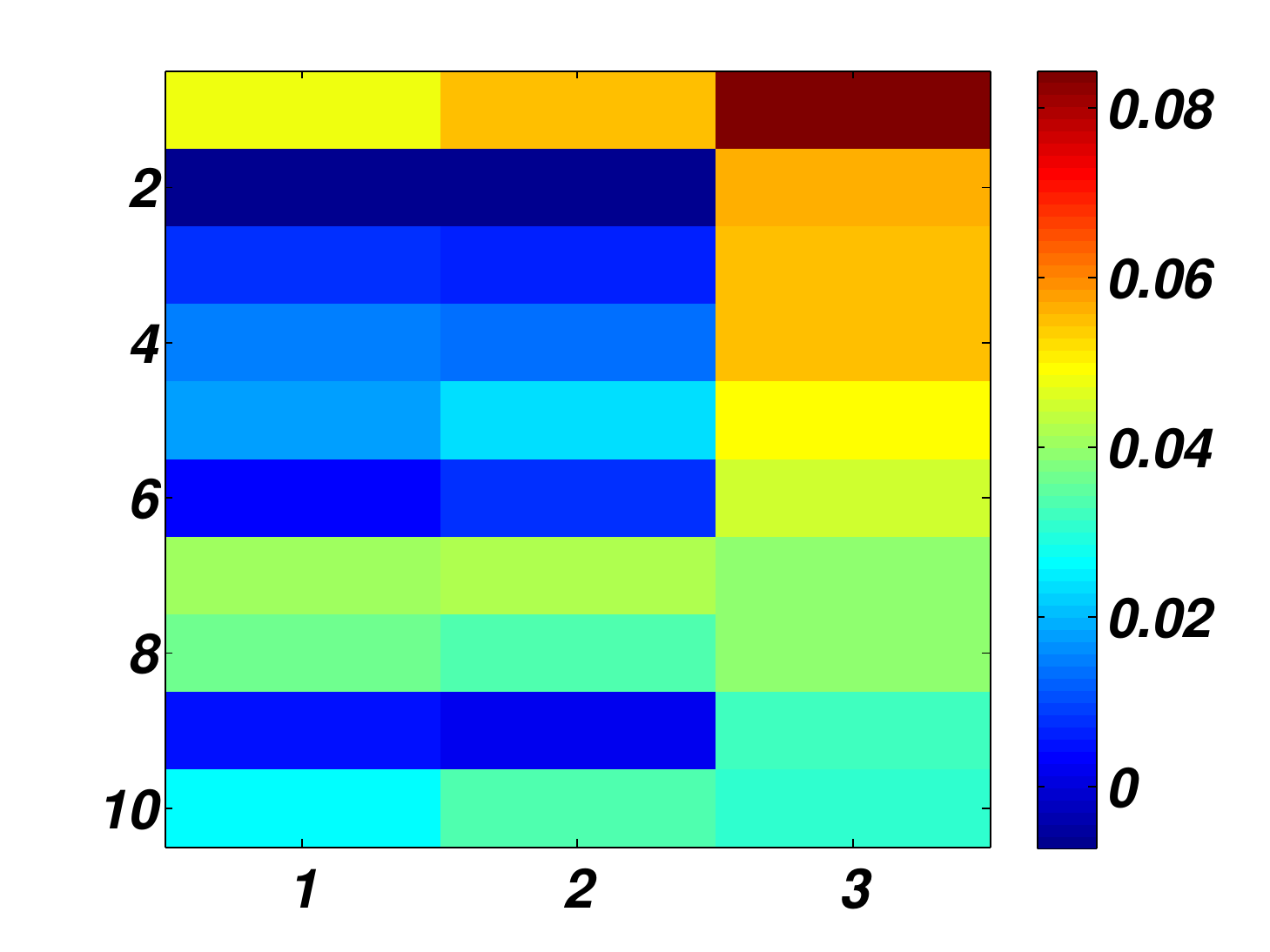}}
\subfigure[$\mathbf{y}(t)$ top 10 for document]{\includegraphics[width = 0.24\textwidth]{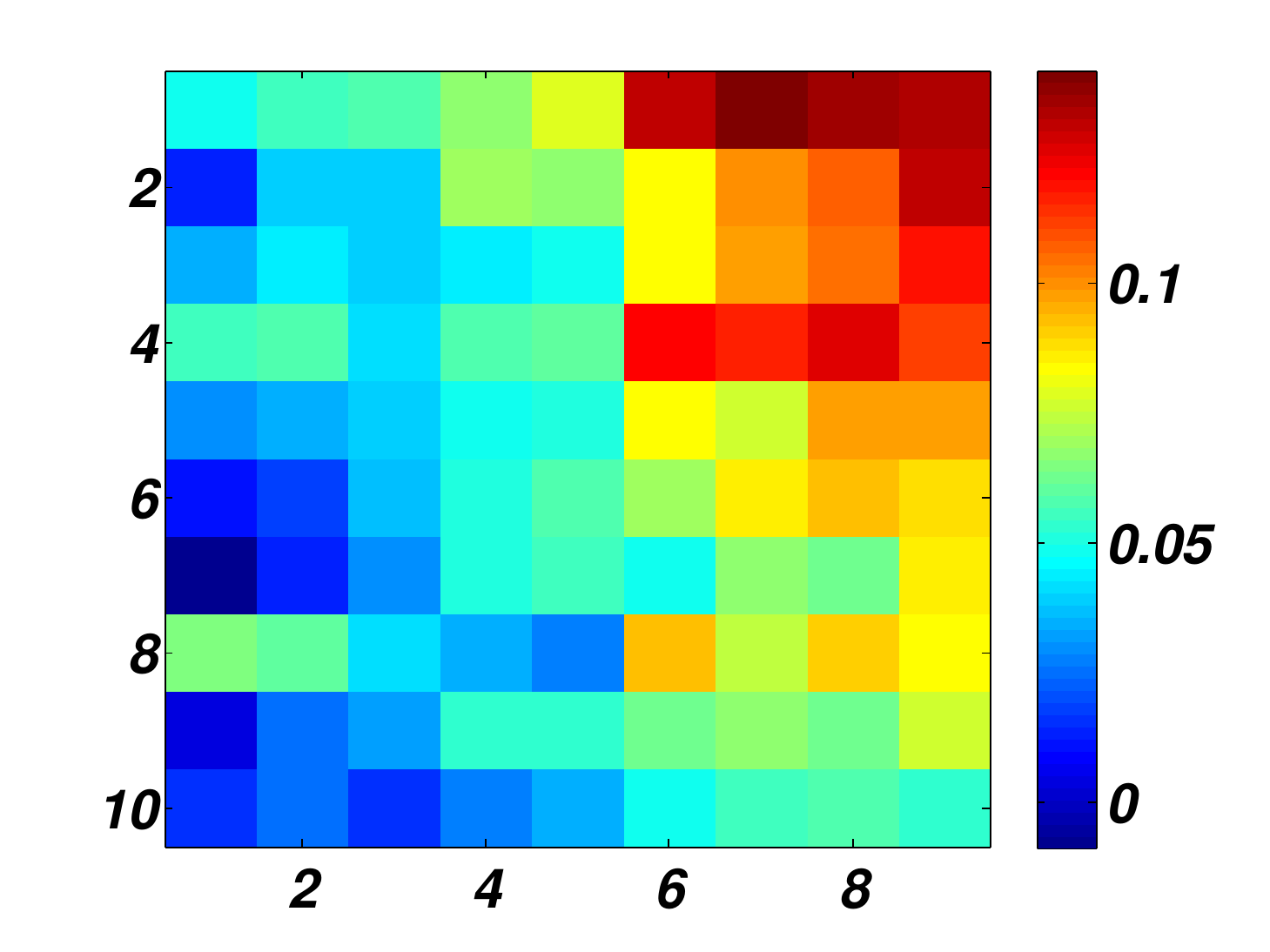}}
}
\caption{Activation values, $\mathbf{y}(t)$, of 10 most active cells for Query: ``\emph{hotels in  shanghai}'' and Document: ``\emph{shanghai hotels accommodation hotel in shanghai discount and reservation}''}
\label{fig:KeyWordShanghaiQD}
\end{figure}

In this section, we show how the trained LSTM-RNN extracts the important information, i.e., keywords, from the input sentences. To this end, we backtrack semantic representations, $\mathbf{y}(t)$, over time. We focus on the 10 most active cells in final semantic representation. Whenever there is a large enough change in cell activation value ($\mathbf{y}(t)$), we assume an important keyword has been detected by the model. We illustrate the result using the above example (``\emph{hotels  in   shanghai}''). The evolution of the 10 most active cells activation, $\mathbf{y}(t)$, over time are shown in Fig. \ref{fig:KeyWordShanghaiQD} for the query and the document sentences.\footnote{Likewise, the vertical axis is the cell index and horizontal axis is the word index in the sentence.}%
From Fig. \ref{fig:KeyWordShanghaiQD}, we also observe that different words activate different cells. In Tables \ref{table:Q1}--\ref{table:D1}, we show the number of cells each word activates.\footnote{Note that before presenting the first word of the sequence,  activation values are initially zero so that there is always a considerable change in the cell states after presenting the first word. For this reason, we have not indicated the number of cells detecting the first word as a keyword. Moreover, another keyword extraction example can be found in section IV of supplementary materials.
}
We used Bidirectional LSTM-RNN to get the results of these tables where in the first row, LSTM-RNN reads sentences from left to right and in the second row it reads sentences from right to left. In these tables we labelled a word as a keyword if more than $40\%$ of top $10$ active cells in both directions declare it as keyword. The boldface numbers in the table show that the number of cells assigned to that word is more than $4$, i.e., $40\%$ of top $10$ active cells.  
From the tables, we observe that the keywords activate more cells than the unimportant words, meaning that they are selectively embedded into the semantic vector.
\begin{table}[t]
\caption{Key words for query: ``$hotels\; in \; shanghai$''}
\label{table:Q1}
\vspace{-.5cm}
\begin{center}
\scriptsize
\begin{tabular}{ | c | c | c |c | } 
\hline
     Query & \color{red}{$hotels$} & $in$ & \color{red}{$shanghai$}\\ 
\hline
Number of assigned &  &  & \\
cells out of 10 &  &  & \\
Left to Right & - & 0 & \textbf{7} \\ 
\hline
Number of assigned &  &  & \\
cells out of 10 &  &  & \\
Right to Left & \textbf{6} & 0 & - \\ 
\hline
\end{tabular}
\end{center}
\end{table}
\begin{table*}[ht]
\caption{Key words for document: ``$shanghai\; hotels\; accommodation\; hotel\; in\; shanghai\; discount\; and\; reservation$''}
\label{table:D1}
\vspace{-.5cm}
\begin{center}
\scriptsize
\begin{tabular}{ | c | c | c |c |c| c | c | c |c |c| }
\hline
     & \color{red}{$shanghai$} & \color{red}{$hotels$} & $accommodation$ & \color{red}{$hotel$} & $in$ & $shanghai$ & \color{red}{$discount$} & $and$ & \color{red}{$reservation$}\\ 
\hline
Number of assigned &  &  &  &  &  &  &  &  & \\
cells out of 10 &  &  &  &  &  &  &  &  & \\
Left to Right & - & \textbf{4} & 3 & \textbf{8} & 1 & \textbf{8} & \textbf{5} & 3 & \textbf{4} \\ 
\hline
Number of assigned &  &  &  &  &  &  &  &  & \\
cells out of 10 &  &  &  &  &  &  &  &  & \\
Right to Left & \textbf{4} & \textbf{6} & \textbf{5} & \textbf{4} & \textbf{5} & 1 & \textbf{7} & \textbf{5} & - \\ 
\hline
\end{tabular}
\end{center}
\end{table*}
 
\subsubsection{Topic Allocation}
\label{sec:exTopicTransition}
Now, we further show that the trained LSTM-RNN model not only detects the keywords, but also allocates them properly to different cells according to the topics they belong to. To do this, we go through the test dataset using the trained LSTM-RNN model and search for the keywords that are detected by a specific cell. For simplicity, we use the following simple approach: for each given query we look into the keywords that are extracted by the 5 most active cells of LSTM-RNN and list them in Table \ref{tab:TopicModel}. Interestingly, each cell collects keywords of a specific topic. For example, cell 26 in Table \ref{tab:TopicModel} extracts keywords related to the topic ``food'' and cells 2 and 6 mainly focus on the keywords related to the topic ``health''.
\begin{table*}[t]
\Huge
\label{tab:TopicModel}
    \caption{Keywords assigned to each cell of LSTM-RNN for different queries of two topics, ``\textcolor{green}{food}'' and ``\textcolor{red}{health}''} 
    \vspace{-.2cm}    
  \resizebox{\linewidth}{!}{
  \centering
    \label{tab:TopicModel}    
    \begin{tabular}{*{17}{|c}|}
    \hline
    Query & cell 1 & cell 2 & cell 3 & cell 4 & cell 5 & cell 6 & cell 7 & cell 8 & cell 9 & cell 10 & cell 11 & cell 12 & cell 13 & cell 14 & cell 15 & cell 16 \\\hline 
    \textcolor{green}{al yo yo sauce} &       &       &       &       & \textcolor{green}{yo}    &       &       & \textcolor{green}{sauce} &       &       & \textcolor{green}{sauce} &       &       & &  &\\\hline
    \textcolor{green}{atkins diet lasagna} &       &       &       &       &       &       &       & \textcolor{green}{diet}  &       &       &       &       &       &  & &\\\hline
    \textcolor{green}{blender recipes} &       &       &       &       &       &       &       &       &       &       &       &       &       &  & &\\\hline
    \textcolor{green}{cake bakery edinburgh} &       &       &       &       &       &       &       &       &       & \textcolor{green}{bakery} &       &       &       &  & &\\\hline
    \textcolor{green}{canning corn beef hash} &       &       &       &       & \textcolor{green}{beef, hash} &       &       &       &       &       &       &       &       &  & &\\\hline
    \textcolor{green}{torre de pizza} &       &       &       &       &       &       &       &       &       &       &       &       &       &  & &\\\hline
    \textcolor{green}{famous desserts} &       &       &       &       &       &       & \textcolor{green}{desserts} &       &       &       &       &       &       &  & &\\\hline
    \textcolor{green}{fried chicken} &       &       & \textcolor{green}{chicken} &       &       &       & \textcolor{green}{chicken} &       &       &       &       &       &       &  & &\\\hline
    \textcolor{green}{smoked turkey recipes} &       &       &       &       &       &       &       &       &       &       &       &       &       &  & &\\\hline
    \textcolor{green}{italian sausage hoagies} &       &       &       &       &       &       &       & \textcolor{green}{sausage} &       &       &       &       &       &  & &\\\hline
    \textcolor{red}{do you get allergy} &       & \textcolor{red}{allergy} &       &       &       &       &       &       &       &       &       &       &       &  & &\\\hline
    \textcolor{red}{much pain will after total knee replacement} & \textcolor{red}{pain}  &       &       &       &       & \textcolor{red}{pain, knee} &       &       &       &       &       &       &       &  & &\\\hline
    \textcolor{red}{how to make whiter teeth} &       &       &       &       &       &       &       &       &       &       &       &       & \textcolor{red}{make, teeth} &  &\textcolor{red}{to} &\\\hline
    \textcolor{red}{illini community hospital} &       & \textcolor{red}{community, hospital} &       &       &       &       &       & \textcolor{red}{hospital} &       & \textcolor{red}{community} &       &       &       &  & &\\\hline
    \textcolor{red}{implant infection} &       & \textcolor{red}{infection} &       &       &       & \textcolor{red}{infection} &       &       &       &       &       &       &       &  & &\\\hline
    \textcolor{red}{introductory psychology} &       & \textcolor{red}{psychology} &       &       &       & \textcolor{red}{psychology} &       &       &       &       &       &       &       &  & &\\\hline
    \textcolor{red}{narcotics during pregnancy side effects} &       & \textcolor{red}{pregnancy} &       &       &       & \textcolor{red}{pregnancy,effects, during} &       &       &       &       &       &       & \textcolor{red}{during} &  & &\\\hline
    \textcolor{red}{fight sinus infections} &       &       &       &       &       & \textcolor{red}{infections} &       &       &       &       &       &       &       &  & &\\\hline
    \textcolor{red}{health insurance high blood pressure} &       & \textcolor{red}{insurance} &       &       &       & \textcolor{red}{blood} &       & \textcolor{red}{high, blood} &       &       &       &       &       &  & &\\\hline
    \textcolor{red}{all antidepressant medications} &       & \textcolor{red}{antidepressant, medications} &       &       &       &       &       &       &       &       &       &       &       &  & &\\\hline
    \end{tabular}%
}
\resizebox{.95\linewidth}{!}{
    \begin{tabular}{*{17}{|c}|}
    \hline
    Query & cell 17 & cell 18 & cell 19 & cell 20 & cell 21 & cell 22 & cell 23 & cell 24 & cell 25 & cell 26 & cell 27 & cell 28 & cell 29 & cell 30& cell 31 & cell 32 \\\hline
    \textcolor{green}{al yo yo sauce} &           &       &       &       &       &       &       &       &       &       &       &       &  &&&\\\hline
    \textcolor{green}{atkins diet lasagna} &           &       &       &       &       &       & \textcolor{green}{diet}  &       &       &       &       &       & &\textcolor{green}{diet}&& \\\hline
    \textcolor{green}{blender recipes} &           &       &       &       &       &       &       &       &       & \textcolor{green}{recipes} &       &       &  &&&\\\hline
    \textcolor{green}{cake bakery edinburgh} &              &       &       & \textcolor{green}{bakery} &       &       &       &       &       & \textcolor{green}{bakery} &       &       &  &&&\\\hline
    \textcolor{green}{canning corn beef hash} &              &       &       &       &       &       &       &       &       & \textcolor{green}{corn, beef} &       &       &  &&&\\\hline
    \textcolor{green}{torre de pizza} &             &       &       &       &       &       &       &       &       & \textcolor{green}{pizza} &       &       & \textcolor{green}{pizza} &&&\\\hline
    \textcolor{green}{famous desserts} &              &       &       &       &       &       &       &       &       &       &       &       &  &&&\\\hline
    \textcolor{green}{fried chicken} &              &       &       &       &       &       &       &       &       & \textcolor{green}{chicken} &       &       &  &&&\\\hline
    \textcolor{green}{smoked turkey recipes} &             &       &       & \textcolor{green}{turkey} &       &       &       &       &       & \textcolor{green}{recipes} &       &       &  &&&\\\hline
    \textcolor{green}{italian sausage hoagies} &        \textcolor{green}{hoagies} &       &       &       &       & \textcolor{green}{sausage} &       &       &       & \textcolor{green}{sausage} &       &       &  &&&\\\hline
   \textcolor{red}{ do you get allergy} &              &       &       &       &       &       &       &       &       &       &       &       &  &&&\\\hline
    \textcolor{red}{much pain will after total knee replacement} &        \textcolor{red}{knee}  &       &       &       &       &       & \textcolor{red}{replacement} &       &       &       &       &       &  &&&\\\hline
    \textcolor{red}{how to make whiter teeth} &        &       &       &       &       &       &       &       &       & \textcolor{red}{whiter} &       &       &  &&&\\\hline
    \textcolor{red}{illini community hospital} &              &       &       &       &       & \textcolor{red}{hospital} &       &       &       &       &       &       &  &\textcolor{red}{hospital}&&\\\hline
    \textcolor{red}{implant infection} &       &              &       &       &       &       &       &       & \textcolor{red}{infection} &       &       &       &  &&&\\\hline
    \textcolor{red}{introductory psychology} &              &       &       &       &       &       &       &       &       &       &       & \textcolor{red}{psychology} &  &&&\\\hline
    \textcolor{red}{narcotics during pregnancy side effects} &            &       &       &       &       &       &       &       &       &       &       &       &  &&&\\\hline
    \textcolor{red}{fight sinus infections} &        \textcolor{red}{sinus, infections} &       &       &       &       &       &       &       & \textcolor{red}{infections} &       &       &       &  &&&\\\hline
    \textcolor{red}{health insurance high blood pressure} &              &       &       &       &       &       & \textcolor{red}{high, pressure} &       &       &       &       &       &  &\textcolor{red}{insurance,high}&&\\\hline
    \textcolor{red}{all antidepressant medications} &       &              &       &       &       &       &       &       & \textcolor{red}{antidepressant} &       &       &       &  &\textcolor{red}{medications}&&\\\hline
    \end{tabular}%
}   
\end{table*}%

\subsection{Performance Evaluation}
\label{sec:performance}
\subsubsection{Web Document Retrieval Task}
\label{sec:exRanking}
In this section, we apply the proposed sentence embedding method to an important web document retrieval task for a commercial web search engine. Specifically, the RNN models (with and without LSTM cells) embed the sentences from the query and the document sides into their corresponding semantic vectors, and then compute the cosine similarity between these vectors to measure the semantic similarity between the query and candidate documents. 

Experimental results for this task are shown in Table \ref{table:Results} using the standard metric mean Normalized Discounted Cumulative Gain (NDCG) \cite{refNDCG} (the higher the better) for evaluating the ranking performance of the RNN and LSTM-RNN on a standalone human-rated test dataset. We also trained several strong baselines, such as DSSM \cite{DSSM} and CLSM \cite{CDSSM}, on the same training dataset and evaluated their performance on the same task. For fair comparison, our proposed RNN and LSTM-RNN models are trained with the same number of parameters as the DSSM and CLSM models ($14.4$M parameters). Besides, we also include in Table \ref{table:Results} two well-known information retrieval (IR) models, BM25 and PLSA, for the sake of benchmarking. The BM25 model uses the bag-of-words representation for queries and documents, which is a state-of-the-art document ranking model based on term matching, widely used as a baseline in IR society. PLSA (Probabilistic Latent Semantic Analysis) is a topic model proposed in \cite{ref19}, which is trained using the Maximum A Posterior estimation \cite{ref15} on the documents side from the same training dataset. We experimented with a varying number of topics from 100 to 500 for PLSA, which gives similar performance, and we report in Table \ref{table:Results} the results of using 500 topics. Results for a language model based method, uni-gram language model (ULM) with Dirichlet smoothing, are also presented in the table. 

To compare the performance of the proposed method with general sentence embedding methods in document retrieval task, we also performed experiments using two general sentence embedding methods. 

\begin{enumerate}
\item In the first experiment, we used the method proposed in \cite{le2014distributed} that generates embedding vectors known as Paragraph Vectors. It is also known as doc2vec. It maps each word to a vector and then uses the vectors representing all words inside a context window to predict the vector representation of the next word. The main idea in this method is to use an additional paragraph token from previous sentences in the document inside the context window. This paragraph token is mapped to vector space using a different matrix from the one used to map the words. A primary version of this method is known as word2vec proposed in \cite{word2vec}. The only difference is that word2vec does not include the paragraph token. 

To use doc2vec on our dataset, we first trained doc2vec model on both train set (about 200,000 query-document pairs) and test set (about 900,000 query-document pairs). This gives us an embedding vector for every query and document in the dataset. We used the following parameters for training:
\begin{itemize}
\item min-count=1 : minimum number of of words per sentence, sentences with words less than this will be ignored. We set it to 1 to make sure we do not throw away anything. 
\item window=5 : fixed window size explained in \cite{le2014distributed}. We used different window sizes, it resulted in about just 0.4\% difference in final NDCG values.
\item size=100 : feature vector dimension. We used 400 as well but did not get significantly different NDCG values.
\item sample=1e-4 : this is the down sampling ratio for the words that are repeated a lot in corpus.
\item negative=5 : the number of noise words, i.e., words used for negative sampling as explained in \cite{le2014distributed}.
\item We used 30 epochs of training. We ran an experiment with 100 epochs but did not observe much difference in the results.
\item We used \textit{gensim} \cite{gensim} to perform experiments. 
\end{itemize}

To make sure that a meaningful model is trained, we used the trained doc2vec model to find the most similar words to two sample words in our dataset, e.g., the words ``pizza'' and ``infection''. The resulting words and corresponding scores are presented in section V of Supplementary Materials. As it is observed from the resulting words, the trained model is a meaningful model and can recognise semantic similarity. 

Doc2vec also assigns an embedding vector for each query and document in our test set. We used these embedding vectors to calculate the cosine similarity score between each query-document pair in the test set. We used these scores to calculate NDCG values reported in Table \ref{table:Results} for the Doc2Vec model.

Comparing the results of doc2vec model with our proposed method for document retrieval task shows that the proposed method in this paper significantly outperforms doc2vec. One reason for this is that we have used a very general sentence embedding method, doc2vec, for document retrieval task. This experiment shows that it is not a good idea to use a general sentence embedding method and using a better task oriented cost function, like the one proposed in this paper, is necessary.

\item In the second experiment, we used the Skip-Thought vectors proposed in \cite{Skip-Thought}. During training, skip-thought method gets a tuple ($s(t-1),s(t),s(t+1)$) where it encodes the sentence $s(t)$ using one encoder, and tries to reconstruct the previous and next sentences, i.e., $s(t-1)$ and $s(t+1)$, using two separate decoders. The model uses RNNs with Gated Recurrent Unit (GRU) which is shown to perform as good as LSTM. In the paper, authors have emphasized that: "\textit{Our model depends on having a training corpus of contiguous text}". Therefore, training it on our training set where we barely have more than one sentence in query or document title is not fair. However, since their model is trained on 11,038 books from BookCorpus dataset \cite{BookCorpus} which includes about 74 million sentences, we can use the trained model as an off-the-shelf sentence embedding method as authors have concluded in the conclusion of the paper. 

To do this we downloaded their trained models and word embeddings (its size was more than 2GB) available from ``\url{https://github.com/ryankiros/skip-thoughts}''. Then we encoded each query and its corresponding document title in our test set as vector.

We used the combine-skip sentence embedding method, a vector of size $4800 \times 1$, where it is concatenation of a uni-skip, i.e., a unidirectional encoder resulting in a $2400 \times 1$ vector, and a bi-skip, i.e., a bidirectional encoder resulting in a $1200 \times 1$ vector by forward encoder and another $1200 \times 1$ vector by backward encoder. The authors have reported their best results with the combine-skip encoder.

Using the $4800 \times 1$ embedding vectors for each query and document we calculated the scores and NDCG for the whole test set which are reported in Table \ref{table:Results}.

The proposed method in this paper is performing significantly better than the off-the-shelf skip-thought method for document retrieval task. Nevertheless, since we used skip-thought as an off-the-shelf sentence embedding method, its result is good. This result also confirms that learning embedding vectors using a model and cost function specifically designed for document retrieval task is necessary.

\end{enumerate}

As shown in Table \ref{table:Results}, the LSTM-RNN significantly outperforms all these models, and exceeds the best baseline model (CLSM) by $1.3\%$ in NDCG@1 score, which is a statistically significant improvement. As we pointed out in Sec. \ref{sec:Analysis}, such an improvement comes from the LSTM-RNN's ability to embed the contextual and semantic information of the sentences into a finite dimension vector. In Table \ref{table:Results}, we have also presented the results when different number of negative samples, $n$, is used. Generally, by increasing $n$ we expect the performance to improve. This is because more negative samples results in a more accurate approximation of the partition function in \eqref{eq:L_r}. The results of using Bidirectional LSTM-RNN are also presented in Table \ref{table:Results}. In this model, one LSTM-RNN reads queries and documents from left to right, and the other LSTM-RNN reads queries and documents from right to left. Then the embedding vectors from left to right and right to left LSTM-RNNs are concatenated to compute the cosine similarity score and NDCG values.

\begin{table}[t]
\caption{Comparisons of NDCG performance measures (the higher the better) of proposed models and a series of baseline models, where {\it nhid} refers to the number of hidden units, {\it ncell} refers to number of cells, {\it win} refers to window size, and {\it n} is the number of negative samples which is set to 4 unless otherwise stated. Unless stated otherwise, the RNN and LSTM-RNN models are chosen to have the same number of model parameters as the DSSM and CLSM models: $14.4$M, where $1 \mathrm{M} = 10^6$. The boldface numbers are the best results.}
\label{table:Results}
\vspace{-.5cm}
\begin{center}
\begin{tabular}{ | c | c | c | c| } 
\hline
Model & NDCG & NDCG & NDCG\\ 
 & @1 & @3 & @10\\ \hline
Skip-Thought & 26.9\% & 29.7\% & 36.2\% \\ 
off-the-shelf & & & \\
\hline
Doc2Vec & 29.1\% & 31.8\% & 38.4\% \\ \hline
ULM & 30.4\% & 32.7\% & 38.5\% \\ \hline
BM25 & 30.5\% & 32.8\% & 38.8\% \\ \hline
PLSA (T=500) & 30.8\% & 33.7\% & 40.2\% \\ \hline
DSSM (nhid = 288/96)  & 31.0\% & 34.4\% & 41.7\% \\
2 Layers & & & \\
\hline
CLSM (nhid = 288/96, win=1)  & 31.8\% & 35.1\% & 42.6\% \\
2 Layers, 14.4 M parameters & & & \\
\hline
CLSM (nhid = 288/96, win=3)  & 32.1\% & 35.2\% & 42.7\% \\
2 Layers, 43.2 M parameters & & & \\
\hline
CLSM (nhid = 288/96, win=5)  & 32.0\% & 35.2\% & 42.6\% \\
2 Layers, 72 M parameters & & & \\
\hline
RNN (nhid = 288)  & 31.7\% & 35.0\% & 42.3\% \\
1 Layer & & & \\
\hline
LSTM-RNN (ncell = 32)  & 31.9\% & 35.5\% & 42.7\% \\
1 Layer, 4.8 M parameters  &  &  &  \\ \hline
LSTM-RNN (ncell = 64)  & 32.9\% & 36.3\% & 43.4\% \\
1 Layer, 9.6 M parameters &  &  &  \\ \hline
LSTM-RNN (ncell = 96) & { 32.6\%} & { 36.0\%} & { 43.4\%} \\
1 Layer, n = 2 & & & \\
\hline
LSTM-RNN (ncell = 96) & {33.1\%} & {36.5\%} & {43.6\%} \\
1 Layer, n = 4 & & & \\
\hline
LSTM-RNN (ncell = 96) & {33.1\%} & {36.6\%} & {43.6\%} \\
1 Layer, n = 6 & & & \\
\hline
LSTM-RNN (ncell = 96) & {\bf 33.1\%} & {\bf 36.4\%} & {\bf 43.7\%} \\
1 Layer, n = 8 & & & \\
\hline
Bidirectional LSTM-RNN & {\bf 33.2\%} & {\bf 36.6\%} & {\bf 43.6\%} \\
(ncell = 96), 1 Layer & & & \\
\hline
\end{tabular}
\end{center}
\end{table}

\begin{figure}[t]
\center
\includegraphics[width=0.48\textwidth]{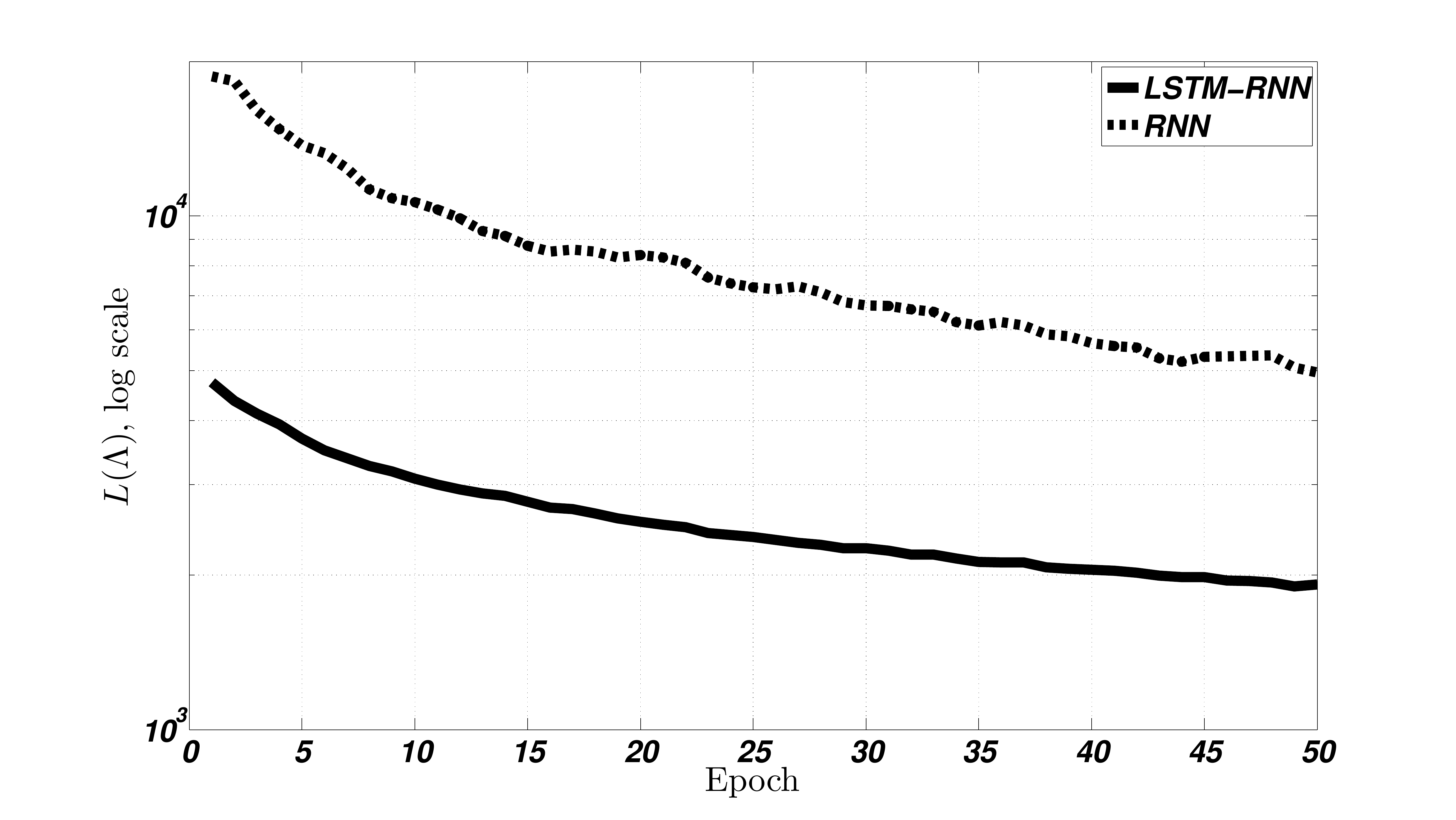}
\vspace{-.5cm}
\caption{LSTM-RNN compared to RNN during training: The vertical axis is logarithmic scale of the training cost, $L(\mathbf{\Lambda})$, in \eqref{eq:objective_func}. Horizontal axis is the number of epochs during training.}
   \label{fig:LSTMvsRDSSMcurve}
\end{figure}

A comparison between the value of the cost function during training for LSTM-RNN and RNN on the click-through data is shown in Fig. \ref{fig:LSTMvsRDSSMcurve}. From this figure, we conclude that LSTM-RNN is optimizing the cost function in \eqref{eq:objective_func} more effectively. Please note that all parameters of both models are initialized randomly.

\section{Conclusions and Future Work}
\label{sec:conclusion}

This paper addresses deep sentence embedding. We propose a model based on long short-term memory to model the long range context information and embed the key information of a sentence in one semantic vector. We show that the semantic vector evolves over time and only takes useful information from any new input. This has been made possible by input gates that detect useless information and attenuate it. Due to general limitation of available human labelled data, we proposed and implemented training the model with a \textit{weak supervision} signal using user click-through data of a commercial web search engine. 

By performing a detailed analysis on the model, we showed that: 1) The proposed model is robust to noise, i.e., it mainly embeds keywords in the final semantic vector representing the whole sentence and 2) In the proposed model, each cell is usually allocated to keywords from a specific topic. These findings have been supported using extensive examples. As a concrete sample application of the proposed sentence embedding method, we evaluated it on the important language processing task of web document retrieval. We showed that, for this task, the proposed method outperforms all existing state of the art methods significantly.

This work has been motivated by the earlier successes of deep learning methods in speech \cite{dahl2011large,Deng1,Dahl,hinton2012deep,DSN} and in semantic modelling \cite{DSSM,CDSSM,Interestingness,ACL}, and it adds further evidence for the effectiveness of these methods.
Our future work will further extend the methods to include 1) Using the proposed sentence embedding method for other important language processing tasks for which we believe sentence embedding plays a key role, e.g., the question / answering task. 2) Exploit the prior information about the structure of the different matrices in Fig. \ref{fig:LSTM Architecture} to develop a more effective cost function and learning method. 3) Exploiting attention mechanism in the proposed model to improve the performance and find out which words in the query are aligned to which words of the document.

\appendices
\section{Expressions for the Gradients}
\label{sec:appGradientExpress}
In this appendix we present the final gradient expressions that are necessary to use for training the proposed models. Full derivations of these gradients are presented in section III of supplementary materials.
\subsection{RNN}
\label{sec:app:rdssm}
For the recurrent parameters, $\mathbf{\Lambda} = \mathbf{W}_{rec}$ (we have ommitted $r$ subscript for simplicity):
\begin{align}
\label{eq:rdssmTrain3}
&\frac{\partial\Delta_{j,\tau}}{\partial\mathbf{W}_{rec}} = [\mathbf{\delta}_{y_Q}^{D^+}(t-\tau)\mathbf{y}_{Q}^T(t-\tau-1) + \nn\\
&\mathbf{\delta}_{y_D}^{D^+}(t-\tau)\mathbf{y}_{D^+}^T(t-\tau-1)]-[\mathbf{\delta}_{y_Q}^{D_j^-}(t-\tau)\mathbf{y}_{Q}^T(t-\tau-1)\nn\\
&+\mathbf{\delta}_{y_D}^{D_j^-}(t-\tau)\mathbf{y}_{D_j^-}^T(t-\tau-1)]
\end{align}
where $D_j^-$ means $j$-th candidate document that is not clicked and
\begin{align}
\label{eq:rdssmTrain4}
&\mathbf{\delta}_{\mathbf{y}_Q}(t-\tau-1) = (1 - \mathbf{y}_{Q}(t-\tau-1))\circ\nn\\
&(1 + \mathbf{y}_{Q}(t-\tau-1))\circ\mathbf{W}_{rec}^T\mathbf{\delta}_{\mathbf{y}_Q}(t-\tau)
\end{align}
and the same as \eqref{eq:rdssmTrain4} for $\mathbf{\delta}_{\mathbf{y}_D}(t-\tau-1)$ with $D$ subscript for document side model. Please also note that: 
\begin{align}
\label{eq:rdssmTrain5}
&\mathbf{\delta}_{\mathbf{y}_Q}(T_Q) = (1-\mathbf{y}_Q(T_Q))\circ(1+\mathbf{y}_Q(T_Q))\circ\nn\\
&(b.c.\mathbf{y}_D(T_D)-a.b^3.c.\mathbf{y}_Q(T_Q)),\nn\\
&\mathbf{\delta}_{\mathbf{y}_D}(T_D) = (1-\mathbf{y}_D(T_D))\circ(1+\mathbf{y}_D(T_D))\circ\nn\\
&(b.c.\mathbf{y}_Q(T_Q)-a.b.c^3.\mathbf{y}_D(T_D))
\end{align}
where
\begin{align}
\label{eq:rdssmTrain6}
& a = \mathbf{y}_Q(t=T_Q)^T\mathbf{y}_D(t=T_D)\nn\\
& b = \frac{1}{\Vert\mathbf{y}_Q(t=T_Q)\Vert},\; c = \frac{1}{\Vert\mathbf{y}_D(t=T_D)\Vert}
\end{align}

For the input parameters, $\mathbf{\Lambda} = \mathbf{W}$:
\begin{align}
\label{eq:rdssmTrain7}
&\frac{\partial\Delta_{j,\tau}}{\partial\mathbf{W}} = [\mathbf{\delta}_{y_Q}^{D^+}(t-\tau)\mathbf{l}_{Q}^T(t-\tau) + \nn\\
&\mathbf{\delta}_{y_D}^{D^+}(t-\tau)\mathbf{l}_{D^+}^T(t-\tau)] - \nn\\
&[\mathbf{\delta}_{y_Q}^{D_j^-}(t-\tau)\mathbf{l}_{Q}^T(t-\tau) + \mathbf{\delta}_{y_D}^{D_j^-}(t-\tau)\mathbf{l}_{D_j^-}^T(t-\tau)]
\end{align}
A full derivation of BPTT for RNN is presented in section III of supplementary materials. 

\subsection{LSTM-RNN}
\label{sec:app:lstm}
Starting with the cost function in \eqref{eq:objective_func}, we use the Nesterov method described in \eqref{eq:Nesterov} to update LSTM-RNN model parameters. Here, $\mathbf{\Lambda}$ is one of the weight matrices or bias vectors $\{\mathbf{W}_1,\mathbf{W}_2,\mathbf{W}_3,\mathbf{W}_4,\mathbf{W}_{rec1},\mathbf{W}_{rec2},\mathbf{W}_{rec3},\mathbf{W}_{rec4}$\\ $,\mathbf{W}_{p1}, \mathbf{W}_{p2}, \mathbf{W}_{p3}, \mathbf{b}_1, \mathbf{b}_2, \mathbf{b}_3, \mathbf{b}_4\}$ in the LSTM-RNN architecture. The general format of the gradient of the cost function,  $\nabla L(\mathbf{\Lambda})$, is the same as \eqref{eq:rdssmTrain1}. By definition of $\Delta_{r,j}$, we have:
\begin{equation}
\label{eq:lstmTrain1}
\frac{\partial \Delta_{r,j}}{\partial\mathbf{\Lambda}} = \frac{\partial R(Q_r,D_r^+)}{\partial\mathbf{\Lambda}} - \frac{\partial R(Q_r,D_{r,j})}{\partial\mathbf{\Lambda}}
\end{equation}
We omit $r$ and $j$ subscripts for simplicity and present  $\frac{\partial R(Q,D)}{\partial\mathbf{\Lambda}}$ for different parameters of each cell of LSTM-RNN in the following subsections. This will complete the process of calculating $\nabla L(\mathbf{\Lambda})$ in \eqref{eq:rdssmTrain1} and then we can use \eqref{eq:Nesterov} to update LSTM-RNN model parameters. In the subsequent subsections vectors $\mathbf{v}_Q$ and $\mathbf{v}_D$ are defined as:
\begin{align}
\label{eq:lstmTrain2}
&\mathbf{v}_Q = (b.c.\mathbf{y}_D(t=T_D) - a.b^3.c.\mathbf{y}_Q(t=T_Q))\nn\\
&\mathbf{v}_D = (b.c.\mathbf{y}_Q(t=T_Q) - a.b.c^3.\mathbf{y}_D(t=T_D))
\end{align}
where $a,\;b$ and $c$ are defined in \eqref{eq:rdssmTrain6}. Full derivation of truncated BPTT for LSTM-RNN model is presented in section III of supplementary materials.

\subsubsection{Output Gate}
\label{sec:o_g}
For recurrent connections we have:
\begin{equation}
\label{eq:LSTM8c}
\frac{\partial R(Q,D)}{\partial\mathbf{W}_{rec1}} = \mathbf{\delta}_{y_Q}^{rec1}(t).\mathbf{y}_Q(t-1)^T + \mathbf{\delta}_{y_D}^{rec1}(t).\mathbf{y}_D(t-1)^T
\end{equation}
where
\begin{equation}
\label{eq:LSTM9c}
\mathbf{\delta}_{y_Q}^{rec1}(t) = \mathbf{o}_Q(t)\circ(1-\mathbf{o}_Q(t))\circ h(\mathbf{c}_Q(t))\circ \mathbf{v}_Q(t)
\end{equation}
and the same as \eqref{eq:LSTM9c} for $\mathbf{\delta}_{y_D}^{rec1}(t)$ with subscript $D$ for document side model.  
For input connections, $\mathbf{W}_1$, and peephole connections, $\mathbf{W}_{p1}$, we will have:
\begin{equation}
\label{eq:LSTM10}
\frac{\partial R(Q,D)}{\partial\mathbf{W}_1} = \mathbf{\delta}_{y_Q}^{rec1}(t).\mathbf{l}_Q(t)^T + \mathbf{\delta}_{y_D}^{rec1}(t).\mathbf{l}_D(t)^T
\end{equation}
\begin{equation}
\label{eq:LSTM11}
\frac{\partial R(Q,D)}{\partial\mathbf{W}_{p1}} = \mathbf{\delta}_{y_Q}^{rec1}(t).\mathbf{c}_Q(t)^T + \mathbf{\delta}_{y_D}^{rec1}(t).\mathbf{c}_D(t)^T
\end{equation}
The derivative for output gate bias values will be:
\begin{equation}
\label{eq:LSTM11Bias}
\frac{\partial R(Q,D)}{\partial\mathbf{b}_{1}} = \mathbf{\delta}_{y_Q}^{rec1}(t) + \mathbf{\delta}_{y_D}^{rec1}(t)
\end{equation}

\subsubsection{Input Gate}
\label{sec:o_i}
For the recurrent connections we have:
\begin{align}
\label{eq:LSTM20c}
&\frac{\partial R(Q,D)}{\partial \mathbf{W}_{rec3}} =\nn\\
&diag(\mathbf{\delta}_{y_Q}^{rec3}(t)).\frac{\partial \mathbf{c}_Q(t)}{\partial \mathbf{W}_{rec3}}
+ diag(\mathbf{\delta}_{y_D}^{rec3}(t)).\frac{\partial \mathbf{c}_D(t)}{\partial \mathbf{W}_{rec3}}
\end{align}
where
\begin{align}
\label{eq:LSTM21c}
&\mathbf{\delta}_{y_Q}^{rec3}(t) = (1-h(\mathbf{c}_Q(t)))\circ (1+h(\mathbf{c}_Q(t)))\circ \mathbf{o}_Q(t) \circ \mathbf{v}_Q(t)\nn\\
&\frac{\partial \mathbf{c}_Q(t)}{\partial \mathbf{W}_{rec3}} = diag(\mathbf{f}_Q(t)).\frac{\partial \mathbf{c}_Q(t-1)}{\partial \mathbf{W}_{rec3}} + \mathbf{b}_{i,Q}(t).\mathbf{y}_Q(t-1)^T\nn\\
&\mathbf{b}_{i,Q}(t) = \mathbf{y}_{g,Q}(t)\circ \mathbf{i}_Q(t) \circ (1-\mathbf{i}_Q(t))
\end{align}
In equation \eqref{eq:LSTM20c},  $\mathbf{\delta}_{y_D}^{rec3}(t)$ and $\frac{\partial \mathbf{c}_D(t)}{\partial \mathbf{W}_{rec3}} $ are the same as \eqref{eq:LSTM21c} with $D$ subscript.
For the input connections we will have the following:
\begin{align}
\label{eq:LSTM22c}
&\frac{\partial R(Q,D)}{\partial \mathbf{W}_{3}} =\nn\\
&diag(\mathbf{\delta}_{y_Q}^{rec3}(t)).\frac{\partial \mathbf{c}_Q(t)}{\partial \mathbf{W}_{3}}
+ diag(\mathbf{\delta}_{y_D}^{rec3}(t)).\frac{\partial \mathbf{c}_D(t)}{\partial \mathbf{W}_{3}}
\end{align}
where
\begin{equation}
\label{eq:LSTM23c}
\frac{\partial \mathbf{c}_Q(t)}{\partial \mathbf{W}_{3}} = diag(\mathbf{f}_Q(t)).\frac{\partial \mathbf{c}_Q(t-1)}{\partial \mathbf{W}_{3}} + \mathbf{b}_{i,Q}(t).\mathbf{x}_Q(t)^T
\end{equation}
For the peephole connections we will have:
\begin{align}
\label{eq:LSTM24c}
&\frac{\partial R(Q,D)}{\partial \mathbf{W}_{p3}} = \nn\\
&diag(\mathbf{\delta}_{y_Q}^{rec3}(t)).\frac{\partial \mathbf{c}_Q(t)}{\partial \mathbf{W}_{p3}} + diag(\mathbf{\delta}_{y_D}^{rec3}(t)).\frac{\partial \mathbf{c}_D(t)}{\partial \mathbf{W}_{p3}}
\end{align}
where
\begin{equation}
\label{eq:LSTM25c}
\frac{\partial \mathbf{c}_Q(t)}{\partial \mathbf{W}_{p3}} = diag(\mathbf{f}_Q(t)).\frac{\partial \mathbf{c}_Q(t-1)}{\partial \mathbf{W}_{p3}} + \mathbf{b}_{i,Q}(t).\mathbf{c}_Q(t-1)^T
\end{equation}
For bias values, $\mathbf{b}_3$, we will have:
\begin{align}
\label{eq:LSTM26c}
&\frac{\partial R(Q,D)}{\partial \mathbf{b}_{3}} = \nn\\
&diag(\mathbf{\delta}_{y_Q}^{rec3}(t)).\frac{\partial \mathbf{c}_Q(t)}{\partial \mathbf{b}_{3}} + diag(\mathbf{\delta}_{y_D}^{rec3}(t)).\frac{\partial \mathbf{c}_D(t)}{\partial \mathbf{b}_{3}}
\end{align}
where
\begin{equation}
\label{eq:LSTM27c}
\frac{\partial \mathbf{c}_Q(t)}{\partial \mathbf{b}_{3}} = diag(\mathbf{f}_Q(t)).\frac{\partial \mathbf{c}_Q(t-1)}{\partial \mathbf{b}_{3}} + \mathbf{b}_{i,Q}(t)
\end{equation}
\subsubsection{Forget Gate}
\label{sec:o_f}
For the recurrent connections we will have:
\begin{align}
\label{eq:LSTM30c}
&\frac{\partial R(Q,D)}{\partial \mathbf{W}_{rec2}} = \nn\\
&diag(\mathbf{\delta}_{y_Q}^{rec2}(t)).\frac{\partial \mathbf{c}_Q(t)}{\partial \mathbf{W}_{rec2}} + diag(\mathbf{\delta}_{y_D}^{rec2}(t)).\frac{\partial \mathbf{c}_D(t)}{\partial \mathbf{W}_{rec2}}
\end{align}
where
\begin{align}
\label{eq:LSTM31c}
&\mathbf{\delta}_{y_Q}^{rec2}(t) = (1-h(\mathbf{c}_Q(t)))\circ (1+h(\mathbf{c}_Q(t)))\circ \mathbf{o}_Q(t) \circ \mathbf{v}_Q(t)\nn\\
&\frac{\partial \mathbf{c}_Q(t)}{\partial \mathbf{W}_{rec2}} = diag(\mathbf{f}_Q(t)).\frac{\partial \mathbf{c}_Q(t-1)}{\partial \mathbf{W}_{rec2}} + \mathbf{b}_{f,Q}(t).\mathbf{y}_Q(t-1)^T\nn\\
&\mathbf{b}_{f,Q}(t) = \mathbf{c}_Q(t-1)\circ \mathbf{f}_Q(t) \circ (1-\mathbf{f}_Q(t))
\end{align}
For input connections to forget gate we will have:
\begin{align}
\label{eq:LSTM32c}
&\frac{\partial R(Q,D)}{\partial \mathbf{W}_{2}} = \nn\\
&diag(\mathbf{\delta}_{y_Q}^{rec2}(t)).\frac{\partial \mathbf{c}_Q(t)}{\partial \mathbf{W}_{2}} + diag(\mathbf{\delta}_{y_D}^{rec2}(t)).\frac{\partial \mathbf{c}_D(t)}{\partial \mathbf{W}_{2}}
\end{align}
where
\begin{equation}
\label{eq:LSTM33c}
\frac{\partial \mathbf{c}_Q(t)}{\partial \mathbf{W}_{2}} = diag(\mathbf{f}_Q(t)).\frac{\partial \mathbf{c}_Q(t-1)}{\partial \mathbf{W}_{2}} + \mathbf{b}_{f,Q}(t).\mathbf{x}_Q(t)^T
\end{equation}
For peephole connections we have:
\begin{align}
\label{eq:LSTM34c}
&\frac{\partial R(Q,D)}{\partial \mathbf{W}_{p2}} = \nn\\
&diag(\mathbf{\delta}_{y_Q}^{rec2}(t)).\frac{\partial \mathbf{c}_Q(t)}{\partial \mathbf{W}_{p2}} + diag(\mathbf{\delta}_{y_D}^{rec2}(t)).\frac{\partial \mathbf{c}_D(t)}{\partial \mathbf{W}_{p2}}
\end{align}
where
\begin{equation}
\label{eq:LSTM35c}
\frac{\partial \mathbf{c}_Q(t)}{\partial \mathbf{W}_{p2}} = diag(\mathbf{f}_Q(t)).\frac{\partial \mathbf{c}_Q(t-1)}{\partial \mathbf{W}_{p2}} + \mathbf{b}_{f,Q}(t).\mathbf{c}_Q(t-1)^T
\end{equation}
For forget gate's bias values we will have:
\begin{align}
\label{eq:LSTM36c}
&\frac{\partial R(Q,D)}{\partial \mathbf{b}_{2}} = \nn\\
&diag(\mathbf{\delta}_{y_Q}^{rec2}(t)).\frac{\partial \mathbf{c}_Q(t)}{\partial \mathbf{b}_{2}} + diag(\mathbf{\delta}_{y_D}^{rec2}(t)).\frac{\partial \mathbf{c}_D(t)}{\partial \mathbf{b}_{2}}
\end{align}
where
\begin{equation}
\label{eq:LSTM37c}
\frac{\partial \mathbf{c}_Q(t)}{\partial \mathbf{b}_{2}} = diag(\mathbf{f}_Q(t)).\frac{\partial \mathbf{c}_Q(t-1)}{\partial \mathbf{b}_{3}} + \mathbf{b}_{f,Q}(t)
\end{equation}

\subsubsection{Input without Gating ($\mathbf{y}_g(t)$)}
\label{sec:o_yg}
For recurrent connections we will have:
\begin{align}
\label{eq:LSTM40c}
&\frac{\partial R(Q,D)}{\partial \mathbf{W}_{rec4}} = \nn\\
&diag(\mathbf{\delta}_{y_Q}^{rec4}(t)).\frac{\partial \mathbf{c}_Q(t)}{\partial \mathbf{W}_{rec4}} + diag(\mathbf{\delta}_{y_D}^{rec4}(t)).\frac{\partial \mathbf{c}_D(t)}{\partial \mathbf{W}_{rec4}}
\end{align}
where
\begin{align}
\label{eq:LSTM41c}
&\mathbf{\delta}_{y_Q}^{rec4}(t) = (1-h(\mathbf{c}_Q(t)))\circ (1+h(\mathbf{c}_Q(t)))\circ \mathbf{o}_Q(t) \circ \mathbf{v}_Q(t)\nn\\
&\frac{\partial \mathbf{c}_Q(t)}{\partial \mathbf{W}_{rec4}} = diag(\mathbf{f}_Q(t)).\frac{\partial \mathbf{c}_Q(t-1)}{\partial \mathbf{W}_{rec4}} + \mathbf{b}_{g,Q}(t).\mathbf{y}_Q(t-1)^T\nn\\
&\mathbf{b}_{g,Q}(t) = \mathbf{i}_Q(t) \circ (1-\mathbf{y}_{g,Q}(t))\circ (1+\mathbf{y}_{g,Q}(t))
\end{align}
For input connection we have:
\begin{align}
\label{eq:LSTM42c}
&\frac{\partial R(Q,D)}{\partial \mathbf{W}_{4}} = \nn\\
&diag(\mathbf{\delta}_{y_Q}^{rec4}(t)).\frac{\partial \mathbf{c}_Q(t)}{\partial \mathbf{W}_{4}} + diag(\mathbf{\delta}_{y_D}^{rec4}(t)).\frac{\partial \mathbf{c}_D(t)}{\partial \mathbf{W}_{4}}
\end{align}
where
\begin{equation}
\label{eq:LSTM43c}
\frac{\partial \mathbf{c}_Q(t)}{\partial \mathbf{W}_{4}} = diag(\mathbf{f}_Q(t)).\frac{\partial \mathbf{c}_Q(t-1)}{\partial \mathbf{W}_{4}} + \mathbf{b}_{g,Q}(t).\mathbf{x}_Q(t)^T
\end{equation}
For bias values we will have:
\begin{align}
\label{eq:LSTM44c}
&\frac{\partial R(Q,D)}{\partial \mathbf{b}_{4}} = \nn\\
&diag(\mathbf{\delta}_{y_Q}^{rec4}(t)).\frac{\partial \mathbf{c}_Q(t)}{\partial \mathbf{b}_{4}} + diag(\mathbf{\delta}_{y_D}^{rec4}(t)).\frac{\partial \mathbf{c}_D(t)}{\partial \mathbf{b}_{4}}
\end{align}
where
\begin{equation}
\label{eq:LSTM45c}
\frac{\partial \mathbf{c}_Q(t)}{\partial \mathbf{b}_{4}} = diag(\mathbf{f}_Q(t)).\frac{\partial \mathbf{c}_Q(t-1)}{\partial \mathbf{b}_{4}} + \mathbf{b}_{g,Q}(t)
\end{equation}

\subsubsection{Error signal backpropagation}
\label{sec:errBPP}
Error signals are back propagated through time using following equations:
\begin{align}
\label{eq:LSTM46}
&\mathbf{\delta}_Q^{rec1}(t-1) = \nn\\
&[\mathbf{o}_Q(t-1)\circ (1-\mathbf{o}_Q(t-1))\circ h(\mathbf{c}_Q(t-1))]\nn\\
&\circ \mathbf{W}_{rec1}^T.\mathbf{\delta}_Q^{rec1}(t)
\end{align}
\begin{align}
\label{eq:LSTM47}
&\mathbf{\delta}_Q^{rec_i}(t-1) = [(1-h(\mathbf{c}_Q(t-1)))\circ (1+h(\mathbf{c}_Q(t-1)))\nn\\
&\circ \mathbf{o}_Q(t-1)]\circ \mathbf{W}_{rec_i}^T.\mathbf{\delta}_Q^{rec_i}(t),\;\;\;\;for \;\;\; i\in \{2,3,4\}
\end{align}

\ifCLASSOPTIONcaptionsoff
  \newpage
\fi

\bibliographystyle{IEEEtran}
\bibliography{refs}

\clearpage
\newpage

\section*{\huge Supplementary Material}

\vspace{2em}

\section{A more clear figure for input gate for ``$hotels\; in\; shanghai$'' example}
\label{sec:app:igate}
In this section we present a more clear figure for part (a) of Fig. 5 that shows the structure of the input gate for document side of ``$hotels\; in\; shanghai$'' example. As it is clearly visible from this figure, the input gate values for most of the cells corresponding to word $3$, word $7$ and word $9$ in document side of LSTM-RNN have very small values (light green-yellow color). These are corresponding to words ``$accommodation$'', ``$discount$'' and ``$reservation$'' respectively in the document title. Interestingly, input gates are trying to reduce effect of these three words in the final representation ($\mathbf{y}(t)$) because the LSTM-RNN model is trained to maximize the similarity between query and document if they are a good match. 
\begin{figure}[t]
\includegraphics[width = 0.48\textwidth]{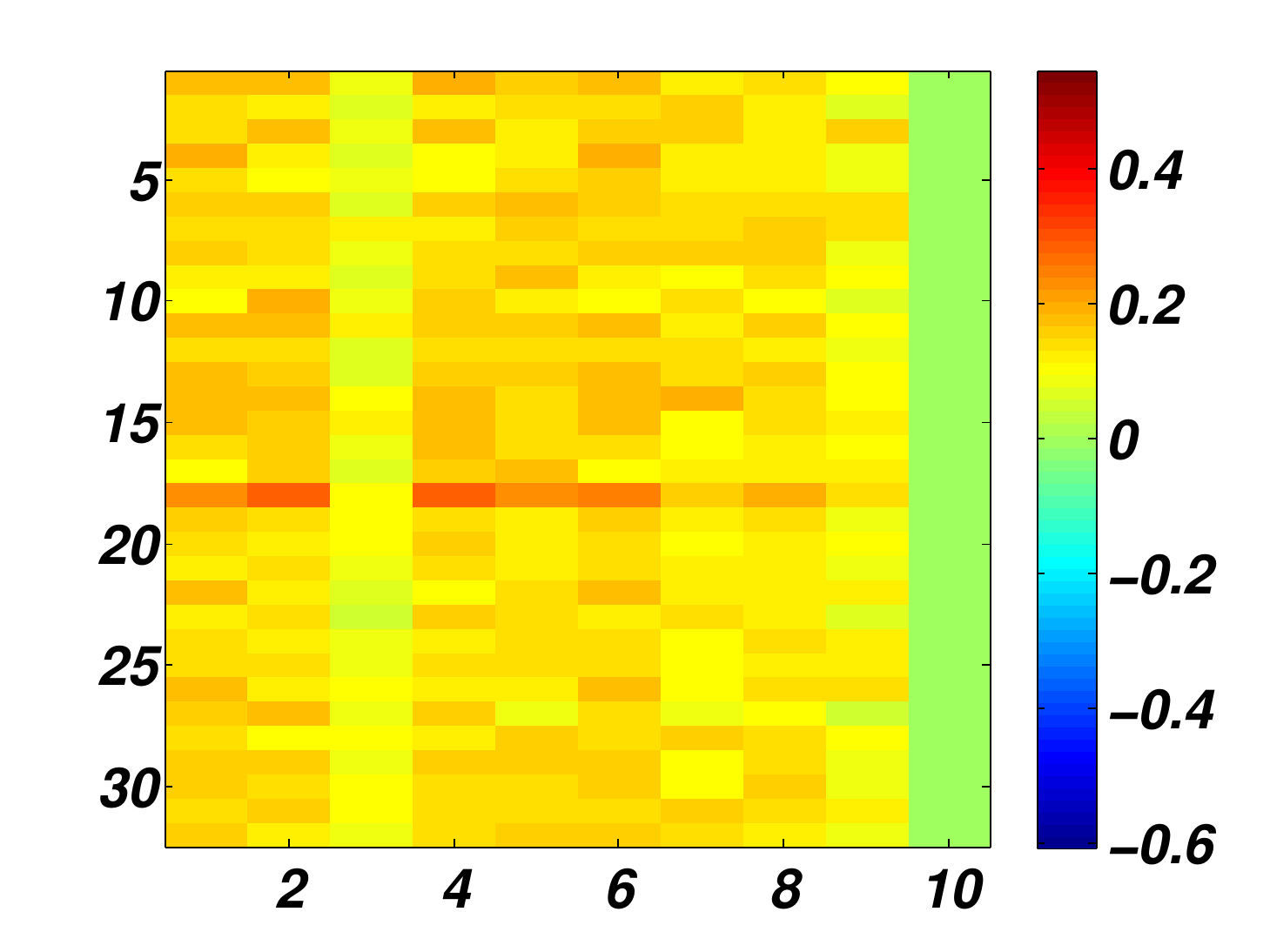}
\caption{Input gate, $\mathbf{i}(t)$, for Document: ``\emph{shanghai hotels accommodation hotel in shanghai discount and reservation}''}
\label{fig:igateQ}
\end{figure}

\section{A Closer Look at RNNs with and without LSTM Cells in Web Document Retrieval Task}
\label{sec:app:RDSSMvsLSTM-DSSM}

In this section we further show examples to reveal the advantage of LSTM-RNN sentence embedding compared to the RNN sentence embedding.

First, we compare the scores assigned by trained RNN and LSTM-RNN to our ``\emph{hotels in shanghai}'' example. On average, each query in our test dataset is associated with 15 web documents (URLs). Each query / document pair has a relevance label which is human generated. These relevance labels are ``Bad'', ``Fair'', ``Good'' and ``Excellent''. This example is rated as a ``Good'' match in the dataset. The score for this pair assigned by RNN is ``$0.8165$'' while the score assigned by LSTM-RNN is ``$0.9161$''. Please note that the score is between $0$ and $1$. This means that the score assigned by LSTM-RNN is more correspondent with the human generated label.

Second, we compare the number of assigned neurons and cells to each word by RNN and LSTM-RNN respectively. To do this, we rely on the 10 most active cells and neurons in the final semantic vectors in both models. Results are presented in Table \ref{table:Q1RDSSM} and Table \ref{table:D1RDSSM} for query and document respectively. An interesting observation is that RNN sometimes assigns neurons to unimportant words, e.g., 6 neurons are assigned to the word ``$in$'' in Table \ref{table:D1RDSSM}. 
\begin{table}[t]
\caption{RNNs with \& without LSTM cells for the same query: ``$hotels\; in \; shanghai$''}
\label{table:Q1RDSSM}
\vspace{-.5cm}
\begin{center}
\scriptsize
\begin{tabular}{ | c | c | c |c | } 
\hline
     & $hotels$ & $in$ & $shanghai$\\ 
\hline
Number of assigned &  &  & \\
cells out of 10 (LSTM-RNN) & - & 0 & 7 \\ 
\hline
Number of assigned &  &  & \\
neurons out of 10 (RNN) & - & 2 & 9 \\
\hline 
\end{tabular}
\end{center}
\end{table}
\begin{table*}[t]
\caption{RNNs with \& without LSTM cells for the same Document:
``\emph{shanghai hotels accommodation hotel in shanghai discount and reservation}''}
\label{table:D1RDSSM}
\vspace{-.5cm}
\begin{center}
\scriptsize
\begin{tabular}{ | c | c | c |c |c| c | c | c |c |c| }
\hline
     & $shanghai$ & $hotels$ & $accommodation$ & $hotel$ & $in$ & $shanghai$ & $discount$ & $and$ & $reservation$\\ 
\hline
Number of assigned &  &  &  &  &  &  &  &  & \\
cells out of 10 (LSTM-RNN) & - & 4 & 3 & 8 & 1 & 8 & 5 & 3 & 4 \\ 
\hline
Number of assigned &  &  &  &  &  &  &  &  & \\
neurons out of 10 (RNN) & - & 10 & 7 & 9 & 6 & 8 & 3 & 2 & 6 \\ 
\hline
\end{tabular}
\end{center}
\end{table*}

As another example we consider the query, ``$how\; to \; fix\; bath\; tub\; wont\; turn\; off$''. This example is rated as a ``Bad'' match in the dataset by human. It is good to know that the score for this pair assigned by RNN is ``$0.7016$'' while the score assigned by LSTM-RNN is ``$0.5944$''. This shows the score generated by LSTM-RNN is closer to human generated label.

Number of assigned neurons and cells to each word by RNN and LSTM-RNN are presented in Table \ref{table:Q2RDSSM} and Table \ref{table:D2RDSSM} for query and document. This is out of 10 most active neurons and cells in the semantic vector of RNN and LSTM-RNN. Examples of RNN assigning neurons to unimportant words are 3 neurons to the word ``a'' and 4 neurons to the word ``you'' in Table \ref{table:D2RDSSM}.  
\begin{table*}[t]
\caption{RNN versus LSTM-RNN for Query: ``$how\; to \; fix\; bath\; tub\; wont\; turn\; off$''}
\label{table:Q2RDSSM}
\vspace{-.5cm}
\begin{center}
\scriptsize
\begin{tabular}{ | c | c | c |c | c | c | c |c |c| }
\hline
     & $how$ & $to$ & $fix$& $bath$& $tub$& $wont$& $turn$& $off$\\ 
\hline
Number of assigned &  &  &  & & & & &  \\
cells out of 10 (LSTM-RNN) & - & 0 & 4 & 7 & 6& 3& 5& 0 \\ 
\hline
Number of assigned &  &  &  & & & & &  \\
neurons out of 10 (RNN) & - & 1 & 10 & 4 & 6& 2& 7& 1 \\ 
\hline
\end{tabular}
\end{center}
\end{table*}
\begin{table*}[t]
\caption{RNN versus LSTM-RNN for Document: ``$how\; do\; you\; paint\; a\; bathtub\; and\; what\; paint\; should\dots$''}
\label{table:D2RDSSM}
\vspace{-.5cm}
\begin{center}
\scriptsize
\begin{tabular}{ | c | c | c |c |c| c | c | c |c |c|c| }
\hline
     & $how$ & $do$ & $you$ & $paint$ & $a$ & $bathtub$ & $and$ & $what$ & $paint$&$should\; you \dots$\\ 
\hline
Number of assigned &  &  &  &  &  &  &  &  & & \\
cells out of 10(LSTM-RNN) & - & 1 & 1 & 7 & 0 & 9 & 2 & 3 & 8 & 4 \\ 
\hline
Number of assigned &  &  &  &  &  &  &  &  & & \\
neurons out of 10(RNN) & - & 1 & 4 & 4 & 3 & 7 & 2 & 5 & 4 & 7 \\ 
\hline
\end{tabular}
\end{center}
\end{table*}

\section{Derivation of BPTT for RNN and LSTM-RNN}
\label{sec:app:derive}
In this appendix we present the full derivation of the gradients for RNN and LSTM-RNN.
\subsection{Derivation of BPTT for RNN}
\label{sec:appRDSSM_Full}
From (4) and (5) we have:
\begin{equation}
\label{eq:appR1}
\frac{\partial L(\mathbf{\Lambda})}{\partial \mathbf{\Lambda}} = \sum_{r=1}^N\frac{\partial l_r(\mathbf{\Lambda})}{\partial \mathbf{\Lambda}}=
-\sum_{r=1}^N\sum_{j=1}^{n}\alpha_{r,j}\frac{\partial \Delta_{r,j}}{\partial\mathbf{\Lambda}}
\end{equation}
where
\begin{equation}
\label{eq:appR2}
\alpha_{r,j} = \frac{-\gamma e^{-\gamma\Delta_{r,j}}}{1+\sum_{j=1}^{n}e^{-\gamma \Delta_{r,j}}}
\end{equation}
and
\begin{equation}
\label{eq:appR3}
\Delta_{r,j} = R(Q_r,D_r^+) - R(Q_r,D_{r,j})
\end{equation}
We need to find $\frac{\partial \Delta_{r,j}}{\partial\mathbf{\Lambda}}$ for input weights and recurrent weights. We omit $r$ subscript for simplicity.
\subsubsection{Recurrent Weights}
\label{subsec:recurrent}
\begin{equation}
\label{eq:appR4}
\frac{\partial\Delta_j}{\partial\mathbf{W}_{rec}} = \frac{\partial R(Q,D^+)}{\partial \mathbf{W}_{rec}} - \frac{\partial R(Q,D_j^-)}{\partial \mathbf{W}_{rec}}
\end{equation}
We divide $R(D,Q)$ into three components:
\begin{align}
\label{eq:appR5}
&R(Q,D) = \underbrace{\mathbf{y}_Q(t=T_Q)^T\mathbf{y}_D(t=T_D)}_{a}.\nn\\
&\underbrace{\frac{1}{\Vert\mathbf{y}_Q(t=T_Q)\Vert}}_{b}.\underbrace{\frac{1}{\Vert\mathbf{y}_D(t=T_D)\Vert}}_{c}
\end{align}
then
\begin{align}
\label{eq:appR6}
&\frac{\partial R(Q,D)}{\partial \mathbf{W}_{rec}} = \underbrace{\frac{\partial a}{\partial \mathbf{W}_{rec}}.b.c}_{\mathbf{D}}+ \underbrace{a.\frac{\partial b}{\partial \mathbf{W}_{rec}}.c}_{\mathbf{E}}+\nn\\
&\underbrace{a.b.\frac{\partial c}{\partial \mathbf{W}_{rec}}}_{\mathbf{F}}
\end{align}
We have
\begin{align}
\label{eq:appR7}
&\mathbf{D} = \frac{\partial \mathbf{y}_Q(t=T_Q)^T\mathbf{y}_D(t=T_D).b.c}{\partial \mathbf{W}_{rec}} \nn\\
&= \frac{\partial \mathbf{y}_Q(t=T_Q)^T\mathbf{y}_D(t=T_D).b.c}{\partial \mathbf{y}_Q(t=T_Q)}.\frac{\partial \mathbf{y}_Q(t=T_Q)}{\partial \mathbf{W}_{rec}} + \nn\\
&\frac{\partial \mathbf{y}_Q(t=T_Q)^T\mathbf{y}_D(t=T_D).b.c}{\partial \mathbf{y}_D(t=T_D)}.\frac{\partial \mathbf{y}_D(t=T_D)}{\partial \mathbf{W}_{rec}}\nn\\
&= \mathbf{y}_D(t=T_D).b.c.\frac{\partial \mathbf{y}_Q(t=T_Q)}{\partial \mathbf{W}_{rec}} + \nn\\
&\mathbf{y}_Q(t=T_Q).\underbrace{(b.c)^T}_{b.c}.\frac{\partial \mathbf{y}_D(t=T_D)}{\partial \mathbf{W}_{rec}}
\end{align}
Since $f(.) = tanh(.)$, using chain rule we have
\begin{align}
\label{eq:appR8}
&\frac{\partial \mathbf{y}_Q(t=T_Q)}{\mathbf{W}_{rec}} = \nn\\
&[(1-\mathbf{y}_Q(t=T_Q))\circ(1+\mathbf{y}_Q(t=T_Q))]\mathbf{y}_Q(t-1)^T
\end{align}
and therefore
\begin{align}
\label{eq:appR9}
&\mathbf{D} = [b.c.\mathbf{y}_D(t=T_D)\circ(1-\mathbf{y}_Q(t=T_Q))\circ\nn\\
&(1+\mathbf{y}_Q(t=T_Q))]\mathbf{y}_Q(t-1)^T+\nn\\ 
&[b.c.\mathbf{y}_Q(t=T_Q)\circ(1-\mathbf{y}_D(t=T_D))\circ\nn\\
&(1+\mathbf{y}_D(t=T_D))]\mathbf{y}_D(t-1)^T 
\end{align}
To find $\mathbf{E}$ we use following basic rule:
\begin{equation}
\label{eq:appR10}
\frac{\partial}{\partial\mathbf{x}}\Vert\mathbf{x}-\mathbf{a}\Vert_2 = \frac{\mathbf{x}-\mathbf{a}}{\Vert\mathbf{x}-\mathbf{a}\Vert_2}
\end{equation}
Therefore
\begin{align}
\label{eq:appR11}
&\mathbf{E} = a.c.\frac{\partial}{\partial\mathbf{W}_{rec}}(\Vert\mathbf{y}_Q(t=T_Q)\Vert)^{-1} = \nn\\
&-a.c.(\Vert\mathbf{y}_Q(t=T_Q)\Vert)^{-2}.\frac{\partial\Vert\mathbf{y}_Q(t=T_Q)\Vert}{\partial \mathbf{W}_{rec}}\nn\\
& = -a.c.(\Vert\mathbf{y}_Q(t=T_Q)\Vert)^{-2}.\frac{\mathbf{y}_Q(t=T_Q)}{\Vert\mathbf{y}_Q(t=T_Q)\Vert}\frac{\partial\mathbf{y}_Q(t=T_Q)}{\partial \mathbf{W}_{rec}} \nn\\
&= -[a.c.b^3.\mathbf{y}_Q(t=T_Q)\circ(1-\mathbf{y}_Q(t=T_Q))\circ\nn\\
&(1+\mathbf{y}_Q(t=T_Q))]\mathbf{y}_Q(t-1)
\end{align}
$\mathbf{F}$ is calculated similar to \eqref{eq:appR11}:
\begin{align}
\label{eq:appR12}
&\mathbf{F} = -[a.b.c^3.\mathbf{y}_D(t=T_D)\circ(1-\mathbf{y}_D(t=T_D))\circ\nn\\
&(1+\mathbf{y}_D(t=T_D))]\mathbf{y}_D(t-1)
\end{align}
Considering \eqref{eq:appR6},\eqref{eq:appR9},\eqref{eq:appR11} and \eqref{eq:appR12} we have:
\begin{equation}
\label{eq:appR13}
\frac{\partial R(Q,D)}{\partial \mathbf{W}_{rec}} = \mathbf{\delta}_{\mathbf{y}_Q}(t)\mathbf{y}_Q(t-1)^T + \mathbf{\delta}_{\mathbf{y}_D}(t)\mathbf{y}_D(t-1)^T
\end{equation}
where
\begin{align}
\label{eq:appR14}
&\mathbf{\delta}_{\mathbf{y}_Q}(t=T_Q) = (1-\mathbf{y}_Q(t=T_Q))\circ(1+\mathbf{y}_Q(t=T_Q))\circ\nn\\
&(b.c.\mathbf{y}_D(t=T_D)-a.b^3.c.\mathbf{y}_Q(t=T_Q)),\nn\\
&\mathbf{\delta}_{\mathbf{y}_D}(t=T_D) = (1-\mathbf{y}_D(t=T_D))\circ(1+\mathbf{y}_D(t=T_D))\circ\nn\\
&(b.c.\mathbf{y}_Q(t=T_Q)-a.b.c^3.\mathbf{y}_D(t=T_D))
\end{align}
Equation \eqref{eq:appR14} will just unfold the network one time step, to unfold it over rest of time steps using backpropagation we have:
\begin{align}
\label{eq:appR15}
&\mathbf{\delta}_{\mathbf{y}_Q}(t-\tau-1) = (1 - \mathbf{y}_{Q}(t-\tau-1))\circ\nn\\
&(1 + \mathbf{y}_{Q}(t-\tau-1))\circ\mathbf{W}_{rec}^T\mathbf{\delta}_{\mathbf{y}_Q}(t-\tau),\nn\\
&\mathbf{\delta}_{\mathbf{y}_D}(t-\tau-1) = (1 - \mathbf{y}_{D}(t-\tau-1))\circ\nn\\
&(1 + \mathbf{y}_{D}(t-\tau-1))\circ\mathbf{W}_{rec}^T\mathbf{\delta}_{\mathbf{y}_D}(t-\tau)
\end{align}
where $\tau$ is the number of time steps that we unfold the network over time which is from $0$ to $T_Q$ and $T_D$ for queries and documents respectively. Now using \eqref{eq:appR4} we have:
\begin{align}
\label{eq:appR16}
&\frac{\partial\Delta_{j,\tau}}{\partial\mathbf{W}_{rec}} = [\mathbf{\delta}_{y_Q}^{D^+}(t-\tau)\mathbf{y}_{Q}^T(t-\tau-1) + \nn\\
&\mathbf{\delta}_{y_D}^{D^+}(t-\tau)\mathbf{y}_{D^+}^T(t-\tau-1)]-[\mathbf{\delta}_{y_Q}^{D_j^-}(t-\tau)\mathbf{y}_{Q}^T(t-\tau-1)\nn\\
&+\mathbf{\delta}_{y_D}^{D_j^-}(t-\tau)\mathbf{y}_{D_j^-}^T(t-\tau-1)]
\end{align}
To calculate final value of gradient we should fold back the network over time and use \eqref{eq:appR1}, we will have:
\begin{equation}
\label{eq:appR17}
\frac{\partial L(\mathbf{\Lambda})}{\partial \mathbf{W}_{rec}} = \underbrace{-\sum_{r=1}^N\sum_{j=1}^{n}\sum_{\tau=0}^{T}\alpha_{r,j,T_{D,Q}}\frac{\partial \Delta_{r,j,\tau}}{\partial\mathbf{W}_{rec}}}_{\mathrm{one\; large\; update}}
\end{equation}
\subsubsection{Input Weights}
\label{subsec:input}
Using a similar procedure we will have the following for input weights:
\begin{equation}
\label{eq:appR18}
\frac{\partial R(Q,D)}{\partial \mathbf{W}} = \mathbf{\delta}_{\mathbf{y}_Q}(t-\tau)\mathbf{l}_{Q}(t-\tau)^T + \mathbf{\delta}_{\mathbf{y}_D}(t-\tau)\mathbf{l}_{D}(t-\tau)^T
\end{equation}
where
\begin{align}
\label{eq:appR19}
&\mathbf{\delta}_{\mathbf{y}_Q}(t-\tau) = (1-\mathbf{y}_Q(t-\tau))\circ(1+\mathbf{y}_Q(t-\tau))\circ\nn\\
&(b.c.\mathbf{y}_D(t-\tau)-a.b^3.c.\mathbf{y}_Q(t-\tau)),\nn\\
& \mathbf{\delta}_{\mathbf{y}_D}(t-\tau) = (1-\mathbf{y}_D(t-\tau))\circ(1+\mathbf{y}_D(t-\tau))\circ\nn\\
&(b.c.\mathbf{y}_Q(t-\tau)-a.b.c^3.\mathbf{y}_D(t-\tau))
\end{align}
Therefore:
\begin{align}
\label{eq:appR20}
&\frac{\partial\Delta_{j,\tau}}{\partial\mathbf{W}} =\nn\\ 
&[\mathbf{\delta}_{y_Q}^{D^+}(t-\tau)\mathbf{l}_{Q}^T(t-\tau) + 
\mathbf{\delta}_{y_D}^{D^+}(t-\tau)\mathbf{l}_{D^+}^T(t-\tau)] - \nn\\
&[\mathbf{\delta}_{y_Q}^{D_j^-}(t-\tau)\mathbf{l}_{Q}^T(t-\tau) + \mathbf{\delta}_{y_D}^{D_j^-}(t-\tau)\mathbf{l}_{D_j^-}^T(t-\tau)]
\end{align}
and therefore:
\begin{equation}
\label{eq:appR21}
\frac{\partial L(\mathbf{\Lambda})}{\partial \mathbf{W}} = \underbrace{-\sum_{r=1}^N\sum_{j=1}^n\sum_{\tau=0}^{T}\alpha_{r,j}\frac{\partial \Delta_{r,j,\tau}}{\partial\mathbf{W}}}_{\mathrm{one\; large\; update}}
\end{equation}

\subsection{Derivation of BPTT for LSTM-RNN}
\label{sec:appLSTM_Full}
Following from \eqref{eq:appR6} for every parameter, $\mathbf{\Lambda}$, in LSTM-RNN architecture we have:  
\begin{equation}
\label{eq:LSTM1}
\frac{\partial R(Q,D)}{\partial\mathbf{\Lambda}} = \underbrace{\frac{\partial a}{\partial \mathbf{\Lambda}}.b.c}_{\mathbf{D}}+ \underbrace{a.\frac{\partial b}{\partial \mathbf{\Lambda}}.c}_{\mathbf{E}}+ \underbrace{a.b.\frac{\partial c}{\partial \mathbf{\Lambda}}}_{\mathbf{F}}
\end{equation}
and from \eqref{eq:appR7}:
\begin{align}
\label{eq:LSTM2}
&\mathbf{D} = \mathbf{y}_D(t=T_D).b.c.\frac{\partial \mathbf{y}_Q(t=T_Q)}{\partial \mathbf{\Lambda}} + \nn\\
&\mathbf{y}_Q(t=T_Q).b.c.\frac{\partial \mathbf{y}_D(t=T_D)}{\partial \mathbf{\Lambda}}
\end{align}
From \eqref{eq:appR11} and \eqref{eq:appR12} we have:
\begin{equation}
\label{eq:LSTM3}
\mathbf{E} = -a.c.b^3.\mathbf{y}_Q(t=T_Q)\frac{\partial \mathbf{y}_Q(t=T_Q)}{\partial \mathbf{\Lambda}}
\end{equation}
\begin{equation}
\label{eq:LSTM4}
\mathbf{F} = -a.b.c^3.\mathbf{y}_D(t=T_D)\frac{\partial \mathbf{y}_D(t=T_D)}{\partial \mathbf{\Lambda}}
\end{equation}
Therefore
\begin{align}
\label{eq:LSTM5}
&\frac{\partial R(Q,D)}{\partial\mathbf{\Lambda}} =\mathbf{D} + \mathbf{E} + \mathbf{F} = \nn\\
&\mathbf{v}_Q\frac{\partial \mathbf{y}_Q(t=T_Q)}{\partial \mathbf{\Lambda}} + \mathbf{v}_D\frac{\partial \mathbf{y}_D(t=T_D)}{\partial \mathbf{\Lambda}}
\end{align}
where
\begin{align}
\label{eq:LSTM6}
&\mathbf{v}_Q = (b.c.\mathbf{y}_D(t=T_D) - a.b^3.c.\mathbf{y}_Q(t=T_Q))\nn\\
&\mathbf{v}_D = (b.c.\mathbf{y}_Q(t=T_Q) - a.b.c^3.\mathbf{y}_D(t=T_D))
\end{align}
\subsubsection{Output Gate}
\label{sec:Goutgate}
Since $\mathbf{\alpha}\circ\mathbf{\beta} = diag(\mathbf{\alpha})\mathbf{\beta} = diag(\mathbf{\beta})\mathbf{\alpha}$ where $diag(\mathbf{\alpha})$ is a diagonal matrix whose main diagonal entries are entries of vector $\mathbf{\alpha}$, we have:
\begin{align}
\label{eq:LSTM7}
&\frac{\partial \mathbf{y}(t)}{\partial \mathbf{W}_{rec1}} = \frac{\partial}{\partial \mathbf{W}_{rec1}}(diag(h(\mathbf{c}(t))).\mathbf{o}(t))\nn\\
& = \underbrace{\frac{\partial diag(h(\mathbf{c}(t)))}{\partial \mathbf{W}_{rec1}}}_{zero}.\mathbf{o}(t) + diag(h(\mathbf{c}(t))).\frac{\partial \mathbf{o}(t)}{\partial \mathbf{W}_{rec1}}\nn\\
& = \mathbf{o}(t)\circ(1-\mathbf{o}(t))\circ h(\mathbf{c}(t)).\mathbf{y}(t-1)^T
\end{align}
Substituting \eqref{eq:LSTM7} in \eqref{eq:LSTM5} we have:
\begin{equation}
\label{eq:LSTM8}
\frac{\partial R(Q,D)}{\partial\mathbf{W}_{rec1}} = \mathbf{\delta}_{y_Q}^{rec1}(t).\mathbf{y}_Q(t-1)^T + \mathbf{\delta}_{y_D}^{rec1}(t).\mathbf{y}_D(t-1)^T
\end{equation}
where
\begin{align}
\label{eq:LSTM9}
&\mathbf{\delta}_{y_Q}^{rec1}(t) = \mathbf{o}_Q(t)\circ(1-\mathbf{o}_Q(t))\circ h(\mathbf{c}_Q(t))\circ \mathbf{v}_Q(t)\nn\\
&\mathbf{\delta}_{y_D}^{rec1}(t) = \mathbf{o}_D(t)\circ(1-\mathbf{o}_D(t))\circ h(\mathbf{c}_D(t))\circ \mathbf{v}_D(t)
\end{align}
with a similar derivation for $\mathbf{W}_1$ and $\mathbf{W}_{p1}$ we get:
\begin{equation}
\label{eq:LSTM10}
\frac{\partial R(Q,D)}{\partial\mathbf{W}_1} = \mathbf{\delta}_{y_Q}^{rec1}(t).\mathbf{l}_Q(t)^T + \mathbf{\delta}_{y_D}^{rec1}(t).\mathbf{l}_D(t)^T
\end{equation}
\begin{equation}
\label{eq:LSTM11}
\frac{\partial R(Q,D)}{\partial\mathbf{W}_{p1}} = \mathbf{\delta}_{y_Q}^{rec1}(t).\mathbf{c}_Q(t)^T + \mathbf{\delta}_{y_D}^{rec1}(t).\mathbf{c}_D(t)^T
\end{equation}

For output gate bias values we have:
\begin{equation}
\label{eq:LSTM11Bias}
\frac{\partial R(Q,D)}{\partial\mathbf{b}_{1}} = \mathbf{\delta}_{y_Q}^{rec1}(t) + \mathbf{\delta}_{y_D}^{rec1}(t)
\end{equation}
\subsubsection{Input Gate}
\label{sec:Gigate}
Similar to output gate we start with:
\begin{align}
\label{eq:LSTM12}
&\frac{\partial \mathbf{y}(t)}{\partial \mathbf{W}_{rec3}} = \frac{\partial}{\partial \mathbf{W}_{rec3}}(diag(\mathbf{o}(t)).h(\mathbf{c}(t)))\nn\\
& = \underbrace{\frac{\partial diag(\mathbf{o}(t))}{\partial \mathbf{W}_{rec3}}}_{zero}.h(\mathbf{c}(t)) + diag(\mathbf{o}(t)).\frac{\partial h(\mathbf{c}(t))}{\partial \mathbf{W}_{rec3}}\nn\\
&= diag(\mathbf{o}(t)).(1-h(\mathbf{c}(t)))\circ(1+h(\mathbf{c}(t)))\frac{\partial \mathbf{c}(t)}{\partial \mathbf{W}_{rec3}}
\end{align}
To find $\frac{\partial \mathbf{c}(t)}{\partial \mathbf{W}_{rec3}}$ assuming $\mathbf{f}(t)=1$ (we derive formulation for $\mathbf{f}(t)\neq 1$ from this simple solution) we have: 
\begin{align}
\label{eq:LSTM13}
&\mathbf{c}(0) = 0\nn\\
&\mathbf{c}(1) = \mathbf{c}(0) + \mathbf{i}(1)\circ \mathbf{y}_g(1) = \mathbf{i}(1)\circ \mathbf{y}_g(1)\nn\\
&\mathbf{c}(2) = \mathbf{c}(1) + \mathbf{i}(2)\circ \mathbf{y}_g(2)\nn\\
& \dots\nn\\
&\mathbf{c}(t) = \sum_{k=1}^{t} \mathbf{i}(k)\circ \mathbf{y}_g(k) = \sum_{k=1}^{t} diag(\mathbf{y}_g(k)).\mathbf{i}(k)
\end{align}
Therefore
\begin{align}
\label{eq:LSTM14}
&\frac{\partial \mathbf{c}(t)}{\partial \mathbf{W}_{rec3}} = \sum_{k=1}^{t} [\underbrace{\frac{\partial diag(\mathbf{y}_g(k))}{\mathbf{W}_{rec3}}}_{zero}.\mathbf{i}(k) + diag(\mathbf{y}_g(k)).\frac{\partial \mathbf{i}(k)}{\mathbf{W}_{rec3}}]\nn\\
&= \sum_{k=1}^{t} diag(\mathbf{y}_g(k)).\mathbf{i}(k)\circ (1-\mathbf{i}(k)).\mathbf{y}(k-1)^T\\
\end{align}
and
\begin{align}
\label{eq:LSTM15}
&\frac{\partial \mathbf{y}(t)}{\partial \mathbf{W}_{rec3}} = \sum_{k=1}^{t} [\underbrace{\mathbf{o}(t)\circ (1-h(\mathbf{c}(t)))\circ (1+h(\mathbf{c}(t)))}_{\mathbf{a}(t)}\nn\\
&\circ \underbrace{\mathbf{y}_g(k)\circ \mathbf{i}(k)\circ (1-\mathbf{i}(k))}_{\mathbf{b}(k)}]\mathbf{y}(k-1)^T
\end{align}
But this is expensive to implement, to resolve it we have:
\begin{align}
\label{eq:LSTM16}
&\frac{\partial \mathbf{y}(t)}{\partial \mathbf{W}_{rec3}} = \underbrace{\sum_{k=1}^{t-1} [\mathbf{a}(t)\circ \mathbf{b}(k)]\mathbf{y}(k-1)^T}_{\mathrm{expensive \; part}}\nn\\
&+ [\mathbf{a}(t)\circ \mathbf{b}(t)]\mathbf{y}(t-1)^T\nn\\
& = diag(\mathbf{a}(t))\underbrace{\sum_{k=1}^{t-1}\mathbf{b}(k).\mathbf{y}(k-1)^T}_{\frac{\partial \mathbf{c}(t-1)}{\partial \mathbf{W}_{rec3}}}\nn\\
&+ diag(\mathbf{a}(t)).\mathbf{b}(t).\mathbf{y}(t-1)^T
\end{align}
Therefore
\begin{equation}
\label{eq:LSTM17}
\frac{\partial \mathbf{y}(t)}{\partial \mathbf{W}_{rec3}} = [diag(\mathbf{a}(t))][\frac{\partial \mathbf{c}(t-1)}{\partial \mathbf{W}_{rec3}} + \mathbf{b}(t).\mathbf{y}(t-1)^T]
\end{equation}
For $\mathbf{f}(t)\neq 1$ we have
\begin{align}
\label{eq:LSTM18}
&\frac{\partial \mathbf{y}(t)}{\partial \mathbf{W}_{rec3}} = [diag(\mathbf{a}(t))][diag(\mathbf{f}(t)).\frac{\partial \mathbf{c}(t-1)}{\partial \mathbf{W}_{rec3}} \nn\\
&+ \mathbf{b}_i(t).\mathbf{y}(t-1)^T]
\end{align}
where
\begin{align}
\label{eq:LSTM19}
&\mathbf{a}(t) = \mathbf{o}(t)\circ (1-h(\mathbf{c}(t)))\circ (1+h(\mathbf{c}(t)))\nn\\
&\mathbf{b}_i(t) = \mathbf{y}_g(t)\circ \mathbf{i}(t)\circ (1-\mathbf{i}(t))
\end{align}
substituting above equation in \eqref{eq:LSTM5} we will have:
\begin{align}
\label{eq:LSTM20}
&\frac{\partial R(Q,D)}{\partial \mathbf{W}_{rec3}} = diag(\mathbf{\delta}_{y_Q}^{rec3}(t)).\frac{\partial \mathbf{c}_Q(t)}{\partial \mathbf{W}_{rec3}}\nn\\
&+ diag(\mathbf{\delta}_{y_D}^{rec3}(t)).\frac{\partial \mathbf{c}_D(t)}{\partial \mathbf{W}_{rec3}}
\end{align}
where
\begin{align}
\label{eq:LSTM21}
&\mathbf{\delta}_{y_Q}^{rec3}(t) = (1-h(\mathbf{c}_Q(t)))\circ (1+h(\mathbf{c}_Q(t)))\circ \mathbf{o}_Q(t) \circ \mathbf{v}_Q(t)\nn\\
&\frac{\partial \mathbf{c}_Q(t)}{\partial \mathbf{W}_{rec3}} = diag(\mathbf{f}_Q(t)).\frac{\partial \mathbf{c}_Q(t-1)}{\partial \mathbf{W}_{rec3}} + \mathbf{b}_{i,Q}(t).\mathbf{y}_Q(t-1)^T\nn\\
&\mathbf{b}_{i,Q}(t) = \mathbf{y}_{g,Q}(t)\circ \mathbf{i}_Q(t) \circ (1-\mathbf{i}_Q(t))
\end{align}
In equation \eqref{eq:LSTM20},  $\mathbf{\delta}_{y_D}^{rec3}(t)$ and $\frac{\partial \mathbf{c}_D(t)}{\partial \mathbf{W}_{rec3}} $ are the same as \eqref{eq:LSTM21} with $D$ subscript. Therefore, update equations for $\mathbf{W}_{rec3}$ are \eqref{eq:LSTM20}, \eqref{eq:LSTM21} for $Q$ and $D$ and (6). 

With a similar procedure for $\mathbf{W}_3$ we will have the following:
\begin{align}
\label{eq:LSTM22}
&\frac{\partial R(Q,D)}{\partial \mathbf{W}_{3}} = diag(\mathbf{\delta}_{y_Q}^{rec3}(t)).\frac{\partial \mathbf{c}_Q(t)}{\partial \mathbf{W}_{3}}\nn\\
&+ diag(\mathbf{\delta}_{y_D}^{rec3}(t)).\frac{\partial \mathbf{c}_D(t)}{\partial \mathbf{W}_{3}}
\end{align}
where
\begin{equation}
\label{eq:LSTM23}
\frac{\partial \mathbf{c}_Q(t)}{\partial \mathbf{W}_{3}} = diag(\mathbf{f}_Q(t)).\frac{\partial \mathbf{c}_Q(t-1)}{\partial \mathbf{W}_{3}} + \mathbf{b}_{i,Q}(t).\mathbf{x}_Q(t)^T
\end{equation}
Therefore, update equations for $\mathbf{W}_{3}$ are \eqref{eq:LSTM22}, \eqref{eq:LSTM23} for $Q$ and $D$ and (6). 

For peephole connections we will have:
\begin{align}
\label{eq:LSTM24}
&\frac{\partial R(Q,D)}{\partial \mathbf{W}_{p3}} = diag(\mathbf{\delta}_{y_Q}^{rec3}(t)).\frac{\partial \mathbf{c}_Q(t)}{\partial \mathbf{W}_{p3}}\nn\\
&+ diag(\mathbf{\delta}_{y_D}^{rec3}(t)).\frac{\partial \mathbf{c}_D(t)}{\partial \mathbf{W}_{p3}}
\end{align}
where
\begin{equation}
\label{eq:LSTM25}
\frac{\partial \mathbf{c}_Q(t)}{\partial \mathbf{W}_{p3}} = diag(\mathbf{f}_Q(t)).\frac{\partial \mathbf{c}_Q(t-1)}{\partial \mathbf{W}_{p3}} + \mathbf{b}_{i,Q}(t).\mathbf{c}_Q(t-1)^T
\end{equation}
Hence, update equations for $\mathbf{W}_{p3}$ are \eqref{eq:LSTM24}, \eqref{eq:LSTM25} for $Q$ and $D$ and  (6).

Following similar derivation for bias values $\mathbf{b}_3$ we will have:
\begin{align}
\label{eq:LSTM26}
&\frac{\partial R(Q,D)}{\partial \mathbf{b}_{3}} = diag(\mathbf{\delta}_{y_Q}^{rec3}(t)).\frac{\partial \mathbf{c}_Q(t)}{\partial \mathbf{b}_{3}}\nn\\
&+ diag(\mathbf{\delta}_{y_D}^{rec3}(t)).\frac{\partial \mathbf{c}_D(t)}{\partial \mathbf{b}_{3}}
\end{align}
where
\begin{equation}
\label{eq:LSTM27}
\frac{\partial \mathbf{c}_Q(t)}{\partial \mathbf{b}_{3}} = diag(\mathbf{f}_Q(t)).\frac{\partial \mathbf{c}_Q(t-1)}{\partial \mathbf{b}_{3}} + \mathbf{b}_{i,Q}(t)
\end{equation}
Update equations for $\mathbf{b}_{3}$ are \eqref{eq:LSTM26}, \eqref{eq:LSTM27} for $Q$ and $D$ and  (6).

\subsubsection{Forget Gate}
\label{sec:Gfgate}
For forget gate, with a similar derivation to input gate we will have
\begin{align}
\label{eq:LSTM28}
&\frac{\partial \mathbf{y}(t)}{\partial \mathbf{W}_{rec2}} = [diag(\mathbf{a}(t))][diag(\mathbf{f}(t)).\frac{\partial \mathbf{c}(t-1)}{\partial \mathbf{W}_{rec2}} \nn\\
&+ \mathbf{b}_f(t).\mathbf{y}(t-1)^T]
\end{align}
where
\begin{align}
\label{eq:LSTM29}
&\mathbf{a}(t) = \mathbf{o}(t)\circ (1-h(\mathbf{c}(t)))\circ (1+h(\mathbf{c}(t)))\nn\\
&\mathbf{b}_f(t) = \mathbf{c}(t-1)\circ \mathbf{f}(t)\circ (1-\mathbf{f}(t))
\end{align}
substituting above equation in \eqref{eq:LSTM5} we will have:
\begin{align}
\label{eq:LSTM30}
&\frac{\partial R(Q,D)}{\partial \mathbf{W}_{rec2}} = diag(\mathbf{\delta}_{y_Q}^{rec2}(t)).\frac{\partial \mathbf{c}_Q(t)}{\partial \mathbf{W}_{rec2}}\nn\\
&+ diag(\mathbf{\delta}_{y_D}^{rec2}(t)).\frac{\partial \mathbf{c}_D(t)}{\partial \mathbf{W}_{rec2}}
\end{align}
where
\begin{align}
\label{eq:LSTM31}
&\mathbf{\delta}_{y_Q}^{rec2}(t) = (1-h(\mathbf{c}_Q(t)))\circ (1+h(\mathbf{c}_Q(t)))\circ \mathbf{o}_Q(t) \circ \mathbf{v}_Q(t)\nn\\
&\frac{\partial \mathbf{c}_Q(t)}{\partial \mathbf{W}_{rec2}} = diag(\mathbf{f}_Q(t)).\frac{\partial \mathbf{c}_Q(t-1)}{\partial \mathbf{W}_{rec2}} + \mathbf{b}_{f,Q}(t).\mathbf{y}_Q(t-1)^T\nn\\
&\mathbf{b}_{f,Q}(t) = \mathbf{c}_Q(t-1)\circ \mathbf{f}_Q(t) \circ (1-\mathbf{f}_Q(t))
\end{align}
Therefore, update equations for $\mathbf{W}_{rec2}$ are \eqref{eq:LSTM30}, \eqref{eq:LSTM31} for $Q$ and $D$ and (6).

For input weights to forget gate, $\mathbf{W}_2$, we have
\begin{align}
\label{eq:LSTM32}
&\frac{\partial R(Q,D)}{\partial \mathbf{W}_{2}} = diag(\mathbf{\delta}_{y_Q}^{rec2}(t)).\frac{\partial \mathbf{c}_Q(t)}{\partial \mathbf{W}_{2}}\nn\\
&+ diag(\mathbf{\delta}_{y_D}^{rec2}(t)).\frac{\partial \mathbf{c}_D(t)}{\partial \mathbf{W}_{2}}
\end{align}
where
\begin{equation}
\label{eq:LSTM33}
\frac{\partial \mathbf{c}_Q(t)}{\partial \mathbf{W}_{2}} = diag(\mathbf{f}_Q(t)).\frac{\partial \mathbf{c}_Q(t-1)}{\partial \mathbf{W}_{2}} + \mathbf{b}_{f,Q}(t).\mathbf{x}_Q(t)^T
\end{equation}
Therefore, update equations for $\mathbf{W}_{2}$ are \eqref{eq:LSTM32}, \eqref{eq:LSTM33} for $Q$ and $D$ and (6).

For peephole connections, $\mathbf{W}_{p2}$, we have
\begin{align}
\label{eq:LSTM34}
&\frac{\partial R(Q,D)}{\partial \mathbf{W}_{p2}} = diag(\mathbf{\delta}_{y_Q}^{rec2}(t)).\frac{\partial \mathbf{c}_Q(t)}{\partial \mathbf{W}_{p2}}\nn\\
&+ diag(\mathbf{\delta}_{y_D}^{rec2}(t)).\frac{\partial \mathbf{c}_D(t)}{\partial \mathbf{W}_{p2}}
\end{align}
where
\begin{equation}
\label{eq:LSTM35}
\frac{\partial \mathbf{c}_Q(t)}{\partial \mathbf{W}_{p2}} = diag(\mathbf{f}_Q(t)).\frac{\partial \mathbf{c}_Q(t-1)}{\partial \mathbf{W}_{p2}} + \mathbf{b}_{f,Q}(t).\mathbf{c}_Q(t-1)^T
\end{equation}
Therefore, update equations for $\mathbf{W}_{p2}$ are \eqref{eq:LSTM34}, \eqref{eq:LSTM35} for $Q$ and $D$ and (6).

Update equations for forget gate bias values, $\mathbf{b}_2$, will be following equations and (6):
\begin{align}
\label{eq:LSTM36}
&\frac{\partial R(Q,D)}{\partial \mathbf{b}_{2}} = diag(\mathbf{\delta}_{y_Q}^{rec2}(t)).\frac{\partial \mathbf{c}_Q(t)}{\partial \mathbf{b}_{2}}\nn\\
&+ diag(\mathbf{\delta}_{y_D}^{rec2}(t)).\frac{\partial \mathbf{c}_D(t)}{\partial \mathbf{b}_{2}}
\end{align}
where
\begin{equation}
\label{eq:LSTM37}
\frac{\partial \mathbf{c}_Q(t)}{\partial \mathbf{b}_{2}} = diag(\mathbf{f}_Q(t)).\frac{\partial \mathbf{c}_Q(t-1)}{\partial \mathbf{b}_{3}} + \mathbf{b}_{f,Q}(t)
\end{equation}

\subsubsection{Input without Gating ($\mathbf{y}_g(t)$)}
\label{sec:Gyg}
Gradients for $\mathbf{y}_g(t)$ parameters are as follows:
\begin{align}
\label{eq:LSTM38}
&\frac{\partial \mathbf{y}(t)}{\partial \mathbf{W}_{rec4}} = [diag(\mathbf{a}(t))][diag(\mathbf{f}(t)).\frac{\partial \mathbf{c}(t-1)}{\partial \mathbf{W}_{rec4}} \nn\\
&+ \mathbf{b}_g(t).\mathbf{y}(t-1)^T]
\end{align}
where
\begin{align}
\label{eq:LSTM39}
&\mathbf{a}(t) = \mathbf{o}(t)\circ (1-h(\mathbf{c}(t)))\circ (1+h(\mathbf{c}(t)))\nn\\
&\mathbf{b}_g(t) = \mathbf{i}(t)\circ (1-\mathbf{y}_g(t))\circ (1+\mathbf{y}_g(t))
\end{align}
substituting above equation in \eqref{eq:LSTM5} we will have:
\begin{align}
\label{eq:LSTM40}
&\frac{\partial R(Q,D)}{\partial \mathbf{W}_{rec4}} = diag(\mathbf{\delta}_{y_Q}^{rec4}(t)).\frac{\partial \mathbf{c}_Q(t)}{\partial \mathbf{W}_{rec4}}\nn\\
&+ diag(\mathbf{\delta}_{y_D}^{rec4}(t)).\frac{\partial \mathbf{c}_D(t)}{\partial \mathbf{W}_{rec4}}
\end{align}
where
\begin{align}
\label{eq:LSTM41}
&\mathbf{\delta}_{y_Q}^{rec4}(t) = (1-h(\mathbf{c}_Q(t)))\circ (1+h(\mathbf{c}_Q(t)))\circ \mathbf{o}_Q(t) \circ \mathbf{v}_Q(t)\nn\\
&\frac{\partial \mathbf{c}_Q(t)}{\partial \mathbf{W}_{rec4}} = diag(\mathbf{f}_Q(t)).\frac{\partial \mathbf{c}_Q(t-1)}{\partial \mathbf{W}_{rec4}} + \mathbf{b}_{g,Q}(t).\mathbf{y}_Q(t-1)^T\nn\\
&\mathbf{b}_{g,Q}(t) = \mathbf{i}_Q(t) \circ (1-\mathbf{y}_{g,Q}(t))\circ (1+\mathbf{y}_{g,Q}(t))
\end{align}
Therefore, update equations for $\mathbf{W}_{rec4}$ are \eqref{eq:LSTM40}, \eqref{eq:LSTM41} for $Q$ and $D$ and (6).

For input weight parameters, $\mathbf{W}_4$, we have
\begin{align}
\label{eq:LSTM42}
&\frac{\partial R(Q,D)}{\partial \mathbf{W}_{4}} = diag(\mathbf{\delta}_{y_Q}^{rec4}(t)).\frac{\partial \mathbf{c}_Q(t)}{\partial \mathbf{W}_{4}}\nn\\
&+ diag(\mathbf{\delta}_{y_D}^{rec4}(t)).\frac{\partial \mathbf{c}_D(t)}{\partial \mathbf{W}_{4}}
\end{align}
where
\begin{equation}
\label{eq:LSTM43}
\frac{\partial \mathbf{c}_Q(t)}{\partial \mathbf{W}_{4}} = diag(\mathbf{f}_Q(t)).\frac{\partial \mathbf{c}_Q(t-1)}{\partial \mathbf{W}_{4}} + \mathbf{b}_{g,Q}(t).\mathbf{x}_Q(t)^T
\end{equation}
Therefore, update equations for $\mathbf{W}_{4}$ are \eqref{eq:LSTM42}, \eqref{eq:LSTM43} for $Q$ and $D$ and (6).

Gradients with respect to bias values, $\mathbf{b}_4$, are
\begin{align}
\label{eq:LSTM44}
&\frac{\partial R(Q,D)}{\partial \mathbf{b}_{4}} = diag(\mathbf{\delta}_{y_Q}^{rec4}(t)).\frac{\partial \mathbf{c}_Q(t)}{\partial \mathbf{b}_{4}}\nn\\
&+ diag(\mathbf{\delta}_{y_D}^{rec4}(t)).\frac{\partial \mathbf{c}_D(t)}{\partial \mathbf{b}_{4}}
\end{align}
where
\begin{equation}
\label{eq:LSTM45}
\frac{\partial \mathbf{c}_Q(t)}{\partial \mathbf{b}_{4}} = diag(\mathbf{f}_Q(t)).\frac{\partial \mathbf{c}_Q(t-1)}{\partial \mathbf{b}_{4}} + \mathbf{b}_{g,Q}(t)
\end{equation}
Therefore, update equations for $\mathbf{b}_{4}$ are \eqref{eq:LSTM44}, \eqref{eq:LSTM45} for $Q$ and $D$ and (6). There is no peephole connections for $\mathbf{y}_g(t)$. 

\section{LSTM-RNN Visualization}
\label{sec:app:lstm_vis}
In this appendix we present more examples of LSTM-RNN visualization.
\subsection{LSTM-RNN Semantic Vectors: Another Example}
\label{sec:app:eg2}
\begin{figure*}[t]
\centerline{
\subfigure[$\mathbf{i}(t)$]{\includegraphics[width = 0.4\textwidth]{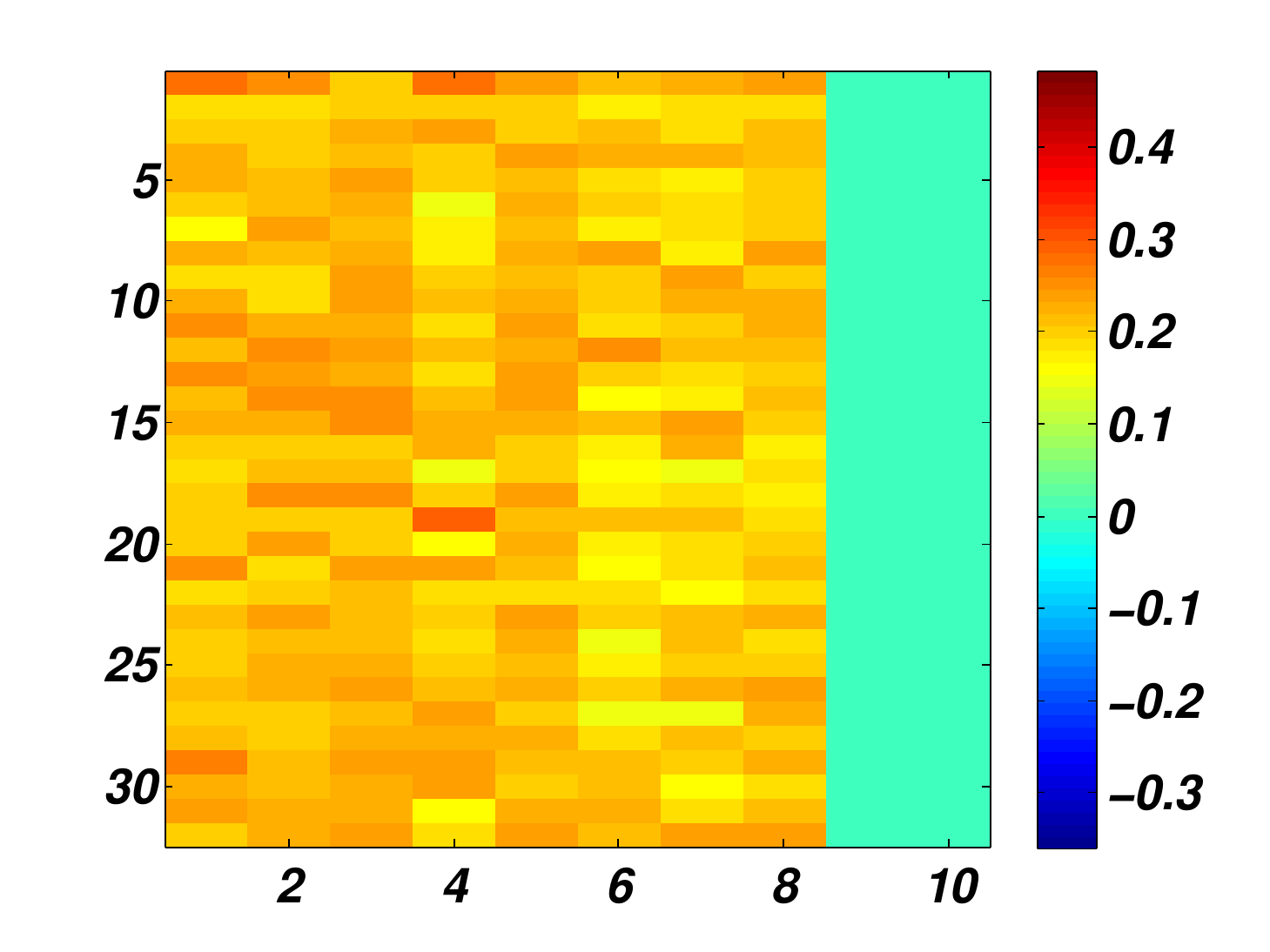}}
\subfigure[$\mathbf{c}(t)$]{\includegraphics[width = 0.4\textwidth]{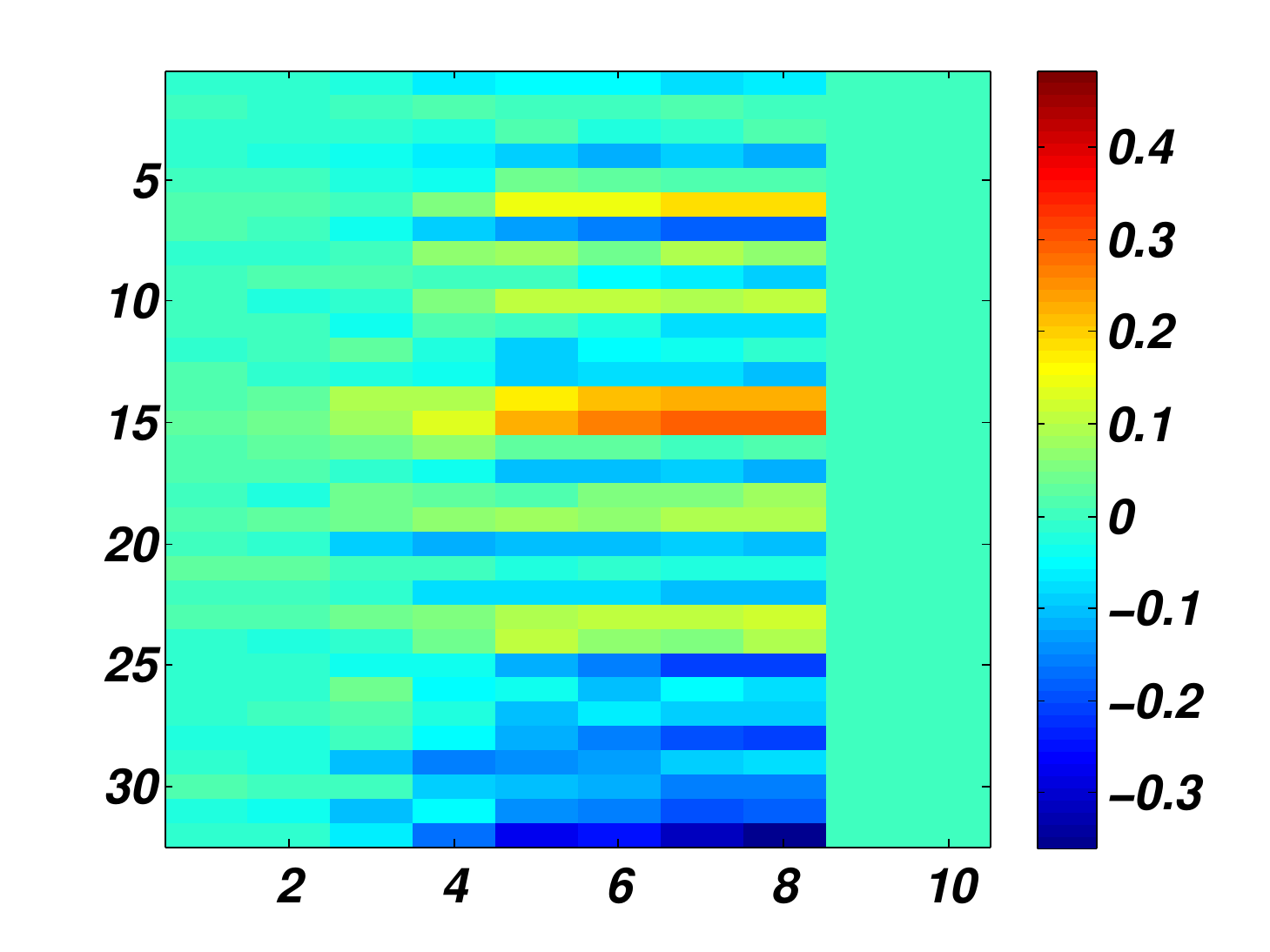}}
}
\centerline{
\subfigure[$\mathbf{o}(t)$]{\includegraphics[width = 0.4\textwidth]{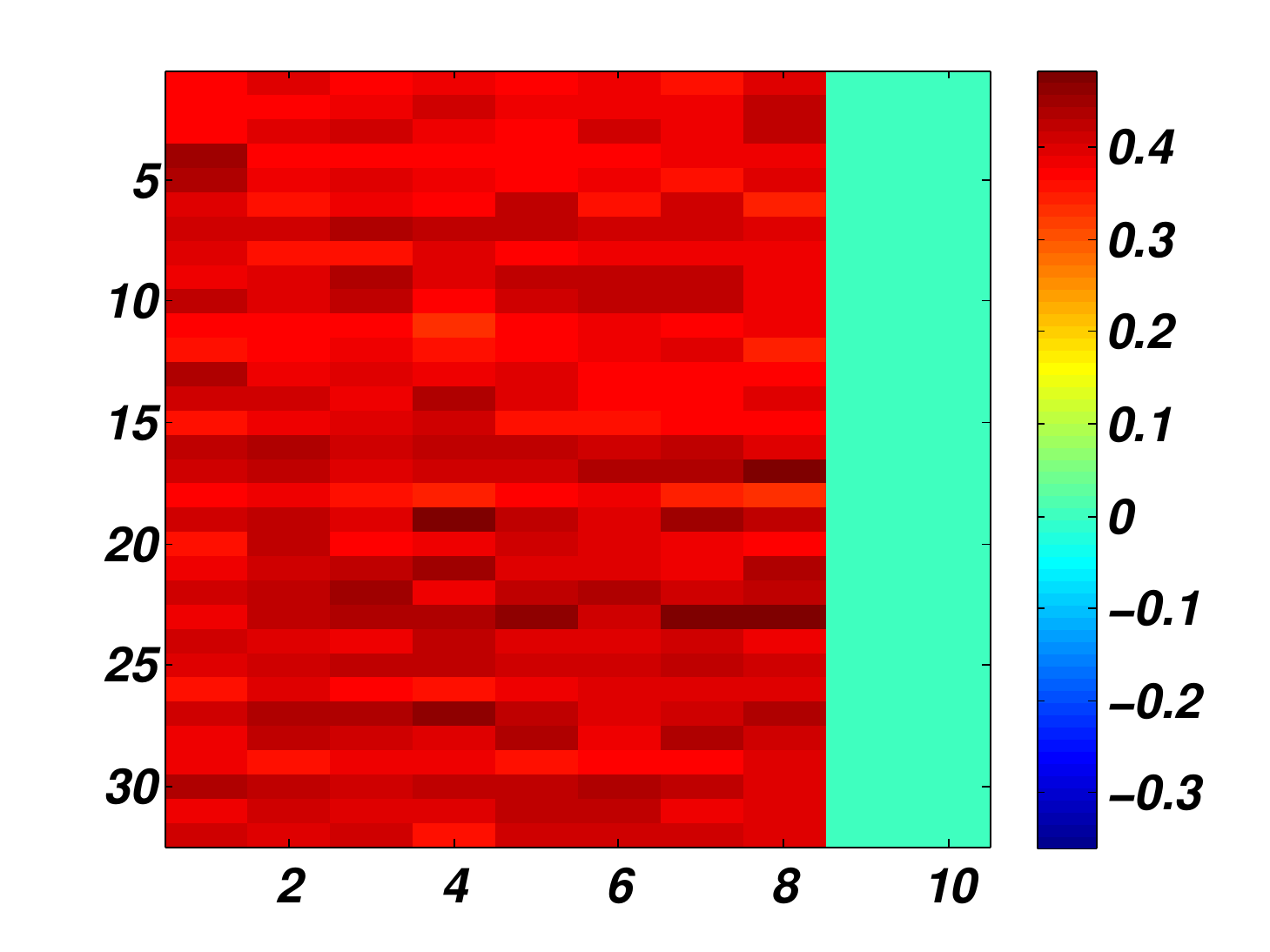}}
\subfigure[$\mathbf{y}(t)$]{\includegraphics[width = 0.4\textwidth]{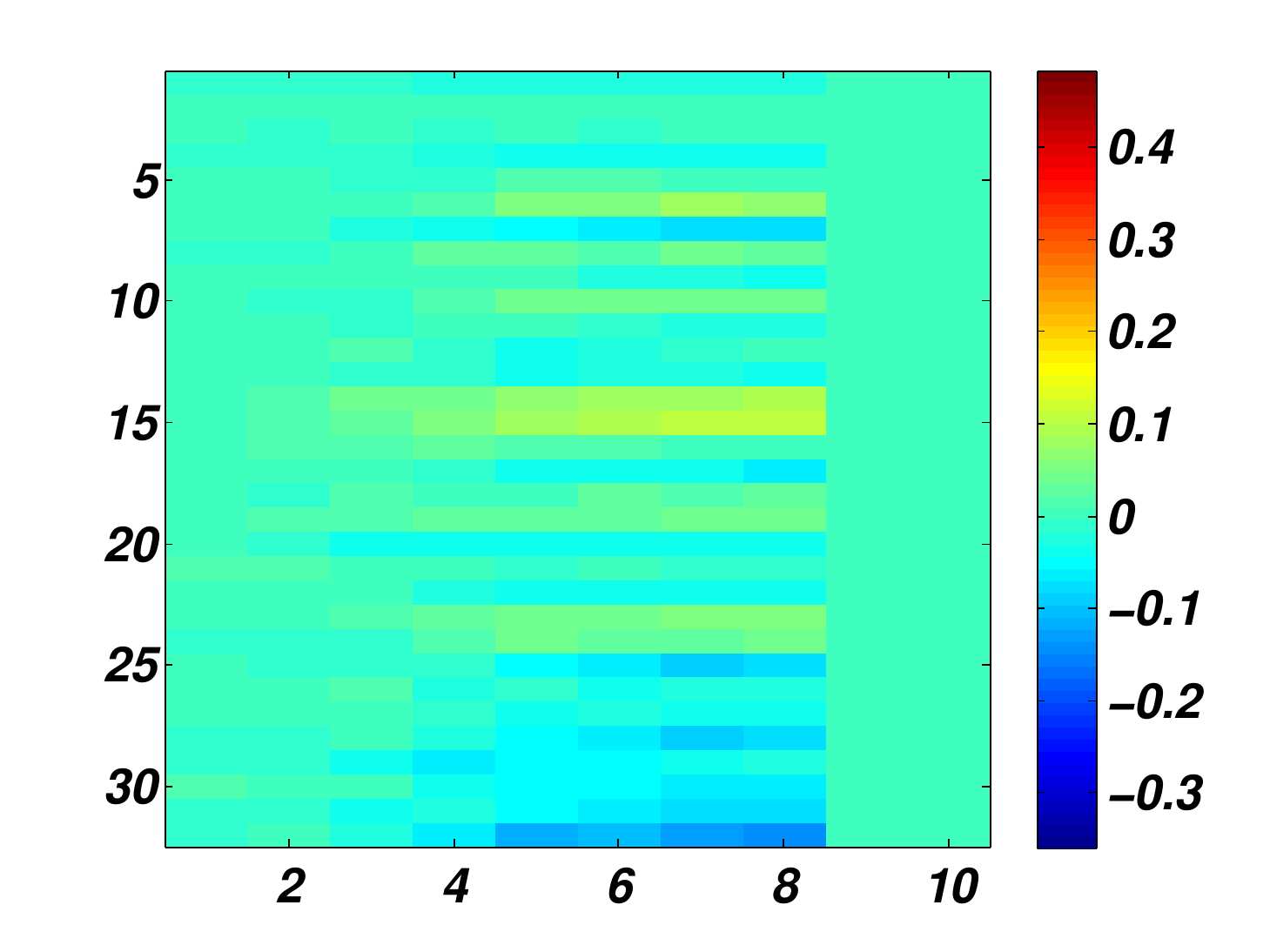}}
}

\caption{Query: ``\emph{how to  fix bath tub wont turn off}''}
\label{fig:BathQ}
\end{figure*}

\begin{figure*}[t]
\centerline{
\subfigure[$\mathbf{i}(t)$]{\includegraphics[width = 0.4\textwidth]{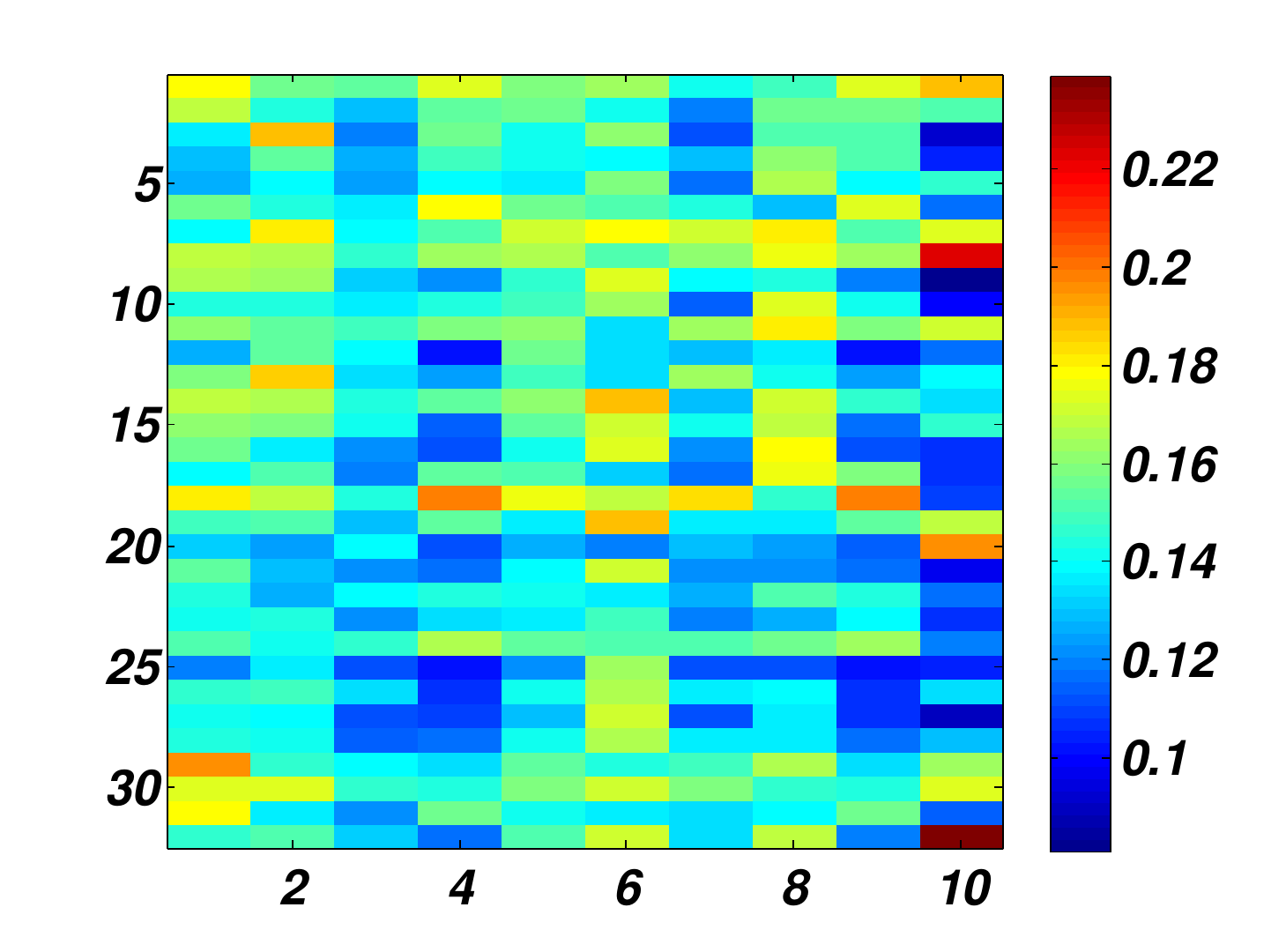}}
\subfigure[$\mathbf{c}(t)$]{\includegraphics[width = 0.4\textwidth]{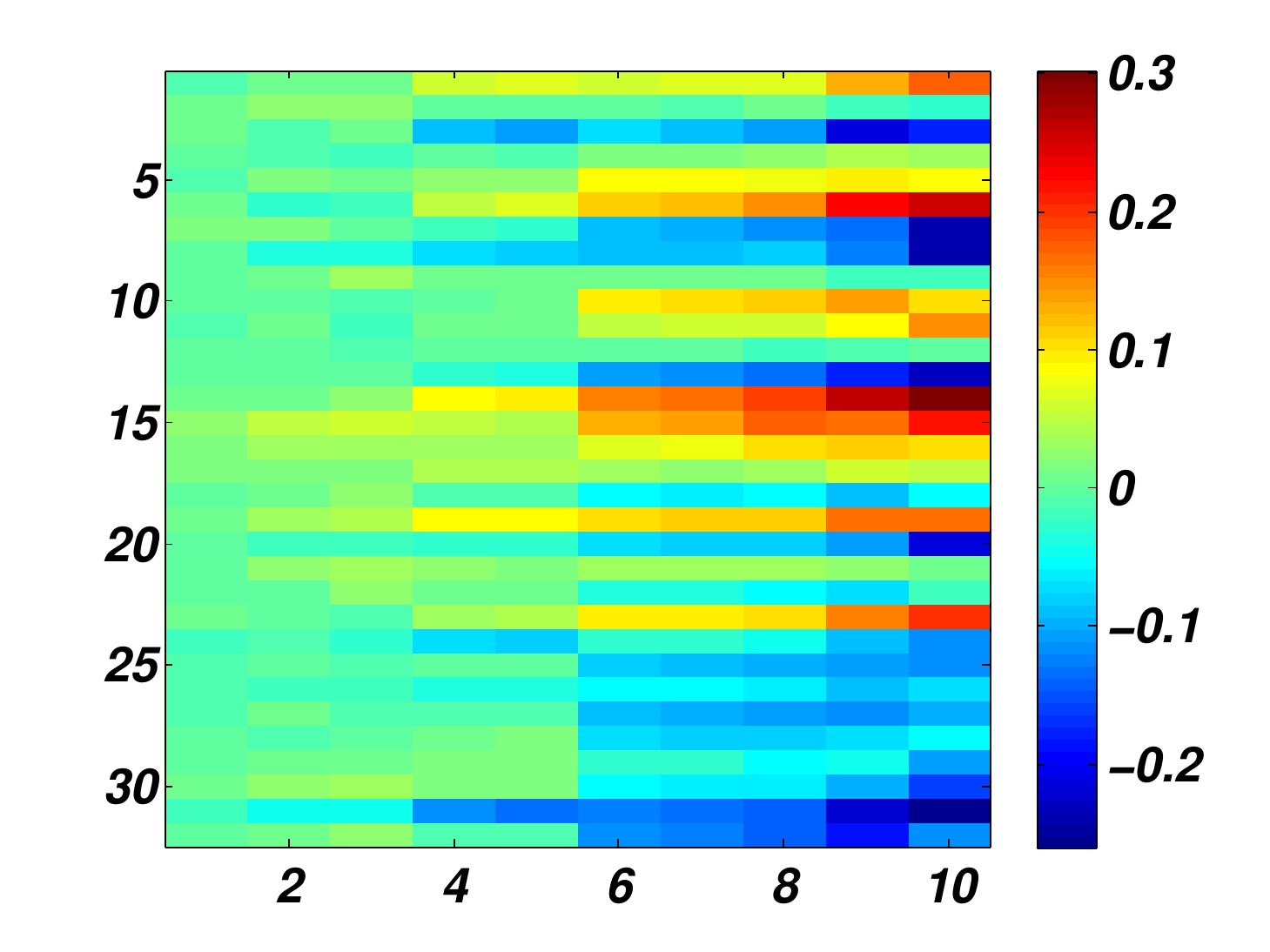}}
}
\centerline{
\subfigure[$\mathbf{o}(t)$]{\includegraphics[width = 0.4\textwidth]{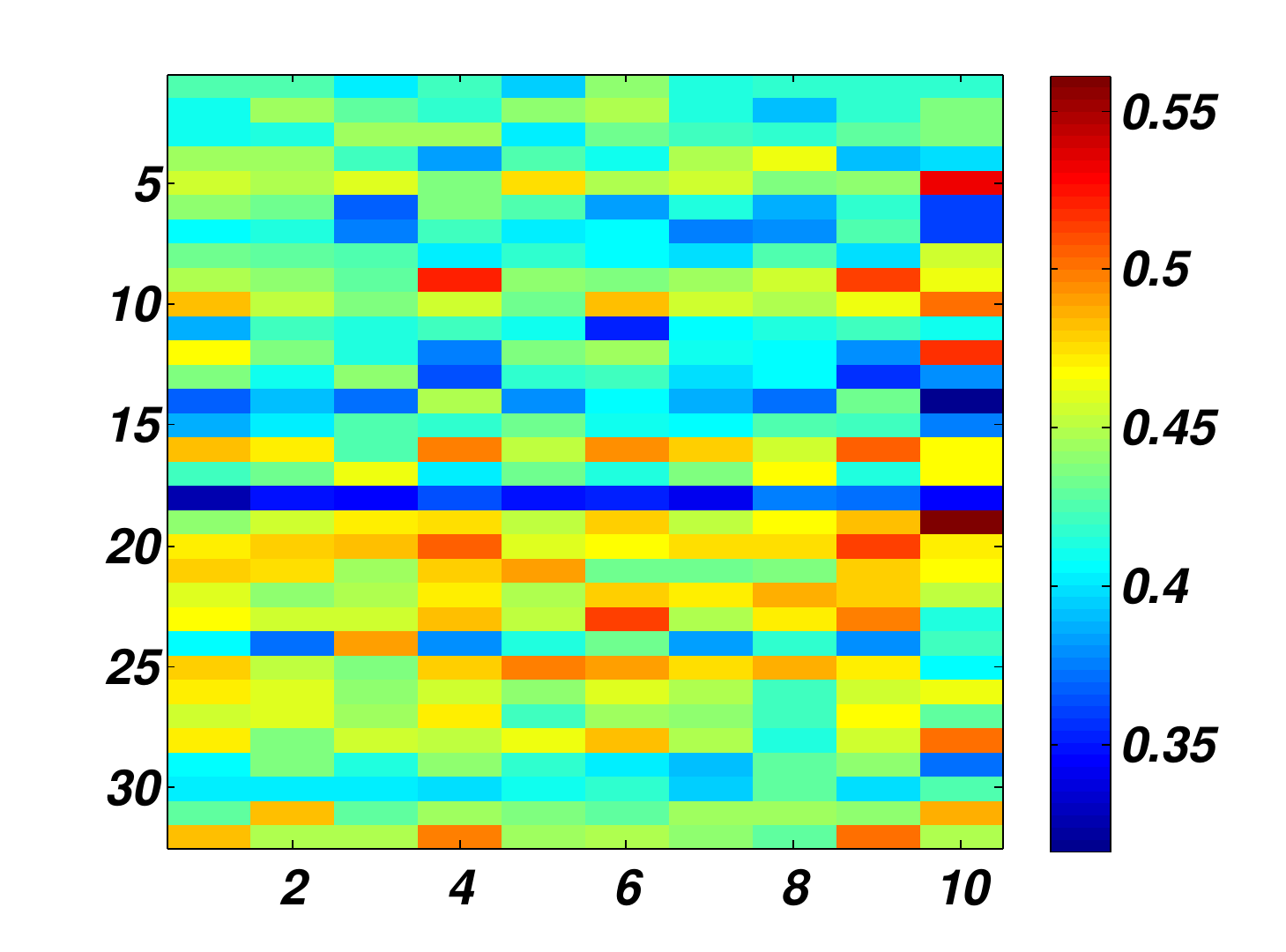}}
\subfigure[$\mathbf{y}(t)$]{\includegraphics[width = 0.4\textwidth]{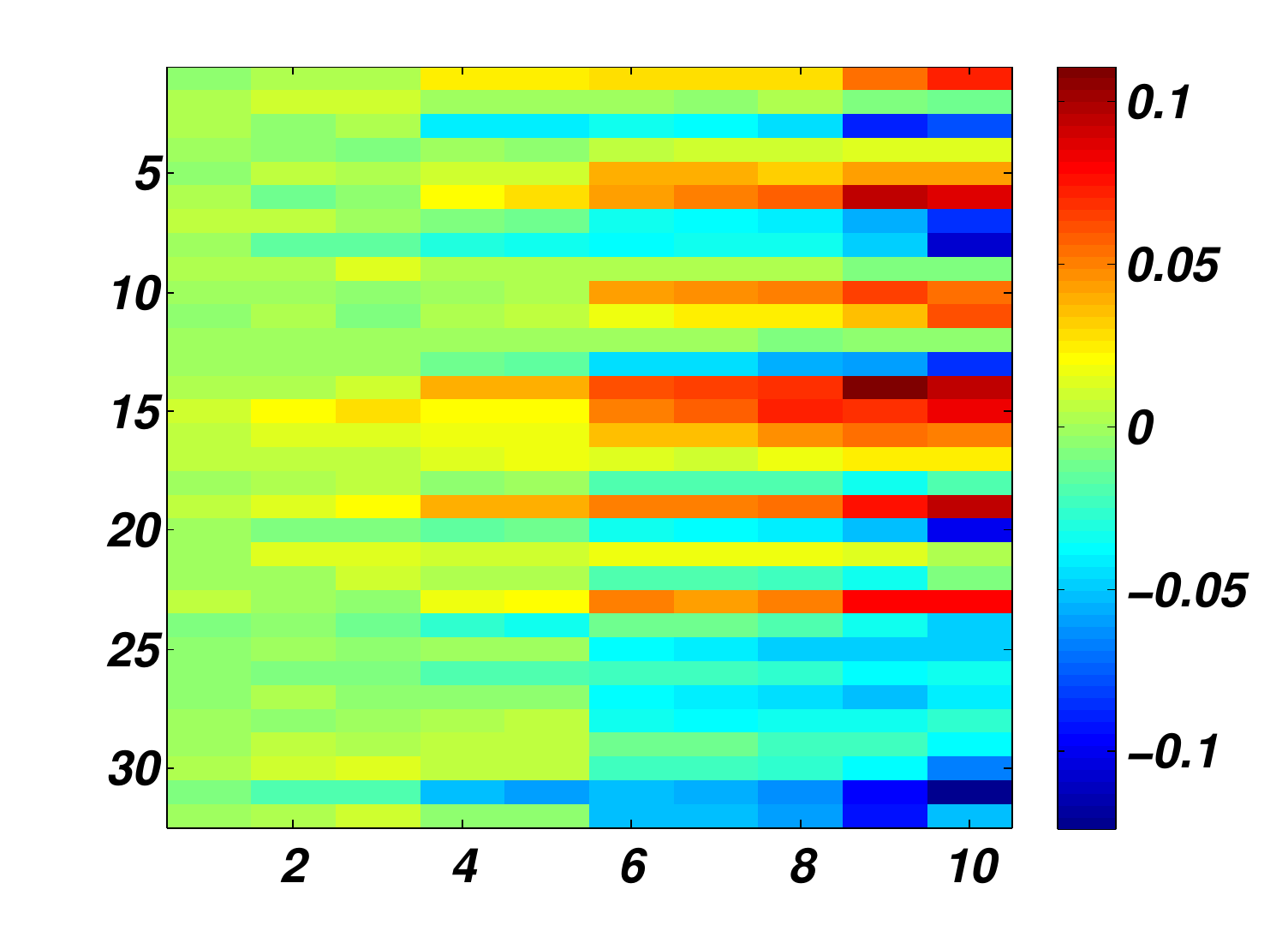}}
}
\caption{Document: ``\emph{how do you paint a bathtub and what paint should \ldots}''}
\label{fig:BathD}
\end{figure*}

Consider the following example from test dataset:
\begin{itemize}
\item Query: ``$how\; to \; fix\; bath\; tub\; wont\; turn\; off$''
\item Document: ``$how\; do\; you\; paint\; a\; bathtub\; and\; what$ \\$paint\; \underbrace{should\; you\; use\; yahoo\; answers}_{\mathrm{treated\; as\; one\; word}}$''
\end{itemize}
Activations of input gate, output gate, cell state and cell output for each cell for query and document are presented in Fig.\ref{fig:BathQ} and Fig.\ref{fig:BathD} respectively based on a trained LSTM-RNN model. 

Three interesting observations from Fig.\ref{fig:BathQ} and Fig.\ref{fig:BathD}:
\begin{itemize}
\item Semantic representation $\mathbf{y}(t)$ and cell states $\mathbf{c}(t)$ are evolving over time.
\item In part (a) of Fig.\ref{fig:BathD}, we observe that input gate values for most of the cells corresponding to word $3$, word $4$, word $7$ and word $9$ in document side of LSTM-RNN have very small values (light blue color). These are corresponding to words ``$you$'', ``$paint$'', ``$and$'' and ``$paint$'' respectively in the document title. Interestingly, input gates are trying to reduce effect of these words in the final representation ($\mathbf{y}(t)$) because the LSTM-RNN model is trained to maximize the similarity between query and document if they are a good match. 
\item $\mathbf{y}(t)$ is used as semantic representation after applying output gate on cell states. Note that valuable context information is stored in cell states $\mathbf{c}(t)$.
\end{itemize}
\subsection{Key Word Extraction: Another Example}
\label{sec:appKeyWord}
\begin{figure*}[t]
\center
\includegraphics[width=0.45\textwidth]{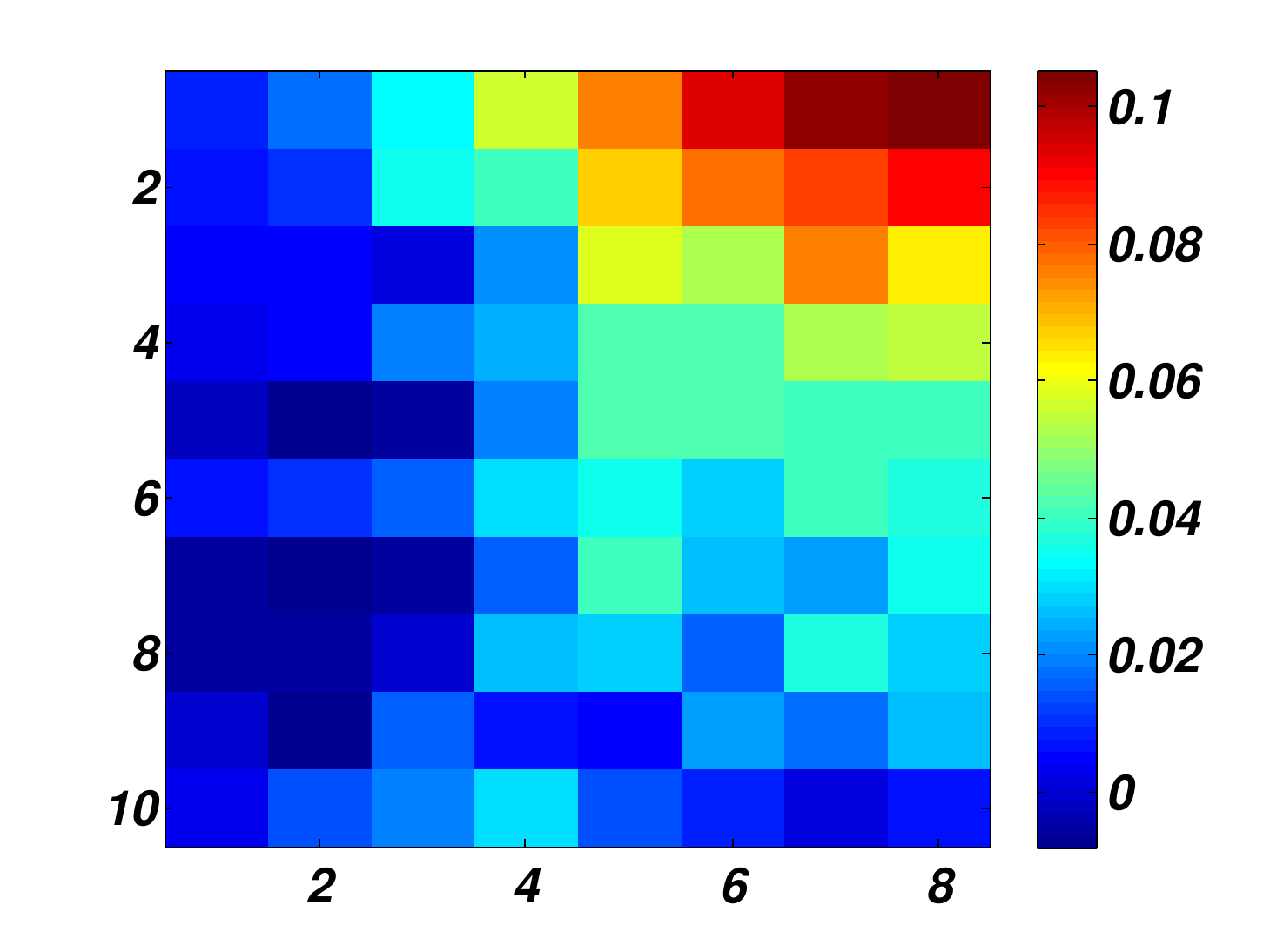}
\caption{Query: ``\emph{how to  fix bath tub wont turn off}''}
\label{fig:KeyWordBathQ}
\end{figure*}

\begin{figure*}[t]
\center
\includegraphics[width = 0.45\textwidth]{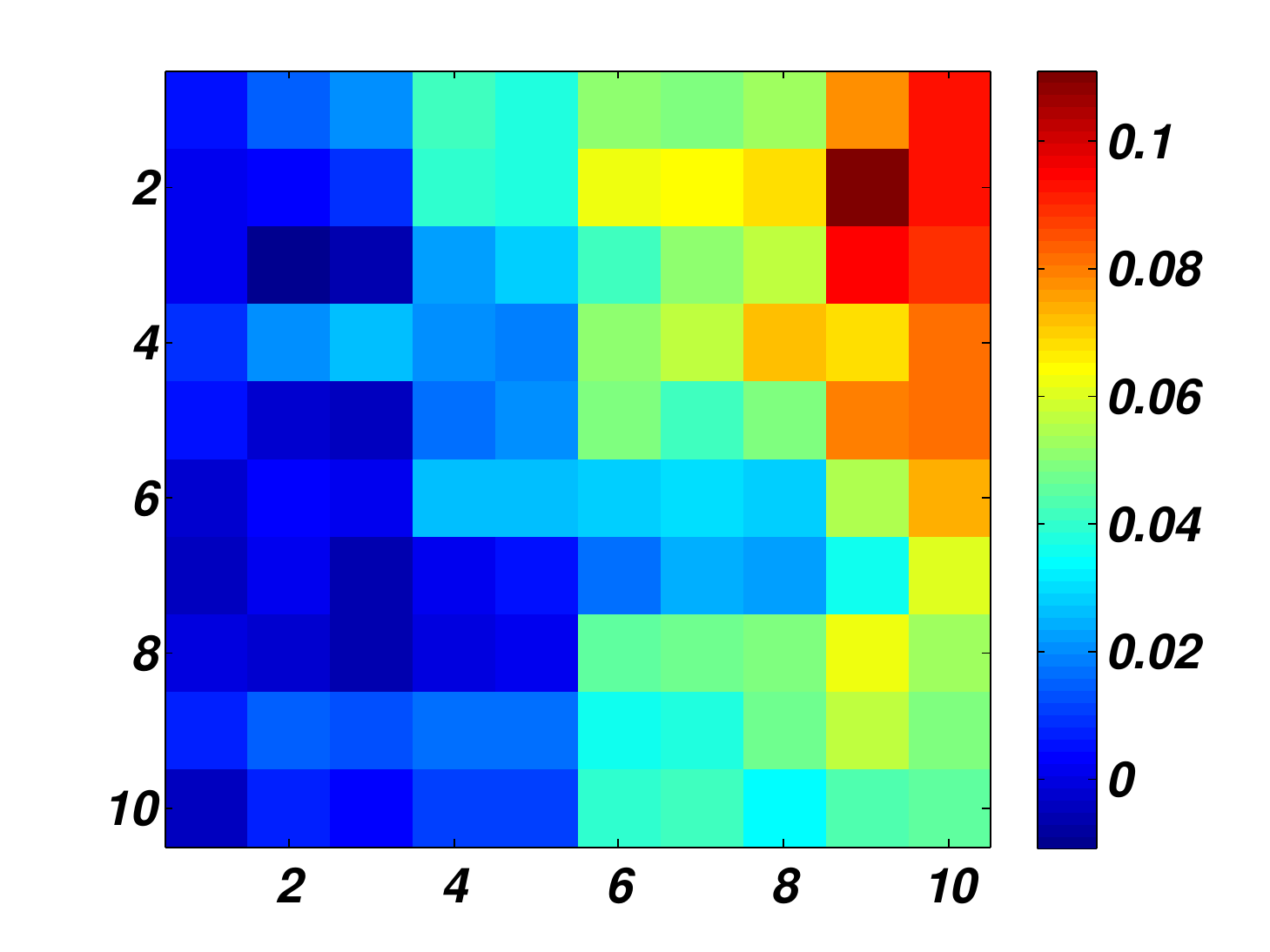}
\caption{Document: ``\emph{how do you paint a bathtub and what paint should \ldots}''}
\label{fig:KeyWordBathD}
\end{figure*}

Evolution of 10 most active cells over time for the second example are presented in Fig. \ref{fig:KeyWordBathQ} for query and Fig. \ref{fig:KeyWordBathD} for document. 
Number of assigned cells out of 10 most active cells to each word are presented in Table \ref{table:Q2} and Table \ref{table:D2}.
\begin{table*}[t]
\caption{Keyword extraction for Query: ``$how\; to \; fix\; bath\; tub\; wont\; turn\; off$''}
\label{table:Q2}
\begin{center}
\scriptsize
\begin{tabular}{ | c | c | c |c | c | c | c |c |c| }
\hline
     & \color{red}{$how$} & $to$ & \color{red}{$fix$}& \color{red}{$bath$}& \color{red}{$tub$}& $wont$& \color{red}{$turn$}& $off$\\ 
\hline
Number of assigned &  &  &  & & & & &  \\
cells out of 10 &  &  &  & & & & &  \\
Left to Right & - & 0 & \textbf{4} & \textbf{7} & \textbf{6}& 3& \textbf{5}& 0 \\ 
\hline
Number of assigned &  &  &  & & & & &  \\
cells out of 10 &  &  &  & & & & &  \\
Right to Left & \textbf{4} & 1 & \textbf{6} & \textbf{7} & \textbf{6} & \textbf{7} & \textbf{7} & - \\ 
\hline
\end{tabular}
\end{center}
\end{table*}
\begin{table*}[t]
\caption{Keyword extraction for Document: ``$how\; do\; you\; paint\; a\; bathtub\; and\; what\; paint\; should\dots$''}
\label{table:D2}
\begin{center}
\scriptsize
\begin{tabular}{ | c | c | c |c |c| c | c | c |c |c|c| }
\hline
     & $how$ & $do$ & $you$ & \color{red}{$paint$} & $a$ & \color{red}{$bathtub$} & $and$ & $what$ & \color{red}{$paint$}&$should\; you \dots$\\ 
\hline
Number of assigned &  &  &  &  &  &  &  &  & & \\
cells out of 10 &  &  &  &  &  &  &  &  & & \\
Left to Right & - & 1 & 1 & \textbf{7} & 0 & \textbf{9} & 2 & 3 & \textbf{8} & 4 \\ 
\hline
Number of assigned &  &  &  &  &  &  &  &  & & \\
cells out of 10 &  &  &  &  &  &  &  &  & & \\
Right to Left & \textbf{5} & \textbf{9} & \textbf{5} & \textbf{4} & \textbf{8} & \textbf{4} & \textbf{5} & \textbf{5} & \textbf{9} & - \\ 
\hline
\end{tabular}
\end{center}
\end{table*}

\section{Doc2Vec Similarity Test}
\label{sec:app:doc2vecTest}
To make sure that a meaningful model is trained, we used the trained doc2vec model to find the most similar words to two sample words in our dataset, the words ``pizza'' and ``infection''. The resulting words and corresponding scores are as follows: 

\hrulefill

\textcolor{blue}{print(model.most-similar('pizza')) :}\\

\textcolor{blue}{[(u'recipes', 0.9316294193267822),\\ 
(u'recipe', 0.9295548796653748),\\ 
(u'food', 0.9250608682632446),\\ 
(u'restaurants', 0.9223555326461792),\\ 
(u'bar', 0.9191627502441406),\\ 
(u'sabayon', 0.916868269443512),\\ 
(u'page', 0.9160783290863037),\\ 
(u'restaurant', 0.9112323522567749),\\ 
(u'house', 0.9104640483856201),\\ 
(u'the', 0.9103578925132751)]}

\hrulefill

\textcolor{blue}{print(model.most-similar('infection')):}\\

\textcolor{blue}{[(u'infections', 0.9698576927185059),\\ 
(u'treatment', 0.9143450856208801),\\ 
(u'symptoms', 0.9138627052307129),\\ 
(u'disease', 0.9100595712661743),\\ 
(u'palpitations', 0.9083651304244995),\\ 
(u'pneumonia', 0.9073051810264587),\\ 
(u'medical', 0.9043352603912354),\\ 
(u'abdomen', 0.9034136533737183),\\ 
(u'medlineplus', 0.9032401442527771),\\ 
(u'gout', 0.9027985334396362)]}

\hrulefill

As it is observed from the resulting words, the trained model is a meaningful model and can recognise semantic similarity.

\section{Diagram of the proposed model}
\label{sec:app:MoreClearDiagram}
\begin{figure*}[t]
\center
\includegraphics[width=1.0\textwidth]{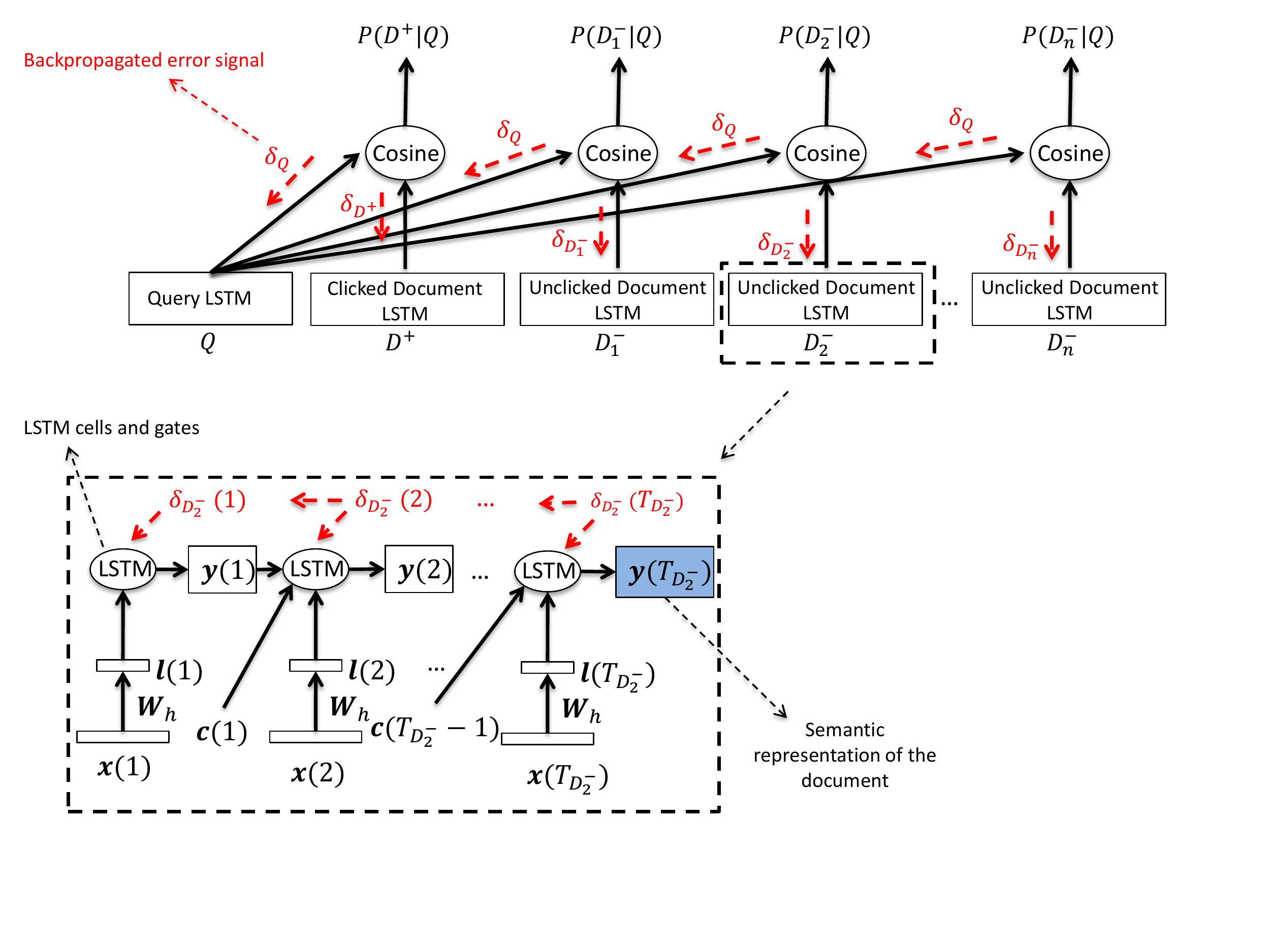}
\vspace{-1.5cm}
\caption{Architecture of the proposed method.}
   \label{fig:Cosine}
\end{figure*}

To clarify the difference between the proposed method and the general sentence embedding methods, in this section we present a diagram illustrating the training procedure of the proposed model. It is presented in Fig. \ref{fig:Cosine}. In this figure $n$ is the number of negative (unclicked) documents. The other parameters in this figure are similar to those used in Fig. 2 and Fig. 3 of the paper.

\end{document}